\documentclass{article}

\PassOptionsToPackage{numbers, compress}{natbib}
\PassOptionsToPackage{hyphens}{url}
\usepackage{hyperref}
\usepackage{xcolor}
\definecolor{myblue}{rgb}{0,0.22,0.45}
\hypersetup{
    colorlinks,
    linkcolor={black},
    citecolor={myblue},
    urlcolor={myblue}
}

\usepackage[final]{neurips_2020} 

\usepackage[noend]{algorithmic}
\usepackage[ruled,vlined,noend]{algorithm2e}

\usepackage{import}
\usepackage{microtype}
\usepackage{graphicx}
\usepackage{subfigure}
\makeatletter
\renewcommand{\thesubfigure}{\alph{subfigure}}
\renewcommand{\@thesubfigure}{(\thesubfigure)\hskip\subfiglabelskip}
\makeatother
\usepackage{wrapfig}
\usepackage{booktabs} 
\usepackage{xr-hyper}

\usepackage[utf8]{inputenc} 
\usepackage[T1]{fontenc}    
\usepackage{hyperref}       
\usepackage{url}            
\usepackage{booktabs}       
\usepackage{amsfonts}       
\usepackage{nicefrac}       
\usepackage{microtype}      

\usepackage{float} 
\usepackage{xspace}
\usepackage{dblfloatfix}
\usepackage{url}
\usepackage{enumitem}
\usepackage{multirow}
\usepackage{caption}
\usepackage{wrapfig}
\usepackage{float}
\usepackage{amssymb}
\usepackage{amsmath}
\usepackage{comment}
\usepackage{blindtext}
\usepackage{subfiles}

\newcommand{\ddpg}{\textsc{ddpg}\xspace}

\newcommand{\imgep}{\textsc{imgep}\xspace}

\newcommand{\her}{\textsc{her}\xspace}

\newcommand{\imagine}{\textsc{imagine}\xspace}

\newcommand{\SP}{\textsc{sp}\xspace}
\newcommand{\NL}{\textsc{nl}\xspace}
\newcommand{\MA}{\textsc{ma}\xspace}
\newcommand{\mar}{\textsc{ma}$^\mathcal{R}$\xspace}
\newcommand{\FA}{\textsc{fa}\xspace}
\newcommand{\far}{\textsc{fa}$^\mathcal{R}$\xspace}
\newcommand{\itwoc}{\textsc{i}\oldstylenums{2}\textsc{c}\xspace}
\newcommand{\SR}{\overline{\mathtt{SR}}}
\newcommand{\G}{\mathcal{G}}
\newcommand{\train}{\text{train}}
\newcommand{\test}{\text{test}}
\newcommand{\CGH}{\textsc{cgh}\xspace}

\newcommand{\balpha}{\boldsymbol{\alpha}}
\newcommand{\bbeta}{\boldsymbol{\beta}}

\newcommand{\bgamma}{\boldsymbol{\gamma}}

\newcommand{\bDelta}{\boldsymbol{\Delta}}

\newcommand{\bi}[1]{\textbf{\textit{#1}}}

\definecolor{myred}{rgb}{0.8,0,0}

\newcommand{\notinsubfile}[1]{}

\def\fauxschelper#1 #2\relax{%
  \fauxschelphelp#1\relax\relax%
  \if\relax#2\relax\else\ \fauxschelper#2\relax\fi%
}
\def\Hscale{.85}\def\Vscale{.74}\def\Cscale{1.12}
\def\fauxschelphelp#1#2\relax{%
  \ifnum`#1>``\ifnum`#1<`\{\scalebox{\Hscale}[\Vscale]{\uppercase{#1}}\else%
    \scalebox{\Cscale}[1]{#1}\fi\else\scalebox{\Cscale}[1]{#1}\fi%
  \ifx\relax#2\relax\else\fauxschelphelp#2\relax\fi}
  
\title{Language as a Cognitive Tool to\\Imagine Goals in Curiosity-Driven Exploration}

\author{C\'edric Colas\thanks{Equal contribution}~, Tristan Karch\footnotemark[1]\\ 
Inria - Flowers team\\
Universit\'e de Bordeaux\\
\texttt{\{firstname.lastname\}@inria.fr}\\
\And
Nicolas Lair\footnotemark[1]\\
Inserm U1093\\
Cloud Temple\\
\texttt{nicolas.lair@inserm.fr}\\
\And
Jean-Michel Dussoux\\
Cloud Temple\\
Paris\\
\AND
Clément Moulin-Frier\\
Inria - Flowers team\\
Universit\'e de Bordeaux\\ 
ENSTA ParisTech\\
\And
Peter Ford Dominey\\
Inserm U1093\\
Universit\'e de Dijon\\
\And
Pierre-Yves Oudeyer\\ 
Inria - Flowers team\\
Universit\'e de Bordeaux\\ 
ENSTA ParisTech\\

}

\makeatletter
\newcommand*{\addFileDependency}[1]{
  \typeout{(#1)}
  \@addtofilelist{#1}
  \IfFileExists{#1}{}{\typeout{No file #1.}}
}
\makeatother

\begin{document}
\maketitle

\begin{abstract}
Developmental machine learning studies how artificial agents can model the way children learn open-ended repertoires of skills. Such agents need to create and represent goals, select which ones to pursue and learn to achieve them. Recent approaches have considered goal spaces that were either fixed and hand-defined or learned using generative models of states. This limited agents to sample goals within the distribution of known effects. We argue that the ability to imagine out-of-distribution goals is key to enable creative discoveries and open-ended learning. Children do so by leveraging the compositionality of language as a tool to imagine descriptions of outcomes they never experienced before, targeting them as goals during play. We introduce \imagine, an intrinsically motivated deep reinforcement learning architecture that models this ability. Such imaginative agents, like children, benefit from the guidance of a social peer who provides language descriptions. To take advantage of goal imagination, agents must be able to leverage these descriptions to interpret their imagined out-of-distribution goals. This generalization is made possible by modularity: a decomposition between learned goal-achievement reward function and policy relying on deep sets, gated attention and object-centered representations. We introduce the Playground environment and study how this form of goal imagination improves generalization and exploration over agents lacking this capacity. In addition, we identify the properties of goal imagination that enable these results and study the impacts of modularity and social interactions.
\end{abstract}

\section{Introduction}
\begin{figure}[ht!]
  \centering
    \includegraphics[width=\textwidth]{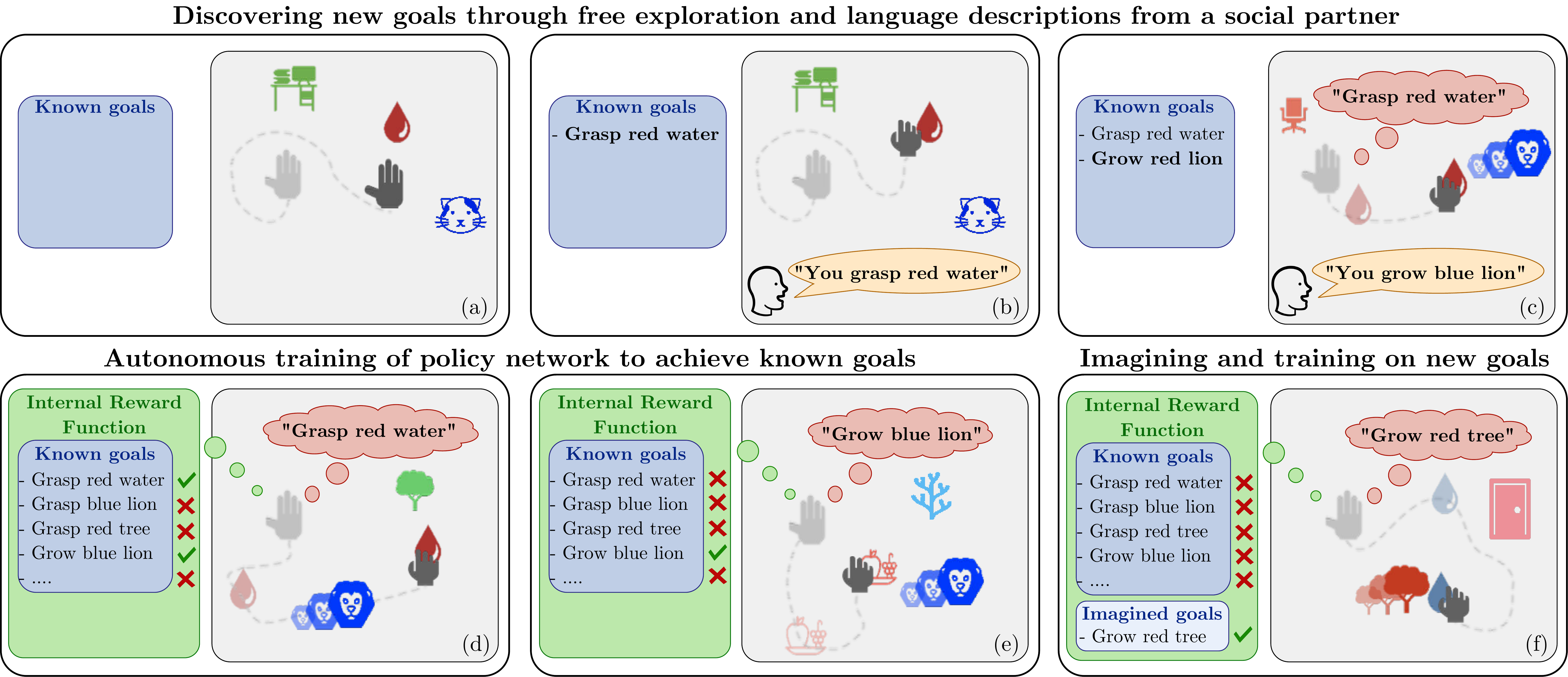}
  \caption{\textbf{\imagine overview}. In the \textit{Playground} environment, the agent (hand) can move, grasp objects and grow some of them.
  Scenes are generated procedurally with objects of different types, colors and sizes. A social partner provides descriptive feedback (orange), that the agent converts into targetable goals (red bubbles). \label{fig:main}
}
  \vspace{-.5cm}
\end{figure}

Building autonomous machines that can discover and learn open-ended skill repertoires is a long-standing goal in Artificial Intelligence. In this quest, we can draw inspiration from children development \cite{cangelosi2015developmental}. In particular, children exploration seems to be driven by intrinsically motivated brain processes that trigger spontaneous exploration for the mere purpose of experiencing novelty, surprise or learning progress \cite{gopnik1999scientist, kaplan2007search, kidd2015psychology}. During \textit{exploratory play}, children can also invent and pursue their own problems \cite{chu2020exploratory}.

Algorithmic models of intrinsic motivation were successfully used in developmental robotics \cite{oudeyer2007intrinsic,baldassarre2013intrinsically}, in reinforcement learning \cite{chentanez2005intrinsically,schmidhuber2010formal} and more recently in deep RL \cite{bellemare2016unifying,pathak2017curiosity}. Intrinsically Motivated Goal Exploration Processes (\imgep), in particular, enable agents to sample and pursue their own goals without external rewards \cite{baranes2013active,forestier2016modular,imgep} and can be formulated within the deep RL framework \cite{florensa2017automatic,nair2018visual,curious,pong2019skew,venkattaramanujam2019self,racaniere2019automated}. However, representing goal spaces and goal-achievement functions remains a major difficulty and often requires hand-crafted definitions. Past approaches proposed to learn image-based representations with generative models such as Variational Auto-Encoders \cite{laversanne2018curiosity,nair2018visual}, but were limited to the generation of goals within the distribution of already discovered effects. Moving beyond \textit{within-distribution} goal generation, \textit{out-of-distribution} goal generation could power creative exploration in agents, a challenge that remains to be tackled. 

In this difficult task, children leverage the properties of language to assimilate thousands of years of experience embedded in their culture, in a only a few years \cite{Tomasello1999,Bruner1991}. As they discover language, their goal-driven exploration changes. \citet{Piaget1926} first identified a form of egocentric speech where children narrate their ongoing activities. Later, \citet{Vygotskii1978}
realized that they were generating novel plans and goals by using the expressive generative properties of language. The harder the task, the more children used egocentric speech to plan their behavior \citep[chap. 2]{Vygotskii1978}. Interestingly, this generative capability can push the limits of the real, as illustrated by \citet{Chomsky1957}’s famous example of a sentence that is syntactically correct but semantically original “\textit{Colorless green ideas sleep furiously}”. Language can thus be used to generate out-of-distributions goals by leveraging compositionality to imagine new goals from known ones.

This paper presents \textbf{I}ntrinsic \textbf{M}otivations \textbf{A}nd \textbf{G}oal \textbf{IN}vention for \textbf{E}xploration (\imagine): a learning architecture which leverages natural language (\NL) interactions with a descriptive social partner (\SP) to explore procedurally-generated scenes and interact with objects. \imagine discovers meaningful environment interactions through its own exploration (Figure~\ref{fig:main}a) and episode-level \NL descriptions provided by \SP (\ref{fig:main}b). These descriptions are turned into targetable goals by the agent (\ref{fig:main}c). The agent learns to represent goals by jointly training a language encoder mapping \NL  to goal embeddings and a goal-achievement reward function (\ref{fig:main}d). The latter evaluates whether the current scene satisfies any given goal. These signals (ticks in Figure~\ref{fig:main}d-e) are then used as training signals for policy learning. More importantly, \imagine can invent new goals by composing known ones (\ref{fig:main}f). Its internal goal-achievement function allows it to train autonomously on these imagined goals.

    
\paragraph{Related work.} The idea that language understanding is grounded in one's experience of the world and should not be secluded from the perceptual and motor systems has a long history in Cognitive Science \cite{Glenberg2002,Zwaan05}. This vision was transposed to intelligent systems \cite{steels2006semiotic,mcclelland2019extending}, applied to human-machine interaction \cite{Dominey2005, Madden2010} and recently to deep RL via frameworks such as \textit{BabyAI} \cite{chevalier-boisvert2018babyai}.

In their review of \textit{RL algorithms informed by NL}, \citet{Luketina2019} distinguish between \textit{language-conditional} problems where language is required to solve the task and \textit{language-assisted} problems where language is a supplementary help. In the first category, most works propose instruction-following agents \cite{Branavan2010, Chen2011, bahdanau2018learning, coreyes2018guiding, Jiang2019, Goyal2019, ther}. Although our system is \textit{language-conditioned}, it is not \textit{language-instructed}: it is never given any instruction or reward but sets its own goals and learns its own internal reward function. \citet{bahdanau2018learning} and \citet{fu2018from} also learn a reward function but require extensive expert knowledge (expert dataset and known environment dynamics respectively), whereas our agent uses experience generated by its own exploration.

Language is also particularly well suited for Hindsight Experience Replay \cite{andrychowicz2017hindsight}: descriptions of the current state can be used to relabel trajectories, enabling agents to transfer skills across goals. While previous works used a hard-coded descriptive function \cite{chan2019actrce, Jiang2019} or trained a generative model \cite{ther} to generate goal substitutes, we leverage the learned reward function to scan goal candidates.

To our knowledge, no previous work has considered the use of compositional goal imagination to enable creative exploration of the environment.
The linguistic basis of our goal imagination mechanism is grounded in construction grammar (CG). CG is a usage-based approach that characterizes language acquisition as a trajectory starting with pattern imitation and the discovery of equivalence classes for argument substitution, before evolving towards the recognition and composition of more abstract patterns \cite{tomasello2000item,goldberg2003constructions}. This results in a structured inventory of constructions as form-to-meaning mappings that can be combined to create novel utterances \cite{goldberg2003constructions}. The discovery and substitution of equivalent words in learned schemas is observed directly in studies of child language \cite{tomasello1993twenty, tomasello2000item}.
Computational implementations of this approach have demonstrated its ability to foster generalization \cite{hinaut2013real} and was also used for data augmentation to improve the performance of neural seq2seq models in NLP \cite{andreas2019goodenough}.

Imagining goals by composing known ones only works in association with \textit{systematic generalization} \cite{bahdanau2018systematic,hill2019emergent}: generalizations of the type \textit{grow any animal + grasp any plant $\to$ grow any plant}. These were found to emerge in instruction-following agents, including generalizations to new combinations of motor predicates, object colors and shapes \cite{Hermann2017,hill2019emergent, bahdanau2018learning}. Systematic generalization can occur when objects share common attributes (e.g. type, color). 
We directly encode that assumption into our models by representing objects as \textit{single-slot object files} \cite{green2017object}: separate entities characterized by shared attributes. Because all objects have similar features, we introduce a new object-centered inductive bias: object-based modular architectures based on Deep Sets \cite{deepset}.

\paragraph{Contributions.} This paper introduces:
\begin{enumerate}
    \item 
     The concept of imagining new goals using language compositionality to drive exploration.
    \item 
    \imagine: an intrinsically motivated agent that uses goal imagination to explore its environment, discover and master object interactions by leveraging \NL descriptions from a social partner.
    \item 
    Modular policy and reward function with systematic generalization properties enabling \imagine to train on imagined goals. Modularity is based on Deep Sets, gated attention mechanisms and object-centered representations.
    \item 
    \textit{Playground}: a procedurally-generated environment designed to study several types of generalizations (across predicates, attributes, object types and categories).
    \item 
    A study of \imagine investigating: 1) the effects of our goal imagination mechanism on generalization and exploration; 2) the identification of general properties of imagined goals required for any algorithm to have a similar impact; 3) the impact of modularity and 4) social interactions. 
    \end{enumerate}

\section{Problem Definition}
\label{sec:pb_def}
\paragraph{Open-ended learning environment.} We consider a setup where agents evolve in an environment filled with objects and have no prior on the set of possible interactions. An agent decides what and when to learn by setting its own goals, and has no access to external rewards.

However, to allow the agent to learn relevant skills, a social partner (\SP) can watch the scene and plays the role of a human caregiver. Following a developmental approach \cite{asada2009cognitive}, we propose a hard-coded surrogate \SP that models important aspects of the developmental processes seen in humans:
\begin{itemize}
    \item 
    At the beginning of each episode, the agent chooses a goal by formulating a sentence. \SP then provides agents with optimal learning opportunities by organizing the scene with: 1) the required objects to reach the goal (not too difficult) 2) procedurally-generated distracting objects (not too easy and providing further discovery opportunities). This constitutes a developmental scaffolding modelling the process of Zone of Proximal Development (ZPD) introduced by Vygotsky to describe infant-parent learning dynamics \cite{Vygotskii1978}. 
    \item 
    At the end of each episode, \SP utters a set of sentences describing achieved and meaningful outcomes (except sentences from a test set). Linguistic guidance given through descriptions are a key component of how parents "teach" language to infants, which contrasts with instruction following (providing a linguistic command and then a reward), that is rarely seen in real parent-child interactions \cite{tomasello2009constructing, bornstein1992maternal}. By default, \SP respects the $3$ following properties: \textit{precision}: descriptions are accurate, \textit{exhaustiveness}: it provides all valid descriptions for each episode and \textit{full-presence}: it is always available. Section~\ref{sec:social_feedbacks} investigates relaxations of the last two assumptions. 
\end{itemize}

Pre-verbal infants are known to acquire object-based representations very early \cite{spelke1992origins,johnson2003development} and, later, to benefit from a simplified parent-child language during language acquisition \cite{mintz2003frequent}. Pursuing a developmental approach \cite{asada2009cognitive}, we assume corresponding object-based representations and a simple grammar. As we aim to design agents that bootstrap creative exploration without prior knowledge of possible interactions or language, we do not consider the use of pre-trained language models.

\paragraph{Evaluation metrics.} This paper investigates how goal imagination can lead agents to efficiently and creatively explore their environment to discover interesting interactions with objects around. In this quest, \SP guides agents towards a set of interesting outcomes by uttering \NL descriptions. Through compositional recombinations of these sentences, goal imagination aims to drive creative exploration, to push agents to discover outcomes beyond the set of outcomes known by \SP. We evaluate this desired behavior by three metrics: 1) the generalization of the policy to new states, using goals from the training set that \SP knows and describes; 2) the generalization of the policy to new language goals, using goals from the testing set unknown to \SP; 3) goal-oriented exploration metrics. These measures assess the quality of the agents' intrinsically motivated exploration. Measures 1) and 2) are also useful to assess the abilities of agents to learn language skills. We measure generalization for each goal as the success rate over $30$ episodes and report $\SR$ the average over goals. We evaluate exploration with the \textit{interesting interaction count} (\itwoc). \itwoc is computed on different sets of interesting interactions: behaviors a human could infer as goal-directed. These sets include the training, testing sets and an extra set containing interactions such as bringing water or food to inanimate objects. $\itwoc_\mathcal{I}$ measures the number of times interactions from $\mathcal{I}$ were observed over the last epoch ($600$ episodes), whether they were targeted or not (see Supplementary Section~\ref{sec:suppl_exploration}). Thus, \itwoc measures the penchant of agents to explore interactions with objects around them. Unless specified  otherwise, we provide means $\mu$ and standard deviations over $10$ seeds and report statistical significance using a two-tail Welch's t-test with null hypothesis $\mu_1=\mu_2$, at level $\alpha=0.05$ (noted by star and circle markers in figures) \cite{colas2019hitchhiker}.

\section{Methods}
\label{sec:methods}

\subsection{The \textit{Playground} environment}
\label{sec:env}
We argue that the study of new mechanisms requires the use of controlled environments. We thus introduce \textit{Playground}, a simple environment designed to study the impact of goal imagination on exploration and generalization by disentangling it from the problems of perception and fully-blown \NL understanding. The \textit{Playground} environment is a continuous $2$D world, with procedurally-generated scenes containing $\textsc{n}=3$ objects, from $32$ different object types (\textit{e.g. dog, cactus, sofa, water, etc.}), organized into $5$ categories (\textit{animals, furniture, plants, etc}), see Figure~\ref{fig:main}. To our knowledge, it is the first environment that introduces object categories and category-dependent combinatorial dynamics, which allows the study of new types of generalization. We release \textit{Playground} in a separate repository.\footnote{\url{https://github.com/flowersteam/playground\_env}}
\paragraph{Agent perception and embodiment.} Agents have access to state vectors describing the scene: the agent's body and the objects. Each object is represented by a set of features describing its type, position, color, size and whether it is grasped. Categories are not explicitly encoded. Objects are made unique by the procedural generation of their color and size. The agent can perform bounded translations in the $2$D plane, grasp and release objects with its gripper. It can make animals and plants grow by bringing them the right supply (food or water for animals, water for plants).  

%
\paragraph{Grammar.}The following grammar generates the descriptions of the $256$ achievable goals ($\G^\text{A}$): 
\begin{enumerate}
    \item 
    Go: <\textit{go} + \bi{zone}>  \textit{(e.g. go bottom left) }
    \item 
    Grasp: < \textit{grasp} + \textit{any} + \bi{color} + \textit{thing}>  \textit{(e.g. grasp any blue thing)} OR \\
    <\textit{grasp} + \bi{color} $\cup$ \{\textit{any}\}  + \textbf{\textit{object type $\cup$ object category}}> \textit{(e.g. grasp red cat)}
    \item 
    Grow: <\textit{grow} + \textit{any} + \bi{color} + \textit{thing}> \textit{(e.g. grow any red thing)} OR \\
    <\textit{grow} + \bi{color} $\cup$ \{\textit{any}\} + \bi{living thing} $\cup$ \{\textit{living\_thing, animal, plant}\}> \textit{(e.g. grow green animal)}
\end{enumerate} 

\textbf{Bold} and \{ \} are sets of words while \textit{italics} are specific words. 
The grammar is structured around the $3$ predicates \textit{go}, \textit{grasp} and \textit{grow}. Objects can be referred to by a combination of their color and either their object name or category, or simply by one of these.
The set of achievable goals is partitioned into \textit{training} $(\G^\train)$ and \textit{testing}  $(\G^\test)$. $\G^\test$ maximizes the compound divergence with a null atom divergence with respect to $\G^\train$: testing sentences (compounds) are out of the distribution of $\G^\train$ sentences, but their words (atoms) belong to the distribution of words in $\G^\train$ \cite{keysers2019measuring}. \SP only provides descriptions from $\G^\train$. We limit the set of goals to better control the complexity of our environment and enable a careful study of the generalization properties. Supplementary Section~\ref{sec:suppl_env_descr} provides more details about the environment, the grammar and \SP as well as the pseudo-code of our learning architecture.


\subsection{The \imagine Architecture}
\label{sec:architecture}
\begin{figure}[!h]
    \centering
     \includegraphics[width=.7\textwidth]{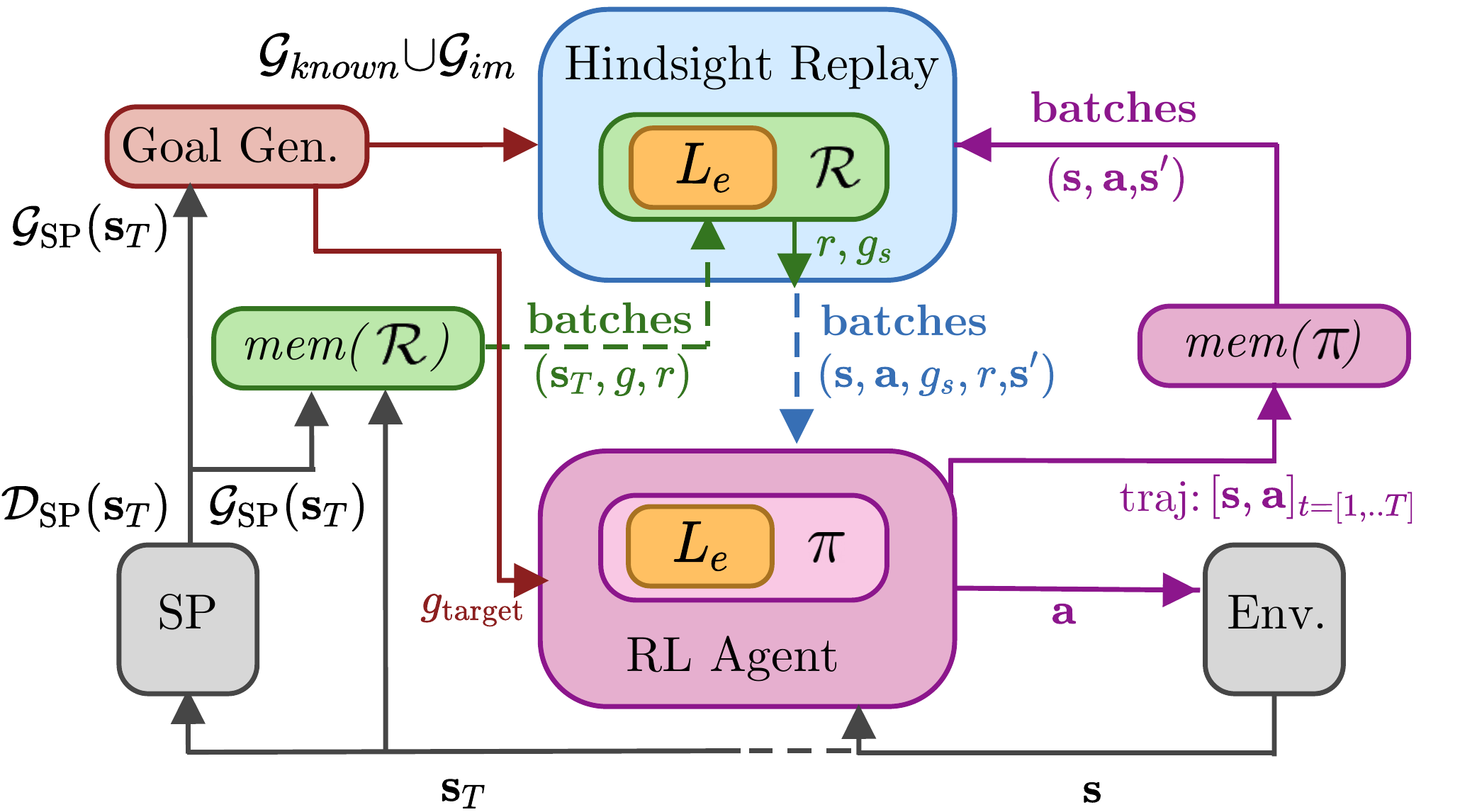}
    
     \caption{\textbf{\imagine architecture.} Colored boxes show the different modules of \imagine. Lines represent update signals (dashed) and function outputs (plain). The language encoder $L_e$ is shared.}  
    \vspace{-0.25cm}
\label{fig:architecture}
\end{figure}

\imagine agents build a repertoire of goals and train two internal models: 1) a goal-achievement reward function $\mathcal{R}$ to predict whether a given description matches a behavioral trajectory; 2) a policy $\pi$ to achieve behavioral trajectories matching descriptions. The architecture is presented in Figure~\ref{fig:architecture} and follows this logic:  
\begin{enumerate}
    \item The \textit{Goal Generator} samples a target goal $g_{\text{target}}$ from known and imagined goals $(\G_\text{known} \cup \G_\text{im})$.
    \item The agent (\textit{RL Agent}) interacts with the environment using its policy $\pi$ conditioned on $g_\text{target}$.
    \item State-action trajectories are stored in a replay buffer \textit{mem$(\pi)$}.
    \item \SP's descriptions of the last state are considered as potential goals $\G_{\text{\SP}}(\mathbf{s}_T)~=~\mathcal{D}_{\text{\SP}}(\textbf{s}_T)$.
    \item \textit{mem$(\mathcal{R})$} stores positive pairs $(\mathbf{s}_T,~ \G_{\text{\SP}}(\mathbf{s}_T))$ and infers negative pairs $(\mathbf{s}_T,~\G_\text{known} \setminus \G_{\text{\SP}}(\mathbf{s}_T))$.
    \item The agent then updates:
    \begin{itemize}
        \item \textit{Goal Gen.}: $\G_\text{known}~\gets~\G_\text{known} \cup \G_{\text{\SP}}(\textbf{s}_T)$ and  $\G_{im}~\gets~\text{Imagination}(\G_\text{known})$.
        \item \textit{Language Encoder} $(L_e)$ and \textit{Reward Function} $(\mathcal{R})$ are updated using data from \textit{mem$(\mathcal{R})$}.
        \item \textit{RL agent}: We sample a batch of state-action transitions $(\mathbf{s},~\mathbf{a},~\mathbf{s}')$ from \textit{mem$(\pi)$}. Then, we use \textit{Hindsight Replay} and $\mathcal{R}$ to bias the selection of substitute goals to train on $(g_\text{s})$ and compute the associated rewards $(\mathbf{s},~\mathbf{a},~\mathbf{s'},~g_\text{s},~r)$. Substituted goals $g_s$ can be known or imagined goals. Finally, the policy and critic are trained via RL.
    \end{itemize}
\end{enumerate}

\paragraph{Goal generator.} It is a generative model of \NL  goals. It generates target goals $g_\text{target}$ for data collection and substitutes goals $g_\text{s}$ for hindsight replay. When goal imagination is disabled, the goal generator samples uniformly from the set of known goals $\G_\text{known}$, sampling random vectors if empty. When enabled, it samples with equal probability from $\G_\text{known}$ and $\G_\text{im}$ (set of imagined goals). $\G_\text{im}$ is generated using a mechanism grounded in construction grammar that leverages the compositionality of language to imagine new goals from $\G_\text{known}$. The heuristic consists in computing sets of \textit{equivalent words}: words that appear in two sentences that only differ by one word. For example, from \textit{grasp red lion} and \textit{grow red lion}, \textit{grasp} and \textit{grow} can be considered \textit{equivalent} and from \textit{grasp green tree} one can imagine a new goal \textit{grow green tree} (see Figure~\ref{fig:main}f). Imagined goals do not include known goals. Among them, some are meaningless, some are syntactically correct but infeasible (e.g. \textit{grow red lamp}) and some belong to $\G^\text{test}$, or even to $\G^\train$ before they are encountered by the agent and described by \SP. The pseudo-code and all imaginable goals are provided in Supplementary Section~\ref{sec:suppl_goal_imagination}.


\paragraph{Language encoder.} The language encoder $(L_e)$ embeds \NL  goals $(L_e:\G^\text{\NL} \to \mathbb{R}^{100})$ using an LSTM \cite{hochreiter1997lstm} trained jointly with the reward function.  $L_e$ acts as a goal translator, turning the goal-achievement reward function, policy and critic into language-conditioned functions.


\paragraph{Object-centered modular architectures.} The goal-achievement reward function, policy and critic leverage novel \textit{modular-attention} (\MA) architectures based on Deep Sets \cite{deepset}, gated attention mechanisms \cite{chaplot2017gatedattention} and object-centered representations. The idea is to ensure efficient skill transfer between objects, no matter their position in the state vector. This is done through the combined use of a shared neural network that encodes object-specific features and a permutation-invariant function to aggregate the resulting latent encodings. The shared network independently encodes, for each object, an affordance between this object (object observations), the agent (body observations) and its current goal. The goal embedding, generated by $L_e$, is first cast into an attention vector in $[0,~1]$, then fused with the concatenation of object and body features via an Hadamard product (gated-attention \cite{chaplot2017gatedattention}). The resulting object-specific encodings are aggregated by a permutation-invariant function and mapped to the desired output via a final network (e.g. into actions or action-values). Supplementary Section~\ref{sec:suppl_archi} provides visual representations.

\paragraph{Reward function.} Learning a goal-achievement reward function $(\mathcal{R})$ is framed as binary classification: $\mathcal{R}(\mathbf{s}, \mathbf{g}):~\mathcal{S} \times \mathbb{R}^{100} \to \{0, 1\}$. We use the \MA architecture with attention vectors $\balpha^g$, a shared network $\textsc{nn}^\mathcal{R}$ with output size $1$ and a logical OR aggregation. $\textsc{nn}^\mathcal{R}$ computes object-dependent rewards $r_{i}$ in $[0,1]$ from the object-specific inputs and the goal embedding. The final binary reward is computed by \textsc{nn}$^\textsc{or}$ which outputs $1$ whenever $ \exists j:~r_{j}~>~0.5$. We pre-trained a neural-network-based \textsc{or} function to enable end-to-end training with back-propagation. The overall function is:

\begin{equation*}
   \mathcal{R}(\mathbf{s}, g)~=~\textsc{nn}^\textsc{OR}([\textsc{nn}^\mathcal{R}(\mathbf{s}_{obj(i)} \odot \balpha^g)]_{i\in[1..N]})
\end{equation*}

\textit{Data.} 
Interacting with the environment and \SP, the agent builds a set of entries $[\mathbf{s}_T,~ g,~ r]$ with $g~\in~\G_\text{known}$ where $r \in \{0,~1\}$ rewards the achievement of $g$ in state $\mathbf{s}_T$: $r~=~1$ if $g \in \G_\text{\SP}(\mathbf{s}_T)$ and $0$ otherwise. $L_e$ and $\mathcal{R}$ are periodically updated jointly by back-propagation on this dataset.


\paragraph{Multi-goal RL agent.} Our agent is controlled by a goal-conditioned policy $\pi$ \cite{schaul2015universal} based on the \MA architecture (see Supplementary Figure~\ref{fig:archi_MA}). It uses an attention vector $\bbeta^g$, a shared network \textsc{nn}$^\pi$, a sum aggregation and a mapper \textsc{nn}$^\text{a}$ that outputs the actions. Similarly, the critic produces action-values via $\bgamma^g$, \textsc{nn}$^Q$ and $\textsc{nn}^\text{a-v}$ respectively: 

\begin{align*}
    \pi(\mathbf{s}, g)~&=~\textsc{nn}^\text{a}(\sum_{i \in [1..N]}\textsc{nn}^\pi(\mathbf{s}_{obj(i)} \odot \bbeta^g)) 
    \qquad 
    Q(\mathbf{s}, \mathbf{a}, g)~=~\textsc{nn}^\text{a-v}(\sum_{i \in [1..N]}\textsc{nn}^Q([\mathbf{s}_{obj(i)},~\mathbf{a}] \odot \bgamma^g)).
\end{align*}

Both are trained using \ddpg \cite{lillicrap2015continuous}, although any other off-policy algorithm can be used. As detailed in Supplementary Section~\ref{sec:suppl_reward}, our agent uses a form of Hindsight Experience Replay \cite{andrychowicz2017hindsight}.
%

\section{Experiments and Results}
This section first showcases the impact of goal imagination on exploration and generalization (Section~\ref{sec:res_imag_exp}). For a more complete picture, we analyze other goal imagination mechanisms and investigate the properties enabling these effects (Section~\ref{sec:res_im_properties}). Finally, we show that our modular architectures are crucial to a successful goal imagination (Section~\ref{sec:res_archi}) and discuss more realistic interactions with \SP (Section~\ref{sec:social_feedbacks}). \imagine agents achieve near perfect generalizations to new states (training set of goals): $\SR~=~0.95\pm0.05$. We thus focus on language generalization and exploration. Supplementary Sections~\ref{sec:supp_focus_gene} to \ref{sec:suppl_visu} provide additional results and insights organized by theme (Generalization, Exploration, Goal Imagination, Architectures, Reward Function and Visualizations).

\subsection{How does Goal Imagination Impact Generalization and Exploration?}
\label{sec:res_imag_exp}

\begin{figure}[!h]
    \centering
    \subfigure[\label{fig:goal_invention_all}]{\includegraphics[width=0.323\textwidth]{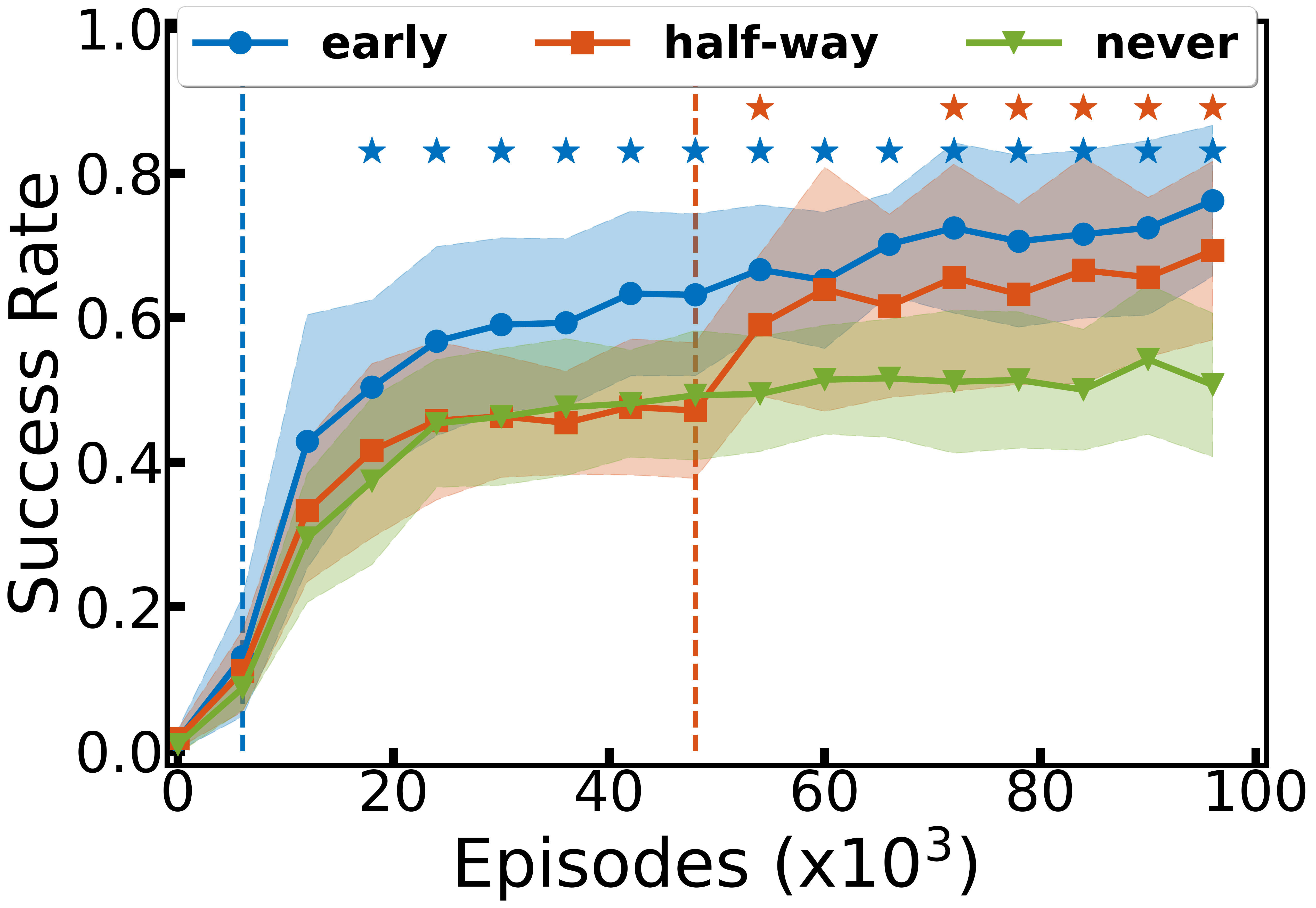}}
    \subfigure[\label{fig:adaptation}]{\includegraphics[width=0.323\textwidth]{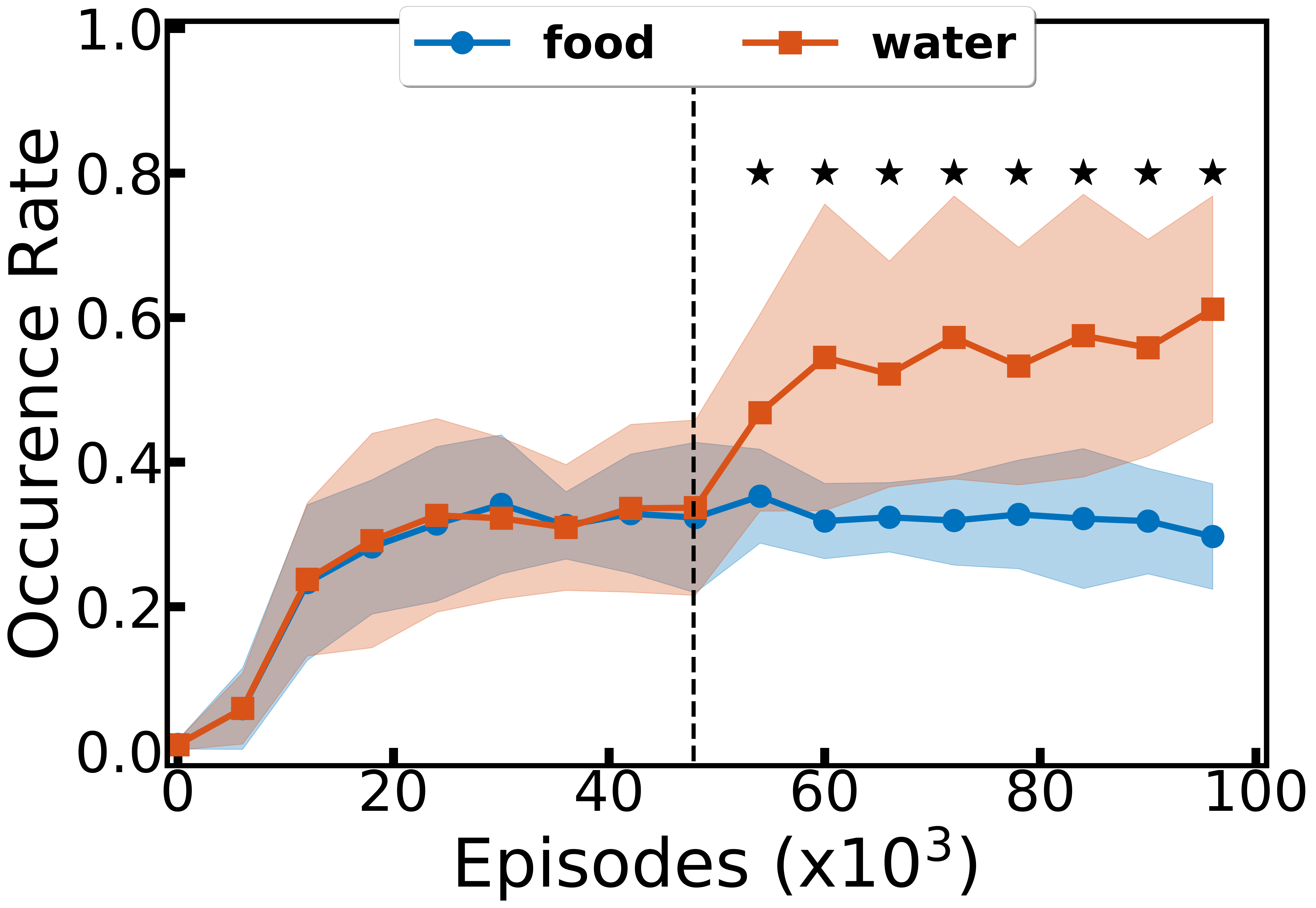}}
    \subfigure[\label{fig:goal_invention_explo}]{\includegraphics[width=0.323\textwidth]{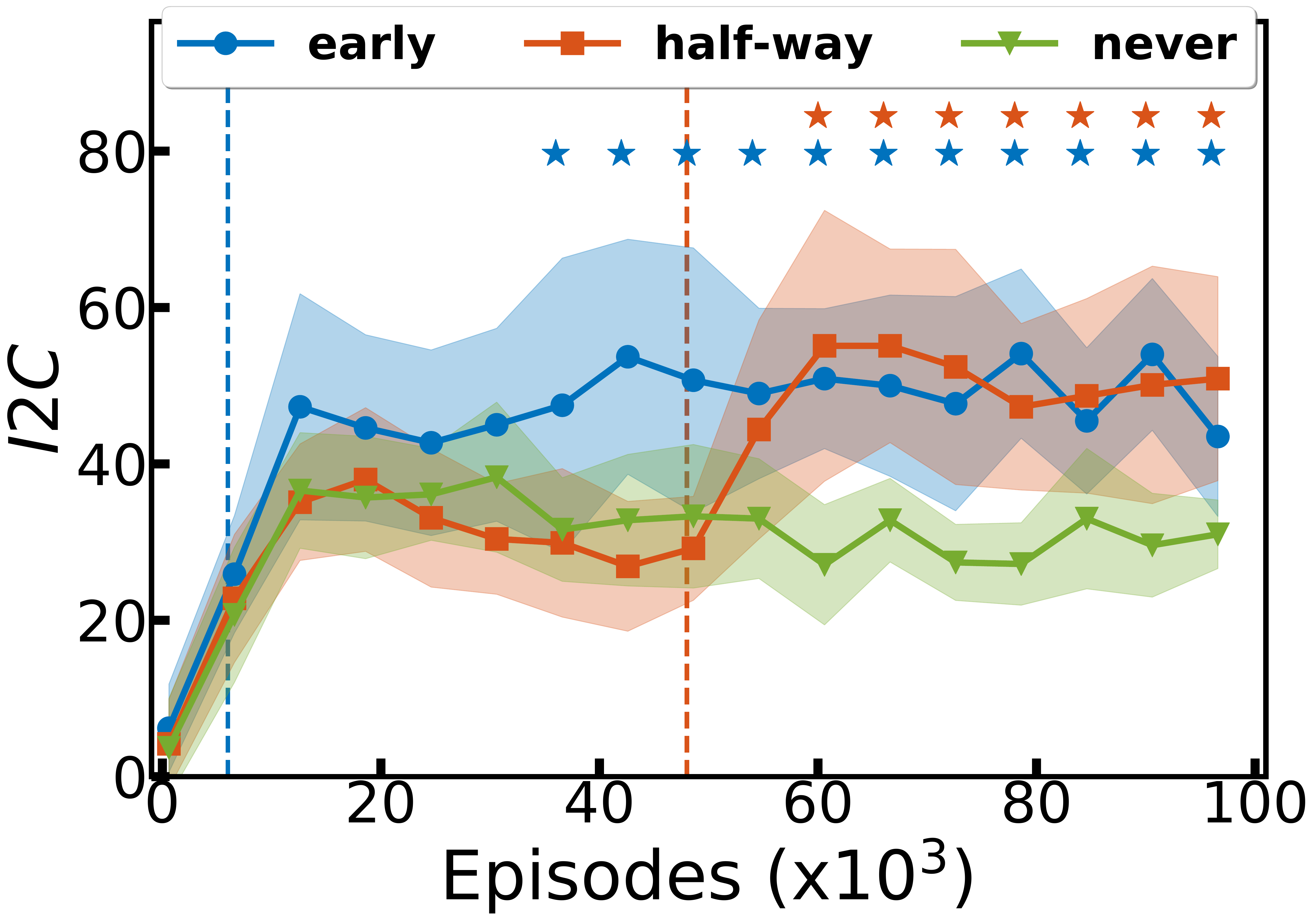}}
    \caption{\textbf{Goal imagination drives exploration and generalization.} Vertical dashed lines mark the onset of goal imagination. (a) $\SR$ on testing set. (b) Behavioral adaptation, empirical probabilities that the agent brings supplies to a plant when trying to grow it. (c) \itwoc computed on the testing set. Stars indicate significance (a and c are tested against \textit{never}).}
    \vspace{-.3cm}
\end{figure}

\paragraph{Global generalization performance.} 
Figure~\ref{fig:goal_invention_all} shows $\SR$ on the set of testing goals, when the agent starts imagining new goals early (after $6\cdot10^3$ episodes), half-way (after $48\cdot10^3$ episodes) or when not allowed to do so. Imagining goals leads to significant improvements in generalization.

\paragraph{A particular generalization: growing plants.} Agents learn to grow animals from \SP's descriptions, but are never told they could grow plants. When evaluated offline on the \textit{growing-plants} goals before goal imagination, agents' policies perform a sensible zero-shot generalization and bring them water or food with equal probability, as they would do for animals (Figure~\ref{fig:adaptation}, left). As they start to imagine and target these goals, their behavior adapts (Figure~\ref{fig:adaptation}, right). If the reward function shows good zero-shot abilities, it only provides positive rewards when the agent brings water. The policy slowly adapts to this internal reward signal and pushes agents to bring more water. We call this phenomenon \textit{behavioral adaptation}. Supplementary Section~\ref{sec:supp_focus_gene} details the generalization abilities of \imagine for $5$ different types of generalizations involving predicates, attributes and categories.

\paragraph{Exploration.} 
Figure~\ref{fig:goal_invention_explo} presents the \itwoc metric computed on the set of interactions related to $\G^\test$ and demonstrates the exploration boost triggered by goal imagination. Supplementary Section~\ref{sec:suppl_exploration} presents other \itwoc metrics computed on additional interactions sets.

\subsection{What If We Used Other Goal Imagination Mechanisms?}
\label{sec:res_im_properties}

\begin{figure}[!h]
\begin{minipage}{.3\textwidth} %
    \centering
    \vspace{.2cm}
    \begin{tabular}{l|cc}
         & \textbf{Cov.} & \textbf{Pre.}\\
         \hline
        \textbf{\CGH} & 0.87 & 0.45  \\
        \textbf{Oracle} &  1 & 1 \\
        \textbf{Low Cov.} & 0.44 & 0.45 \\
        \textbf{Low Pre.} & 0.87 & 0.30 \\
        \textbf{Random G.} & $\approx$0 & $\approx$0 \\
    \end{tabular}
    \\
    \vspace{1.05cm}
    \textit{ }\\
    (a)
\end{minipage}
\hfill
\begin{minipage}{.34\textwidth} %
    \centering
      \includegraphics[width=0.98\linewidth]{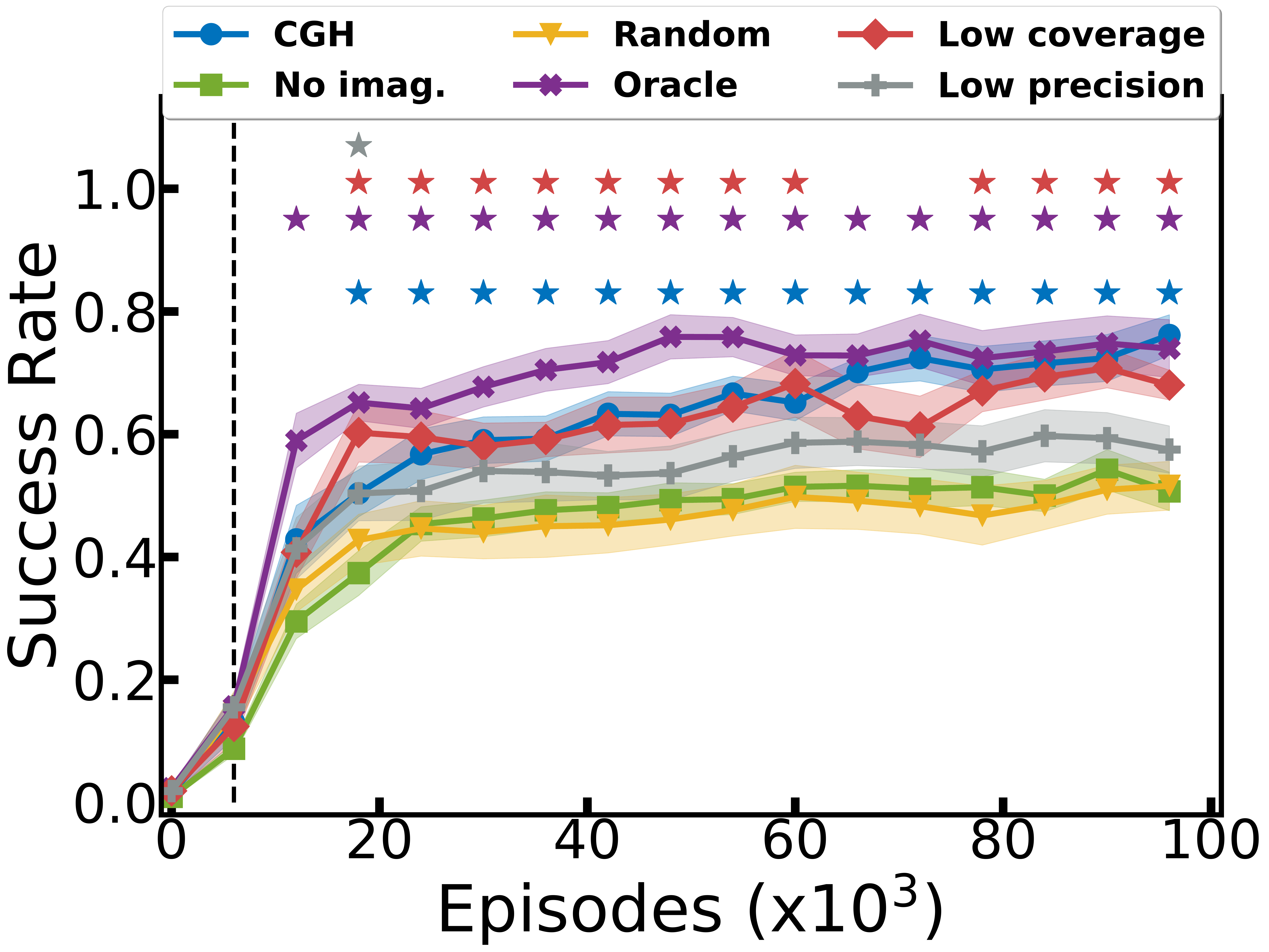}\\
      (b)
\end{minipage} %
\hfill
\begin{minipage}{.34\textwidth} %
    \centering
    \includegraphics[width=0.98\linewidth]{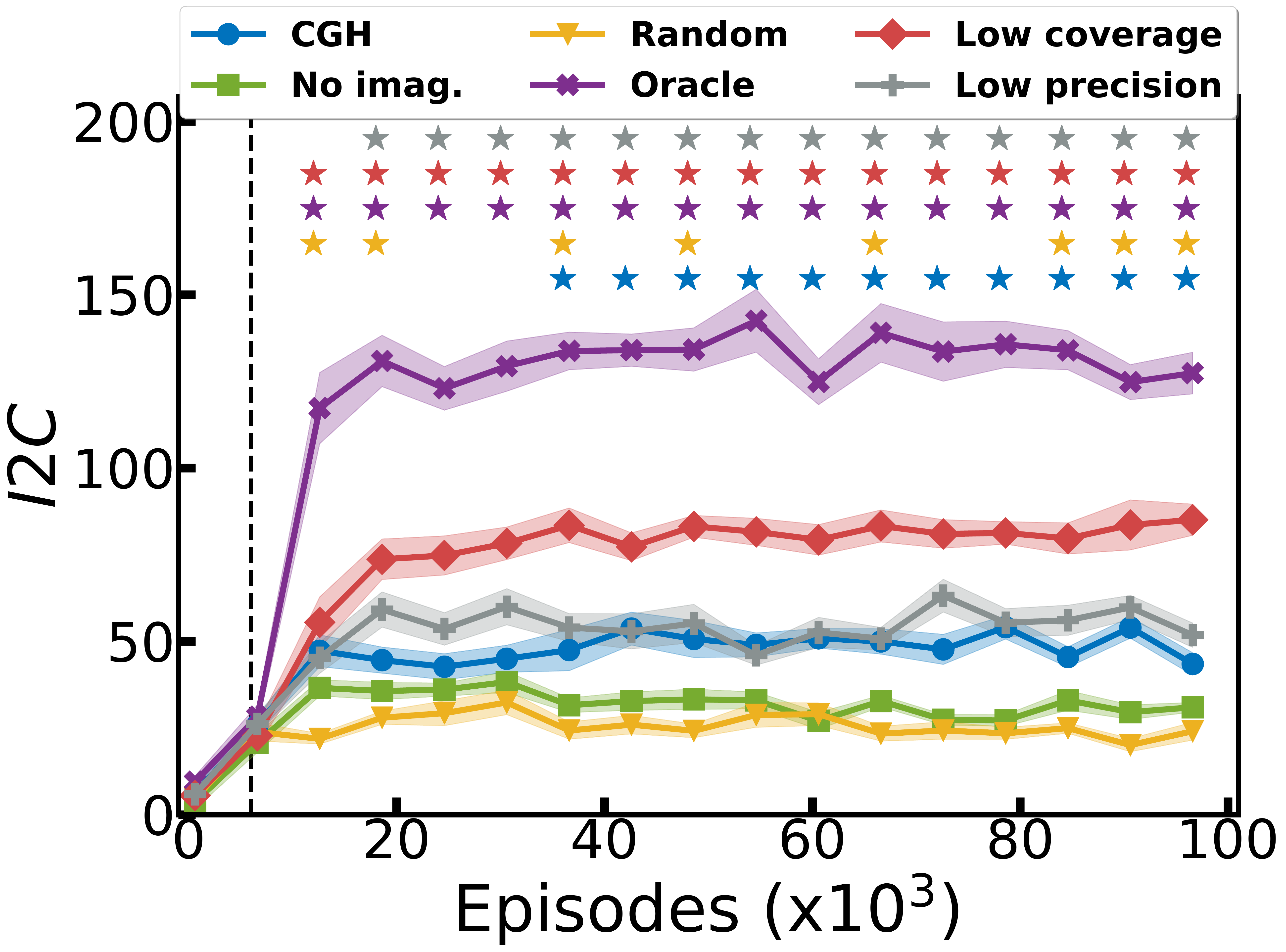}\\
    (c)
\end{minipage} %

\caption{\textbf{Goal imagination properties.} (a) Coverage and precision of different goal imagination heuristics. (b) $\SR$ on testing set. (c) \itwoc on $\G^\test$. We report \textit{sem} (standard error of the mean) instead of \textit{std} to improve readability. Stars indicate significant differences w.r.t the \textit{no imagination} condition.\label{fig:goal_invention_properties_explo}}
\end{figure}

\paragraph{Properties of imagined goals.} We propose to characterize goal imagination mechanisms by two properties: 1) \textit{Coverage}: the fraction of $\mathcal{G^\text{test}}$ found in $\mathcal{G_\text{im}}$ and 2) \textit{Precision}: the fraction of the imagined goals that are achievable. We compare our goal imagination mechanism based on the construction grammar heuristic (\CGH) to variants characterized by 1) lower coverage; 2) lower precision; 3) perfect coverage and precision (oracle); 4) random goal imagination baseline (random sequences of words from $\G^\train$ leading to near null coverage and precision). These measures are computed at the end of experiments, when all goals from $\G^\train$ have been discovered (Figure~\ref{fig:goal_invention_properties_explo}a).

Figure~\ref{fig:goal_invention_properties_explo}b shows that \CGH achieves a generalization performance on par with the oracle. Reducing the coverage of the goal imagination mechanism still brings significant improvements in generalization. Supplementary Section~\ref{sec:suppl_goal_imagination} shows, for the \textit{Low Coverage} condition, that the generalization performance on the testing goals that were imagined is not statistically different from the performance on similar testing goals that could have been imagined but were not. This implies that the generalization for imagined goals also benefits similar non-imagined goals from $\G^\text{test}$. Finally, reducing the precision of imagined goals (gray curve) seems to impede generalization (no significant difference with the \textit{no imagination} baseline). Figure~\ref{fig:goal_invention_properties_explo}c shows that all goal imagination heuristics enable a significant exploration boost. The random goal baseline acts as a control condition. It demonstrates that the generalization boost is not due to a mere effect of network regularization introduced by adding random goals (no significant effect w.r.t. the \textit{no imagination} baseline). In the same spirit, we also ran a control using random goal embeddings, which did not produce any significant effects.

\subsection{How Does Modularity Interact with Goal Imagination?}
\label{sec:res_archi}

\begin{wraptable}{r}{5cm}
    \vspace{-0.3cm}
    \caption{Policy architectures performance. 
    $\SR_\text{test}$ at convergence.}
    \label{tab:archi_comparison}
    \centering
    \begin{tabular}{l|cc}
    & \MA$^\textbf{*}$ & \FA \\
    \hline    
    Im. & $0.76 \pm 0.1$ &  $0.15 \pm 0.05$ \\
    No Im. & $0.51 \pm 0.1$  & $0.17 \pm 0.04$ \\
    \hline
    p-val & $4.8$e-5 & 0.66
    \end{tabular}
\end{wraptable}
We compared \MA to flat architectures (\FA) that consider the whole scene at once. As the use of \FA for the reward function showed poor performance on $\G^\train$, Table~\ref{tab:archi_comparison} only compares the use of \MA and \FA for the policy. \MA shows stronger generalization and is the only architecture allowing an additional boost with goal imagination. Only \MA policy architectures can leverage the novel reward signals coming from imagined goals and turn them into \textit{behavioral adaptation}. Supplementary Section~\ref{sec:suppl_archi} provides additional details.


\subsection{Can We Use More Realistic Feedbacks?}
\label{sec:social_feedbacks}
\begin{wrapfigure}[16]{R}{0.425\textwidth}
\vspace{-.3cm}
     \includegraphics[width=0.425\textwidth]{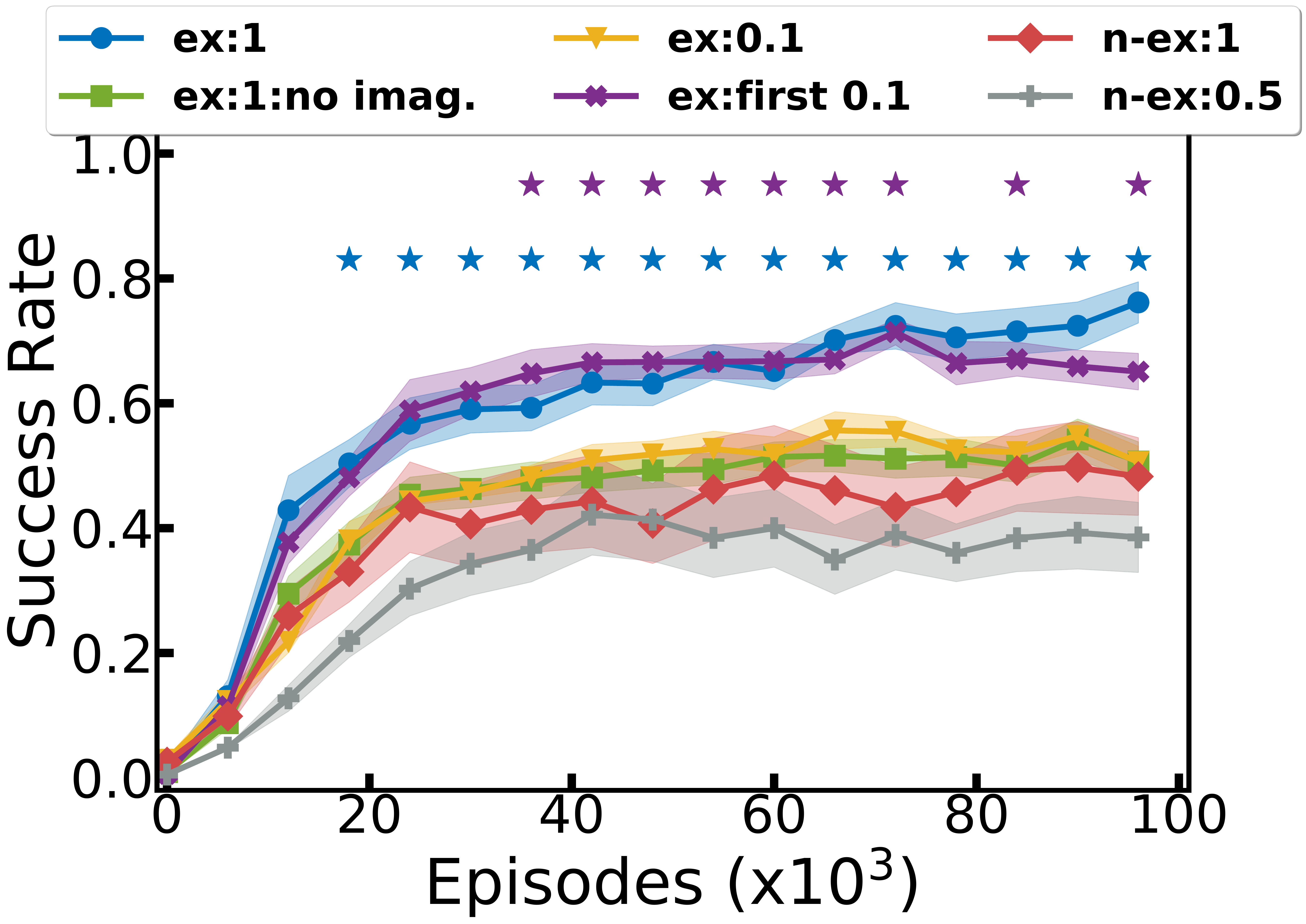}
     \vspace{-0.5cm}
        \caption{\textbf{Influence of social feedbacks.} $\SR$ on $\G^\text{test}$ for different social strategies. Stars indicate significant differences w.r.t. \textit{ex:1 no imag.}. sem plotted, 5 seeds.}
      \label{fig:social}
\end{wrapfigure} 
We study the relaxation of the \textit{full-presence} and \textit{exhaustiveness} assumptions of \SP. We first relax \textit{full-presence} while keeping \textit{exhaustiveness} (blue, yellow and purple curves). When \SP has a 10\% chance of being present (yellow), imaginative agents show generalization performance on par with the unimaginative agents trained in a full-presence setting (green), see Figure~\ref{fig:social}). However, when the same amount of feedback is concentrated in the first 10\% episodes (purple), goal imagination enables significant improvements in generalization (w.r.t. green).  This is reminiscent of children who require less and less attention as they grow into adulthood and is consistent with \citet{chan2019actrce}. Relaxing \textit{exhaustiveness}, \SP only provides one positive and one negative description every episode (red) or in 50\% of the episodes (gray). Then, generalization performance matches the one of unimaginative agents in the exhaustive setting (green). 

\section{Discussion and Conclusion}
\imagine is a learning architecture that enables autonomous learning by leveraging \NL interactions with a social partner. As other algorithms from the \imgep family, \imagine sets its own goals and builds behavioral repertoires without external rewards. As such, it is distinct from traditional instruction-following RL agents. This is done through the joint training of a language encoder for goal representation and a goal-achievement reward function to generate internal rewards. Our proposed modular architectures with gated-attention enable efficient out-of-distribution generalization of the reward function and policy. The ability to imagine new goals by composing known ones leads to further improvements over initial generalization abilities and fosters exploration beyond the set of interactions relevant to \SP. Our agent even tries to grow pieces of furniture with supplies, a behavior that can echo the way a child may try to feed his doll. 

\imagine does not need externally-provided rewards but learns which behaviors are \textit{interesting} from language-based interactions with \SP. In contrast with hand-crafted reward functions, \NL descriptions provide an easy way to guide machines towards relevant interactions. \textit{A posteriori} counterfactual feedback is easier to communicate for humans, especially when possible effects are unknown and, thus, the set of possible instructions is undefined. Hindsight learning also greatly benefits from such counterfactual feedback and improves sample efficiency.
Attention mechanisms further extend the interpretability of the agent's learning by mapping language to attentional scaling factors (see Supplementary Figure~\ref{fig:att}). In addition,  Section~\ref{sec:social_feedbacks} shows that agents can learn to achieve goals from a relatively small number of descriptions, paving the way towards human-provided descriptions. 

\textit{Playground} is a tool that we hope will enable the community to further study under-explored descriptive setups with rich combinatorial dynamics, as well as goal imagination. It is designed for the study of goal imagination and combinatorial generalization. Compared to existing environments \citep{Hermann2017,chevalier-boisvert2018babyai,chan2019actrce}, we allow the use of descriptive feedback, introduce the notion of object categories and category-dependent object interactions (\textit{Grow} refer to different modalities for \textit{plants} or \textit{animals}). Playground can easily be extended by adding objects, attributes, category- or object-type-dependent dynamics.

\imagine could be combined with unsupervised multi-object representation learning algorithms \cite{burgess2019monet, greff2019multi} to work directly from pixels, practically enforcing object-centered representations. The resulting algorithm would still be different from goal-as-state approaches \cite{nair2018visual,pong2019skew,nair2019contextual}. Supplementary Section~\ref{sec:suppl_discu} discusses the relevance of comparing \imagine to these works. Some tasks involve instruction-based navigation in visual environments that do not explictly represent objects \citep{nguyen2019vision,shridhar2020alfred}. Here, also, imagining new instructions from known ones could improve exploration and generalization. Finally, we believe \imagine could provide interesting extensions in hierarchical settings, like in \citet{Jiang2019}, with novel goal imagination boosting low-level exploration.

\paragraph{Future work.} A more complex language could be introduced, for example, by considering object relationships (e.g. \textit{Grasp any X left of Y}), see \cite{karch2020deep} for a preliminary experiment in this direction. While the use of pre-trained language models \cite{radford2019language} does not follow our developmental approach, it would be interesting to study how they would interact with goal imagination. Because \CGH performs well in our setup with a medium precision ($0.45$) and because similar mechanisms were successfully used for data augmentation in complex NLP tasks \cite{andreas2019goodenough}, we believe our goal imagination heuristic could scale to more realistic language. 

We could reduce the burden on \SP by considering unreliable feedbacks (lower precision), or by conditioning goal generation on the initial scene (e.g. using mechanisms from \citet{ther}). One could also add new interaction modalities by letting \SP make demonstrations, propose goals or guide the agent's attention. Our modular architectures, because they are set functions, could also directly be used to consider variable numbers of objects. Finally, we could use off-policy learning \cite{fujimoto2018off} to reinterpret past experience in the light of new imagined goals without any additional environment interactions.

\paragraph{Links.} Demonstration videos are available at \url{https://sites.google.com/view/imagine-drl}. The source code of playground environment can be found at \url{https://github.com/flowersteam/playground\_env} and the source code of the \imagine architecture \url{https://github.com/flowersteam/Imagine}.

\clearpage
\section*{Broader Impact Statement}

We present a reinforcement learning architecture where autonomous agents interact with a social partner to explore a large set of possible interactions and learn to master them. As a result, our work contributes to facilitating human intervention in the learning process of a robot, which we believe is a key step towards more explainable and safer autonomous robots. Besides, by releasing our code, we believe that we help efforts in reproducible science and allow the wider community to build upon and extend our work in the future. In that spirit, we also provide clear explanations on the number of seeds, error bars, and statistical testing when reporting the results.

\begin{ack}
C\'edric Colas and Tristan Karch are partly funded by the French Minist\`ere des Arm\'ees - Direction G\'en\'erale de l’Armement. Nicolas Lair is supported by ANRT/CIFRE contract No. 151575A20 from Cloud Temple.
\end{ack}

\bibliography{biblio}

\begin{thebibliography}{80}
\providecommand{\natexlab}[1]{#1}
\providecommand{\url}[1]{\texttt{#1}}
\expandafter\ifx\csname urlstyle\endcsname\relax
  \providecommand{\doi}[1]{doi: #1}\else
  \providecommand{\doi}{doi: \begingroup \urlstyle{rm}\Url}\fi

\bibitem[Andreas(2020)]{andreas2019goodenough}
Jacob Andreas.
\newblock Good-enough compositional data augmentation.
\newblock In \emph{Proceedings of the 58th Annual Meeting of the Association
  for Computational Linguistics}, pages 7556--7566, Online, July 2020.
  Association for Computational Linguistics.
\newblock \doi{10.18653/v1/2020.acl-main.676}.
\newblock URL \url{https://www.aclweb.org/anthology/2020.acl-main.676}.

\bibitem[Andrychowicz et~al.(2017)Andrychowicz, Wolski, Ray, Schneider, Fong,
  Welinder, McGrew, Tobin, Abbeel, and Zaremba]{andrychowicz2017hindsight}
Marcin Andrychowicz, Filip Wolski, Alex Ray, Jonas Schneider, Rachel Fong,
  Peter Welinder, Bob McGrew, Josh Tobin, OpenAI~Pieter Abbeel, and Wojciech
  Zaremba.
\newblock Hindsight experience replay.
\newblock In \emph{Advances in Neural Information Processing Systems}, pages
  5048--5058, 2017.

\bibitem[Asada et~al.(2009)Asada, Hosoda, Kuniyoshi, Ishiguro, Inui, Yoshikawa,
  Ogino, and Yoshida]{asada2009cognitive}
Minoru Asada, Koh Hosoda, Yasuo Kuniyoshi, Hiroshi Ishiguro, Toshio Inui,
  Yuichiro Yoshikawa, Masaki Ogino, and Chisato Yoshida.
\newblock Cognitive developmental robotics: A survey.
\newblock \emph{IEEE transactions on autonomous mental development}, 1\penalty0
  (1):\penalty0 12--34, 2009.

\bibitem[Bahdanau et~al.(2019{\natexlab{a}})Bahdanau, Hill, Leike, Hughes,
  Kohli, and Grefenstette]{bahdanau2018learning}
Dzmitry Bahdanau, Felix Hill, Jan Leike, Edward Hughes, Pushmeet Kohli, and
  Edward Grefenstette.
\newblock {Learning to Understand Goal Specifications by Modelling Reward}.
\newblock In \emph{International Conference on Learning Representations}, jun
  2019{\natexlab{a}}.

\bibitem[Bahdanau et~al.(2019{\natexlab{b}})Bahdanau, Murty, Noukhovitch,
  Nguyen, de~Vries, and Courville]{bahdanau2018systematic}
Dzmitry Bahdanau, Shikhar Murty, Michael Noukhovitch, Thien~Huu Nguyen, Harm
  de~Vries, and Aaron Courville.
\newblock Systematic generalization: What is required and can it be learned?
\newblock In \emph{ICLR}, 2019{\natexlab{b}}.

\bibitem[Baldassarre and Mirolli(2013)]{baldassarre2013intrinsically}
Gianluca Baldassarre and Marco Mirolli.
\newblock \emph{Intrinsically motivated learning in natural and artificial
  systems}.
\newblock Springer, 2013.

\bibitem[Baranes and Oudeyer(2013)]{baranes2013active}
Adrien Baranes and Pierre-Yves Oudeyer.
\newblock Active learning of inverse models with intrinsically motivated goal
  exploration in robots.
\newblock \emph{Robotics and Autonomous Systems}, 61\penalty0 (1):\penalty0
  49--73, 2013.

\bibitem[Bellemare et~al.(2016)Bellemare, Srinivasan, Ostrovski, Schaul,
  Saxton, and Munos]{bellemare2016unifying}
Marc Bellemare, Sriram Srinivasan, Georg Ostrovski, Tom Schaul, David Saxton,
  and Remi Munos.
\newblock Unifying count-based exploration and intrinsic motivation.
\newblock In \emph{Advances in neural information processing systems}, pages
  1471--1479, 2016.

\bibitem[Bornstein et~al.(1992)Bornstein, Tamis-LeMonda, Tal, Ludemann, Toda,
  Rahn, P{\^e}cheux, Azuma, and Vardi]{bornstein1992maternal}
Marc~H Bornstein, Catherine~S Tamis-LeMonda, Joseph Tal, Pamela Ludemann, Sueko
  Toda, Charles~W Rahn, Marie-Germaine P{\^e}cheux, Hiroshi Azuma, and Danya
  Vardi.
\newblock Maternal responsiveness to infants in three societies: The united
  states, france, and japan.
\newblock \emph{Child development}, 63\penalty0 (4):\penalty0 808--821, 1992.

\bibitem[Bruner(1991)]{Bruner1991}
Jerome Bruner.
\newblock {The Narrative Construction of Reality}.
\newblock \emph{Critical Inquiry}, 18\penalty0 (1):\penalty0 1--21, oct 1991.
\newblock ISSN 0093-1896.
\newblock \doi{10.1086/448619}.

\bibitem[Burgess et~al.(2019)Burgess, Matthey, Watters, Kabra, Higgins,
  Botvinick, and Lerchner]{burgess2019monet}
Christopher~P Burgess, Loic Matthey, Nicholas Watters, Rishabh Kabra, Irina
  Higgins, Matt Botvinick, and Alexander Lerchner.
\newblock Monet: Unsupervised scene decomposition and representation.
\newblock \emph{arXiv preprint arXiv:1901.11390}, 2019.

\bibitem[Cangelosi and Schlesinger(2015)]{cangelosi2015developmental}
Angelo Cangelosi and Matthew Schlesinger.
\newblock \emph{Developmental robotics: From babies to robots}.
\newblock MIT press, 2015.

\bibitem[Chan et~al.(2019)Chan, Wu, Kiros, Fidler, and Ba]{chan2019actrce}
Harris Chan, Yuhuai Wu, Jamie Kiros, Sanja Fidler, and Jimmy Ba.
\newblock Actrce: Augmenting experience via teacher's advice for multi-goal
  reinforcement learning, 2019.

\bibitem[Chaplot et~al.(2017)Chaplot, Sathyendra, Pasumarthi, Rajagopal, and
  Salakhutdinov]{chaplot2017gatedattention}
Devendra~Singh Chaplot, Kanthashree~Mysore Sathyendra, Rama~Kumar Pasumarthi,
  Dheeraj Rajagopal, and Ruslan Salakhutdinov.
\newblock Gated-attention architectures for task-oriented language grounding,
  2017.

\bibitem[Chen and Mooney(2011)]{Chen2011}
David~L. Chen and Raymond~J. Mooney.
\newblock {Learning to Interpret Natural Language Navigation Instructions from
  Observations}.
\newblock In \emph{AAAI Conference on Artificial Intelligence (AAAI), 2011},
  2011.

\bibitem[Chentanez et~al.(2005)Chentanez, Barto, and
  Singh]{chentanez2005intrinsically}
Nuttapong Chentanez, Andrew~G Barto, and Satinder~P Singh.
\newblock Intrinsically motivated reinforcement learning.
\newblock In \emph{Advances in neural information processing systems}, pages
  1281--1288, 2005.

\bibitem[Chevalier-Boisvert et~al.(2019)Chevalier-Boisvert, Bahdanau, Lahlou,
  Willems, Saharia, Nguyen, and Bengio]{chevalier-boisvert2018babyai}
Maxime Chevalier-Boisvert, Dzmitry Bahdanau, Salem Lahlou, Lucas Willems,
  Chitwan Saharia, Thien~Huu Nguyen, and Yoshua Bengio.
\newblock {Baby{\{}AI{\}}: First Steps Towards Grounded Language Learning With
  a Human In the Loop}.
\newblock In \emph{International Conference on Learning Representations}, 2019.

\bibitem[Chomsky(1957)]{Chomsky1957}
Noam. Chomsky.
\newblock \emph{{Syntactic structures}}.
\newblock Mouton, 1957.
\newblock ISBN 9789027933850.

\bibitem[Chu and Schulz(2020)]{chu2020exploratory}
Junyi Chu and Laura Schulz.
\newblock Exploratory play, rational action, and efficient search.
\newblock 2020.

\bibitem[Cideron et~al.(2019)Cideron, Seurin, Strub, and Pietquin]{ther}
Geoffrey Cideron, Mathieu Seurin, Florian Strub, and Olivier Pietquin.
\newblock Self-educated language agent with hindsight experience replay for
  instruction following.
\newblock \emph{arXiv preprint arXiv:1910.09451}, 2019.

\bibitem[Co-Reyes et~al.(2018)Co-Reyes, Gupta, Sanjeev, Altieri, Andreas,
  DeNero, Abbeel, and Levine]{coreyes2018guiding}
John~D. Co-Reyes, Abhishek Gupta, Suvansh Sanjeev, Nick Altieri, Jacob Andreas,
  John DeNero, Pieter Abbeel, and Sergey Levine.
\newblock Guiding policies with language via meta-learning, 2018.

\bibitem[Colas et~al.(2019{\natexlab{a}})Colas, Oudeyer, Sigaud, Fournier, and
  Chetouani]{curious}
C{\'{e}}dric Colas, Pierre{-}Yves Oudeyer, Olivier Sigaud, Pierre Fournier, and
  Mohamed Chetouani.
\newblock {CURIOUS:} intrinsically motivated modular multi-goal reinforcement
  learning.
\newblock In \emph{Proceedings of the 36th International Conference on Machine
  Learning, {ICML} 2019, 9-15 June 2019, Long Beach, California, {USA}}, pages
  1331--1340, 2019{\natexlab{a}}.

\bibitem[Colas et~al.(2019{\natexlab{b}})Colas, Sigaud, and
  Oudeyer]{colas2019hitchhiker}
C{\'e}dric Colas, Olivier Sigaud, and Pierre-Yves Oudeyer.
\newblock A hitchhiker's guide to statistical comparisons of reinforcement
  learning algorithms.
\newblock \emph{arXiv preprint arXiv:1904.06979}, 2019{\natexlab{b}}.

\bibitem[Dominey(2005)]{Dominey2005}
Peter~Ford Dominey.
\newblock {Emergence of grammatical constructions: evidence from simulation and
  grounded agent experiments}.
\newblock \emph{Connection Science}, 17\penalty0 (3-4):\penalty0 289--306, sep
  2005.
\newblock ISSN 0954-0091.
\newblock \doi{10.1080/09540090500270714}.

\bibitem[Ecoffet et~al.(2019)Ecoffet, Huizinga, Lehman, Stanley, and
  Clune]{ecoffet2019go}
Adrien Ecoffet, Joost Huizinga, Joel Lehman, Kenneth~O Stanley, and Jeff Clune.
\newblock Go-explore: a new approach for hard-exploration problems.
\newblock \emph{arXiv preprint arXiv:1901.10995}, 2019.

\bibitem[Florensa et~al.(2018)Florensa, Held, Geng, and
  Abbeel]{florensa2017automatic}
Carlos Florensa, David Held, Xinyang Geng, and Pieter Abbeel.
\newblock Automatic goal generation for reinforcement learning agents.
\newblock \emph{ICML}, 2018.

\bibitem[Forestier and Oudeyer(2016)]{forestier2016modular}
S{\'e}bastien Forestier and Pierre-Yves Oudeyer.
\newblock Modular active curiosity-driven discovery of tool use.
\newblock In \emph{Intelligent Robots and Systems (IROS), 2016 IEEE/RSJ
  International Conference on}, pages 3965--3972. IEEE, 2016.

\bibitem[Forestier et~al.(2017)Forestier, Mollard, and Oudeyer]{imgep}
S{\'{e}}bastien Forestier, Yoan Mollard, and Pierre{-}Yves Oudeyer.
\newblock Intrinsically motivated goal exploration processes with automatic
  curriculum learning.
\newblock \emph{CoRR}, abs/1708.02190, 2017.
\newblock URL \url{http://arxiv.org/abs/1708.02190}.

\bibitem[Fu et~al.(2019)Fu, Korattikara, Levine, and Guadarrama]{fu2018from}
Justin Fu, Anoop Korattikara, Sergey Levine, and Sergio Guadarrama.
\newblock {From Language to Goals: Inverse Reinforcement Learning for
  Vision-Based Instruction Following}.
\newblock In \emph{International Conference on Learning Representations}, 2019.

\bibitem[Fujimoto et~al.(2018)Fujimoto, Meger, and Precup]{fujimoto2018off}
Scott Fujimoto, David Meger, and Doina Precup.
\newblock Off-policy deep reinforcement learning without exploration.
\newblock \emph{arXiv preprint arXiv:1812.02900}, 2018.

\bibitem[Glenberg and Kaschak(2002)]{Glenberg2002}
Arthur~M. Glenberg and Michael~P. Kaschak.
\newblock {Grounding language in action}.
\newblock \emph{Psychonomic Bulletin {\&} Review}, 9\penalty0 (3):\penalty0
  558--565, sep 2002.
\newblock ISSN 1069-9384.
\newblock \doi{10.3758/BF03196313}.

\bibitem[Goldberg(2003)]{goldberg2003constructions}
Adele~E Goldberg.
\newblock Constructions: A new theoretical approach to language.
\newblock \emph{Trends in cognitive sciences}, 7\penalty0 (5):\penalty0
  219--224, 2003.

\bibitem[Gopnik et~al.(1999)Gopnik, Meltzoff, and Kuhl]{gopnik1999scientist}
Alison Gopnik, Andrew~N Meltzoff, and Patricia~K Kuhl.
\newblock \emph{The scientist in the crib: Minds, brains, and how children
  learn.}
\newblock William Morrow \& Co, 1999.

\bibitem[Goyal et~al.(2019)Goyal, Niekum, and Mooney]{Goyal2019}
Prasoon Goyal, Scott Niekum, and Raymond~J. Mooney.
\newblock {Using Natural Language for Reward Shaping in Reinforcement
  Learning}.
\newblock In \emph{IJCAI 2019}, mar 2019.
\newblock URL \url{http://arxiv.org/abs/1903.02020}.

\bibitem[Green and Quilty-Dunn(2017)]{green2017object}
Edwin~James Green and Jake Quilty-Dunn.
\newblock What is an object file?
\newblock \emph{The British Journal for the Philosophy of Science}, 2017.

\bibitem[Greff et~al.(2019)Greff, Kaufmann, Kabra, Watters, Burgess, Zoran,
  Matthey, Botvinick, and Lerchner]{greff2019multi}
Klaus Greff, Rapha{\"e}l~Lopez Kaufmann, Rishab Kabra, Nick Watters, Chris
  Burgess, Daniel Zoran, Loic Matthey, Matthew Botvinick, and Alexander
  Lerchner.
\newblock Multi-object representation learning with iterative variational
  inference.
\newblock \emph{arXiv preprint arXiv:1903.00450}, 2019.

\bibitem[He et~al.(2015)He, Zhang, Ren, and Sun]{he}
Kaiming He, Xiangyu Zhang, Shaoqing Ren, and Jian Sun.
\newblock Delving deep into rectifiers: Surpassing human-level performance on
  imagenet classification.
\newblock In \emph{Proceedings of the IEEE international conference on computer
  vision}, pages 1026--1034, 2015.

\bibitem[Hermann et~al.(2017)Hermann, Hill, Green, Wang, Faulkner, Soyer,
  Szepesvari, Czarnecki, Jaderberg, Teplyashin, Wainwright, Apps, Hassabis, and
  Blunsom]{Hermann2017}
Karl~Moritz Hermann, Felix Hill, Simon Green, Fumin Wang, Ryan Faulkner, Hubert
  Soyer, David Szepesvari, Wojciech~Marian Czarnecki, Max Jaderberg, Denis
  Teplyashin, Marcus Wainwright, Chris Apps, Demis Hassabis, and Phil Blunsom.
\newblock {Grounded Language Learning in a Simulated 3D World}.
\newblock jun 2017.
\newblock URL \url{http://arxiv.org/abs/1706.06551}.

\bibitem[Hill et~al.(2019)Hill, Lampinen, Schneider, Clark, Botvinick,
  McClelland, and Santoro]{hill2019emergent}
Felix Hill, Andrew Lampinen, Rosalia Schneider, Stephen Clark, Matthew
  Botvinick, James~L. McClelland, and Adam Santoro.
\newblock Emergent systematic generalization in a situated agent, 2019.

\bibitem[Hinaut and Dominey(2013)]{hinaut2013real}
Xavier Hinaut and Peter~Ford Dominey.
\newblock Real-time parallel processing of grammatical structure in the
  fronto-striatal system: A recurrent network simulation study using reservoir
  computing.
\newblock \emph{PloS one}, 8\penalty0 (2), 2013.

\bibitem[Hochreiter and Schmidhuber(1997)]{hochreiter1997lstm}
Sepp Hochreiter and J\"{u}rgen Schmidhuber.
\newblock Long short-term memory.
\newblock \emph{Neural Comput.}, 9\penalty0 (8):\penalty0 1735–1780, November
  1997.
\newblock ISSN 0899-7667.
\newblock \doi{10.1162/neco.1997.9.8.1735}.
\newblock URL \url{https://doi.org/10.1162/neco.1997.9.8.1735}.

\bibitem[Jiang et~al.(2019)Jiang, Gu, Murphy, and Finn]{Jiang2019}
Yiding Jiang, Shixiang Gu, Kevin Murphy, and Chelsea Finn.
\newblock {Language as an Abstraction for Hierarchical Deep Reinforcement
  Learning}.
\newblock In \emph{Workshop on “Structure {\&} Priors in Reinforcement
  Learning”at ICLR 2019}, jun 2019.
\newblock URL \url{http://arxiv.org/abs/1906.07343}.

\bibitem[Johnson et~al.(2003)Johnson, Amso, and
  Slemmer]{johnson2003development}
Scott~P Johnson, Dima Amso, and Jonathan~A Slemmer.
\newblock Development of object concepts in infancy: Evidence for early
  learning in an eye-tracking paradigm.
\newblock \emph{Proceedings of the National Academy of Sciences}, 100\penalty0
  (18):\penalty0 10568--10573, 2003.

\bibitem[Kaplan and Oudeyer(2007)]{kaplan2007search}
Frederic Kaplan and Pierre-Yves Oudeyer.
\newblock In search of the neural circuits of intrinsic motivation.
\newblock \emph{Frontiers in neuroscience}, 1:\penalty0 17, 2007.

\bibitem[Karch et~al.(2020)Karch, Colas, Teodorescu, Moulin-Frier, and
  Oudeyer]{karch2020deep}
Tristan Karch, Cédric Colas, Laetitia Teodorescu, Clément Moulin-Frier, and
  Pierre-Yves Oudeyer.
\newblock Deep sets for generalization in rl, 2020.

\bibitem[Keysers et~al.(2019)Keysers, Schärli, Scales, Buisman, Furrer,
  Kashubin, Momchev, Sinopalnikov, Stafiniak, Tihon, Tsarkov, Wang, van Zee,
  and Bousquet]{keysers2019measuring}
Daniel Keysers, Nathanael Schärli, Nathan Scales, Hylke Buisman, Daniel
  Furrer, Sergii Kashubin, Nikola Momchev, Danila Sinopalnikov, Lukasz
  Stafiniak, Tibor Tihon, Dmitry Tsarkov, Xiao Wang, Marc van Zee, and Olivier
  Bousquet.
\newblock Measuring compositional generalization: A comprehensive method on
  realistic data, 2019.

\bibitem[Kidd and Hayden(2015)]{kidd2015psychology}
Celeste Kidd and Benjamin~Y Hayden.
\newblock The psychology and neuroscience of curiosity.
\newblock \emph{Neuron}, 88\penalty0 (3):\penalty0 449--460, 2015.

\bibitem[Kingma and Ba(2014)]{kingma2014adam}
Diederik~P Kingma and Jimmy Ba.
\newblock Adam: A method for stochastic optimization.
\newblock \emph{arXiv preprint arXiv:1412.6980}, 2014.

\bibitem[Laversanne-Finot et~al.(2018)Laversanne-Finot, Pere, and
  Oudeyer]{laversanne2018curiosity}
Adrien Laversanne-Finot, Alexandre Pere, and Pierre-Yves Oudeyer.
\newblock Curiosity driven exploration of learned disentangled goal spaces.
\newblock volume~87 of \emph{Proceedings of Machine Learning Research}, pages
  487--504. PMLR, 29--31 Oct 2018.
\newblock URL \url{http://proceedings.mlr.press/v87/laversanne-finot18a.html}.

\bibitem[Lillicrap et~al.(2015)Lillicrap, Hunt, Pritzel, Heess, Erez, Tassa,
  Silver, and Wierstra]{lillicrap2015continuous}
Timothy~P Lillicrap, Jonathan~J Hunt, Alexander Pritzel, Nicolas Heess, Tom
  Erez, Yuval Tassa, David Silver, and Daan Wierstra.
\newblock Continuous control with deep reinforcement learning.
\newblock \emph{arXiv preprint arXiv:1509.02971}, 2015.

\bibitem[Luketina et~al.(2019)Luketina, Nardelli, Farquhar, Foerster, Andreas,
  Grefenstette, Whiteson, and Rockt{\"{a}}schel]{Luketina2019}
Jelena Luketina, Nantas Nardelli, Gregory Farquhar, Jakob Foerster, Jacob
  Andreas, Edward Grefenstette, Shimon Whiteson, and Tim Rockt{\"{a}}schel.
\newblock {A Survey of Reinforcement Learning Informed by Natural Language}.
\newblock \emph{IJCAI'19}, jun 2019.
\newblock URL \url{http://arxiv.org/abs/1906.03926}.

\bibitem[Madden et~al.(2010)Madden, Hoen, and Dominey]{Madden2010}
Carol Madden, Michel Hoen, and Peter~Ford Dominey.
\newblock {A cognitive neuroscience perspective on embodied language for
  human–robot cooperation}.
\newblock \emph{Brain and Language}, 112\penalty0 (3):\penalty0 180--188, mar
  2010.
\newblock ISSN 0093-934X.
\newblock \doi{10.1016/J.BANDL.2009.07.001}.

\bibitem[Mankowitz et~al.(2018)Mankowitz, Z{\'{\i}}dek, Barreto, Horgan,
  Hessel, Quan, Oh, van Hasselt, Silver, and Schaul]{unicorn}
Daniel~J. Mankowitz, Augustin Z{\'{\i}}dek, Andr{\'{e}} Barreto, Dan Horgan,
  Matteo Hessel, John Quan, Junhyuk Oh, Hado van Hasselt, David Silver, and Tom
  Schaul.
\newblock Unicorn: Continual learning with a universal, off-policy agent.
\newblock \emph{CoRR}, abs/1802.08294, 2018.
\newblock URL \url{http://arxiv.org/abs/1802.08294}.

\bibitem[McClelland et~al.(2019)McClelland, Hill, Rudolph, Baldridge, and
  Sch{\"u}tze]{mcclelland2019extending}
James~L McClelland, Felix Hill, Maja Rudolph, Jason Baldridge, and Hinrich
  Sch{\"u}tze.
\newblock Extending machine language models toward human-level language
  understanding.
\newblock \emph{arXiv preprint arXiv:1912.05877}, 2019.

\bibitem[Mintz(2003)]{mintz2003frequent}
Toben~H Mintz.
\newblock Frequent frames as a cue for grammatical categories in child directed
  speech.
\newblock \emph{Cognition}, 90\penalty0 (1):\penalty0 91--117, 2003.

\bibitem[Nair et~al.(2019)Nair, Bahl, Khazatsky, Pong, Berseth, and
  Levine]{nair2019contextual}
Ashvin Nair, Shikhar Bahl, Alexander Khazatsky, Vitchyr Pong, Glen Berseth, and
  Sergey Levine.
\newblock Contextual imagined goals for self-supervised robotic learning.
\newblock \emph{arXiv preprint arXiv:1910.11670}, 2019.

\bibitem[Nair et~al.(2018)Nair, Pong, Dalal, Bahl, Lin, and
  Levine]{nair2018visual}
Ashvin~V Nair, Vitchyr Pong, Murtaza Dalal, Shikhar Bahl, Steven Lin, and
  Sergey Levine.
\newblock Visual reinforcement learning with imagined goals.
\newblock In \emph{Advances in Neural Information Processing Systems}, pages
  9191--9200, 2018.

\bibitem[Nguyen et~al.(2019)Nguyen, Dey, Brockett, and Dolan]{nguyen2019vision}
Khanh Nguyen, Debadeepta Dey, Chris Brockett, and Bill Dolan.
\newblock Vision-based navigation with language-based assistance via imitation
  learning with indirect intervention.
\newblock In \emph{Proceedings of the IEEE Conference on Computer Vision and
  Pattern Recognition}, pages 12527--12537, 2019.

\bibitem[Oudeyer et~al.(2007)Oudeyer, Kaplan, and Hafner]{oudeyer2007intrinsic}
Pierre-Yves Oudeyer, Frdric Kaplan, and Verena~V Hafner.
\newblock Intrinsic motivation systems for autonomous mental development.
\newblock \emph{IEEE transactions on evolutionary computation}, 11\penalty0
  (2):\penalty0 265--286, 2007.

\bibitem[Pathak et~al.(2017)Pathak, Agrawal, Efros, and
  Darrell]{pathak2017curiosity}
Deepak Pathak, Pulkit Agrawal, Alexei~A Efros, and Trevor Darrell.
\newblock Curiosity-driven exploration by self-supervised prediction.
\newblock In \emph{Proceedings of the IEEE Conference on Computer Vision and
  Pattern Recognition Workshops}, pages 16--17, 2017.

\bibitem[Piaget(1926)]{Piaget1926}
Jean Piaget.
\newblock \emph{{The language and thought of the child}}.
\newblock Routledge, 1926.
\newblock ISBN 0415267501.

\bibitem[Plappert et~al.(2018)Plappert, Andrychowicz, Ray, McGrew, Baker,
  Powell, Schneider, Tobin, Chociej, Welinder, et~al.]{plappert2018multi}
Matthias Plappert, Marcin Andrychowicz, Alex Ray, Bob McGrew, Bowen Baker,
  Glenn Powell, Jonas Schneider, Josh Tobin, Maciek Chociej, Peter Welinder,
  et~al.
\newblock Multi-goal reinforcement learning: Challenging robotics environments
  and request for research.
\newblock \emph{arXiv preprint arXiv:1802.09464}, 2018.

\bibitem[Pong et~al.(2019)Pong, Dalal, Lin, Nair, Bahl, and
  Levine]{pong2019skew}
Vitchyr~H Pong, Murtaza Dalal, Steven Lin, Ashvin Nair, Shikhar Bahl, and
  Sergey Levine.
\newblock Skew-fit: State-covering self-supervised reinforcement learning.
\newblock \emph{arXiv preprint arXiv:1903.03698}, 2019.

\bibitem[{R. K. Branavan} et~al.(2010){R. K. Branavan}, {S. Zettlemoyer}, and
  Barzilay]{Branavan2010}
S~{R. K. Branavan}, Luke {S. Zettlemoyer}, and Regina Barzilay.
\newblock {Reading Between the Lines: Learning to Map High-level Instructions
  to Commands}.
\newblock In \emph{ACL 2010 - 48th Annual Meeting of the Association for
  Computational Linguistics, Proceedings of the Conference}, pages 1268--1277,
  2010.

\bibitem[Racaniere et~al.(2019)Racaniere, Lampinen, Santoro, Reichert, Firoiu,
  and Lillicrap]{racaniere2019automated}
Sebastien Racaniere, Andrew~K Lampinen, Adam Santoro, David~P Reichert, Vlad
  Firoiu, and Timothy~P Lillicrap.
\newblock Automated curricula through setter-solver interactions.
\newblock \emph{arXiv preprint arXiv:1909.12892}, 2019.

\bibitem[Radford et~al.(2019)Radford, Wu, Child, Luan, Amodei, and
  Sutskever]{radford2019language}
Alec Radford, Jeffrey Wu, Rewon Child, David Luan, Dario Amodei, and Ilya
  Sutskever.
\newblock Language models are unsupervised multitask learners.
\newblock \emph{OpenAI Blog}, 1\penalty0 (8):\penalty0 9, 2019.

\bibitem[Schaul et~al.(2015)Schaul, Horgan, Gregor, and
  Silver]{schaul2015universal}
Tom Schaul, Daniel Horgan, Karol Gregor, and David Silver.
\newblock Universal value function approximators.
\newblock In \emph{International Conference on Machine Learning}, pages
  1312--1320, 2015.

\bibitem[Schmidhuber(2010)]{schmidhuber2010formal}
J{\"u}rgen Schmidhuber.
\newblock Formal theory of creativity, fun, and intrinsic motivation
  (1990--2010).
\newblock \emph{IEEE Transactions on Autonomous Mental Development}, 2\penalty0
  (3):\penalty0 230--247, 2010.

\bibitem[Shridhar et~al.(2020)Shridhar, Thomason, Gordon, Bisk, Han, Mottaghi,
  Zettlemoyer, and Fox]{shridhar2020alfred}
Mohit Shridhar, Jesse Thomason, Daniel Gordon, Yonatan Bisk, Winson Han,
  Roozbeh Mottaghi, Luke Zettlemoyer, and Dieter Fox.
\newblock Alfred: A benchmark for interpreting grounded instructions for
  everyday tasks.
\newblock In \emph{Proceedings of the IEEE/CVF Conference on Computer Vision
  and Pattern Recognition}, pages 10740--10749, 2020.

\bibitem[Spelke et~al.(1992)Spelke, Breinlinger, Macomber, and
  Jacobson]{spelke1992origins}
Elizabeth~S Spelke, Karen Breinlinger, Janet Macomber, and Kristen Jacobson.
\newblock Origins of knowledge.
\newblock \emph{Psychological review}, 99\penalty0 (4):\penalty0 605, 1992.

\bibitem[Steels(2006)]{steels2006semiotic}
Luc Steels.
\newblock Semiotic dynamics for embodied agents.
\newblock \emph{IEEE Intelligent Systems}, 21\penalty0 (3):\penalty0 32--38,
  2006.

\bibitem[Tomasello(1999)]{Tomasello1999}
Michael Tomasello.
\newblock \emph{{The cultural origins of human cognition}}.
\newblock Harvard University Press, 1999.
\newblock ISBN 9780674005822.

\bibitem[Tomasello(2000)]{tomasello2000item}
Michael Tomasello.
\newblock The item-based nature of children’s early syntactic development.
\newblock \emph{Trends in cognitive sciences}, 4\penalty0 (4):\penalty0
  156--163, 2000.

\bibitem[Tomasello(2009)]{tomasello2009constructing}
Michael Tomasello.
\newblock \emph{Constructing a language}.
\newblock Harvard university press, 2009.

\bibitem[Tomasello and Olguin(1993)]{tomasello1993twenty}
Michael Tomasello and Raquel Olguin.
\newblock Twenty-three-month-old children have a grammatical category of noun.
\newblock \emph{Cognitive development}, 8\penalty0 (4):\penalty0 451--464,
  1993.

\bibitem[Venkattaramanujam et~al.(2019)Venkattaramanujam, Crawford, Doan, and
  Precup]{venkattaramanujam2019self}
Srinivas Venkattaramanujam, Eric Crawford, Thang Doan, and Doina Precup.
\newblock Self-supervised learning of distance functions for goal-conditioned
  reinforcement learning.
\newblock \emph{arXiv preprint arXiv:1907.02998}, 2019.

\bibitem[Vygotsky(1978)]{Vygotskii1978}
L.~S. Vygotsky.
\newblock {Tool and Symbol in Child Development}.
\newblock In \emph{Mind in Society}, chapter Tool and Symbol in Child
  Development, pages 19--30. Harvard University Press, 1978.
\newblock ISBN 0674576292.
\newblock \doi{10.2307/j.ctvjf9vz4.6}.

\bibitem[Zaheer et~al.(2017)Zaheer, Kottur, Ravanbakhsh, Poczos, Salakhutdinov,
  and Smola]{deepset}
Manzil Zaheer, Satwik Kottur, Siamak Ravanbakhsh, Barnabas Poczos, Russ~R
  Salakhutdinov, and Alexander~J Smola.
\newblock Deep sets.
\newblock In \emph{Advances in neural information processing systems}, pages
  3391--3401, 2017.

\bibitem[Zaremba et~al.(2014)Zaremba, Sutskever, and
  Vinyals]{zaremba2014recurrent}
Wojciech Zaremba, Ilya Sutskever, and Oriol Vinyals.
\newblock Recurrent neural network regularization, 2014.

\bibitem[Zwaan and Madden(2005)]{Zwaan05}
Rolf Zwaan and Carol Madden.
\newblock {Embodied sentence comprehension}.
\newblock \emph{Grounding Cognition: The Role of Perception and Action in
  Memory, Language, and Thinking}, pages 224--245, 2005.
\newblock \doi{10.1017/CBO9780511499968.010}.

\end{thebibliography}
\bibliographystyle{plainnat}

\clearpage

\section*{Supplementary Material}
This supplementary material provides additional methods, results and discussion, as well as implementation details.

\begin{itemize}[nolistsep]
    \item Section~\ref{sec:suppl_env_descr} gives a complete description of our setup and of the \textit{Playground} environment.
    \item Section~\ref{sec:supp_focus_gene} presents a focus on generalization and studies different types of generalization.
    \item Section~\ref{sec:suppl_exploration} presents a focus on exploration and how it is influenced by goal imagination.
    \item Section~\ref{sec:suppl_goal_imagination} presents a focus on the goal imagination mechanism we use for \imagine.
    \item Section~\ref{sec:suppl_archi} presents a focus on the \textit{Modular-Attention} architecture.
    \item Section~\ref{sec:suppl_reward} presents a focus on the benefits of learning the reward function.
    \item Section~\ref{sec:suppl_visu} provides additional visualization of the goal embeddings and the attention vectors.
    \item Section~\ref{sec:suppl_discu} discusses the comparison with goal-as-state approaches.
    \item Section~\ref{sec:supp_impl_details} gives all necessary implementation details.
\end{itemize}





\section{Complete Description of the Playground Environment and Its Language}
\label{sec:suppl_env_descr}

\begin{figure}[ht]
    \centering
        \includegraphics[width=0.9\columnwidth]{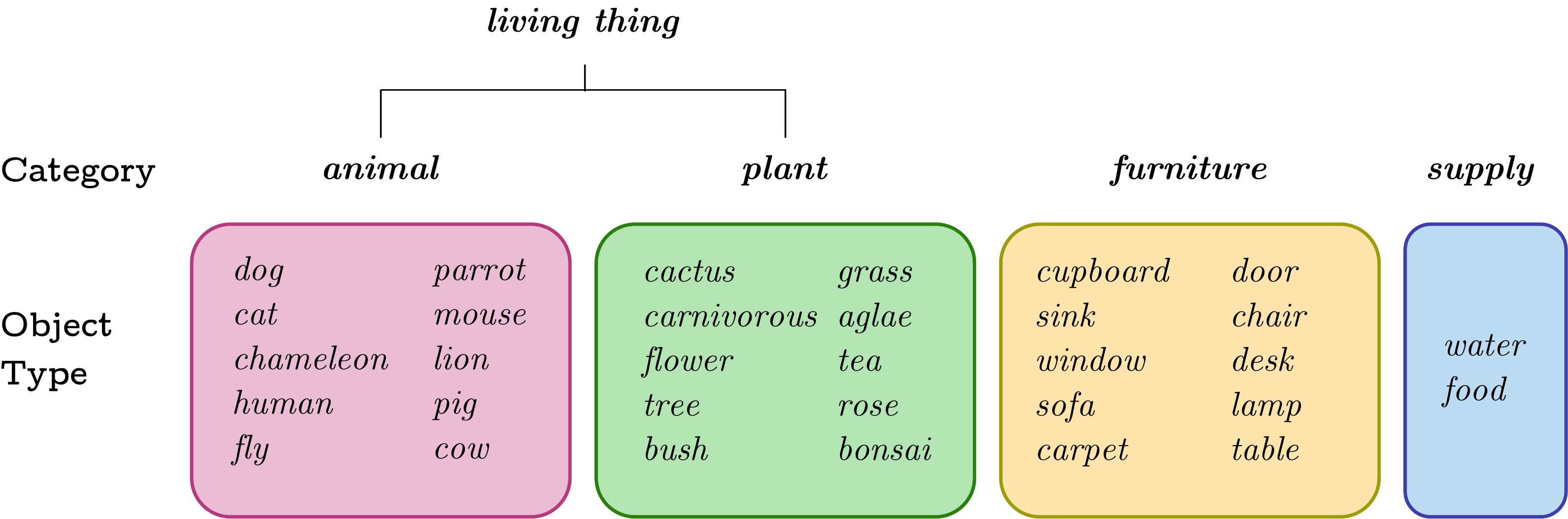}
         \caption{Representation of possible objects types and categories.}  
    \label{fig:env_category}
    \end{figure}

\paragraph{Environment description.} The environment is a $2$D square: $[-1.2,1.2]^2$. The agent is a disc of diameter $0.05$ with an initial position $(0,0)$. Objects have sizes uniformly sampled from $[0.2, 0.3]$ and their initial positions are randomized so that they are not in contact with each other. The agent has an action space of size $3$ bounded in $[-1,1]$. The first two actions control the agent's continuous $2$D translation (bounded to $0.15$ in any direction). The agent can grasp objects by getting in contact with them and closing its gripper (positive third action), unless it already has an object in hand. Objects include $10$ animals, $10$ plants, $10$ pieces of furniture and $2$ supplies. Admissible categories are \textit{animal, plant, furniture, supply} and \textit{living\_thing} (animal or plant), see Figure~\ref{fig:env_category}. Objects are assigned a color attribute (red, blue or green). Their precise color is a continuous RGB code uniformly sampled from  RGB subspaces associated with their attribute color. Each scene contains $3$ of these procedurally-generated objects (see paragraph about the Social Partner below).

\paragraph{Agent perception.} At time step $t$, we can define an observation $\textbf{o}_t$ as the concatenation of body observations ($2$D-position, gripper state) and objects' features. These two types of features form affordances between the agent and the objects around. These affordances are necessary to understand the meaning of object interactions like \textit{grasp}. The state $\textbf{s}_t$ used as input of the models is the concatenation of $\textbf{o}_t$ and $\bDelta\textbf{o}_t = \textbf{o}_t-\textbf{o}_0$ to provide a sense of time. This is required to acquire the understanding and behavior related to the \textit{grow} predicate, as the agent needs to observe and produce a change in the object's size.

\paragraph{Social Partner.} \SP has two roles:
\begin{itemize}
    \item \textit{Scene organization}: \SP organize the scene according to the goal selected by the agent. When the agent selects a goal, it communicates it to \SP. If the goal starts by the word \textit{grow}, \SP adds a procedurally-generated supply (water or food for animals, water for plants) of any size and color to the scene. If the goal contains an object (e.g. \textit{red cat}), \SP adds a corresponding object to the scene (with a procedurally generated size and RGB color). Remaining objects are generated procedurally. As a result, the objects required to fulfill a goal are always present and the scene contains between 1 (\textit{grow} goals) and 3 (\textit{go} goals) random objects. Note that all objects are procedurally generated (random initial position, RGB color and size).
    \item \textit{Scene description}: \SP provides \NL descriptions of interesting outcomes experienced by the agent at the end of episodes. It takes the final state of an episode ($\mathbf{s}_T$) as input and returns matching \NL descriptions: $\mathcal{D}_{\text{\SP}}(\mathbf{s}_T) \subset \mathcal{D}^\text{\SP}$. When \SP provides \textit{descriptions}, the agent considers them as targetable \textit{goals}. This mapping $\mathcal{D}^\text{\SP}\to \G^\train$ simply consists in removing the first \textit{you} token (e.g. turning \textit{you grasp red door} into the goal \textit{grasp red door}). Given the set of previously discovered goals $(\G_\text{known})$ and new descriptions $\mathcal{D}_{\text{\SP}}(\mathbf{s}_T)$, the agent infers the set of goals that were not achieved: $\G_\text{na}(\mathbf{s}_T)~=~\G_\text{known}~\backslash~\mathcal{D}_\text{\SP}(\mathbf{s}_T)$, where $\backslash$ indicates the complement.
\end{itemize}

\paragraph{Grammar.} We now present the grammar that generates descriptions for the set of goals achievable in the Playground environment $(\G^{A})$. \textbf{Bold} and \{ \} refer to sets of words while \textit{italics} refers to particular words:

\begin{enumerate}[leftmargin=0.6cm, nolistsep]
    \item Go: \textit{(e.g. go bottom left)} 
    \begin{itemize}[leftmargin=0.2cm,noitemsep]
            \itemsep-0.3em 
            \item \textit{go} + \bi{zone}
            \end{itemize}
    \item Grasp: \textit{(e.g. grasp any animal)} 
        \begin{itemize}[leftmargin=0.2cm,noitemsep]
            \itemsep-0.3em 
            \item \textit{grasp} + \bi{color} $\cup$ \{\textit{any}\}  + \textbf{\textit{object type $\cup$ object category}}
            \item \textit{grasp} + \textit{any} + \bi{color} + \textit{thing}
        \end{itemize}
    \item Grow: \textit{(e.g. grow blue lion)}
        \begin{itemize}[leftmargin=0.2cm,noitemsep]
            \itemsep-0.3em 
            \item \textit{grow} + \bi{color} $\cup$ \{\textit{any}\} + \bi{living thing} $\cup$ \{\textit{living\_thing, animal, plant}\}
            \item \textit{grow} + \textit{any} + \bi{color} + \textit{thing}
    \end{itemize}
\end{enumerate} 

Word sets are defined by:
\begin{itemize}[noitemsep]
    \item \textbf{\textit{zone}} = \{\textit{center, top, bottom, right, left, top left, top right, bottom left, bottom right}\}
    \item \textbf{\textit{color}} = \{\textit{red, blue, green}\} 
    \item  \textbf{\textit{object type}} = \textbf{\textit{living thing}}  $\cup$ \textbf{\textit{furniture}}  $\cup$ \textbf{\textit{supply}}
    \item  \textbf{\textit{object category}} = \{\textit{living\_thing}, \textit{animal}, \textit{plant}, \textit{furniture}, \textit{supply}\}
     \item  \textbf{\textit{living thing}} = \textbf{\textit{animal}} $\cup$ \textbf{\textit{plant}}
    \item \textbf{\textit{animal}} = \{\textit{dog, cat, chameleon, human, fly, parrot, mouse, lion, pig, cow}\}
    \item \textbf{\textit{plant}} = \{\textit{cactus, carnivorous, flower, tree, bush, grass, algae, tea, rose, bonsai}\}
    \item \textbf{\textit{furniture}} = \{\textit{door, chair, desk, lamp, table, cupboard, sink, window, sofa, carpet}\} 
    \item \textbf{\textit{supply}} = \{\textit{water, food}\}
    \item \textbf{\textit{predicate}} = \{\textit{go, grasp, grow}\}
\end{itemize}
We partition this set of achievable goals into a training $(\G^\train)$ and a testing $(\G^\test)$ set. Goals from $\G^\test$ are intended to evaluate the ability of our agent to explore the set of achievable outcomes beyond the set of outcomes described by \SP. The next section introduces this testing set and focuses on generalization. Note that some goals might be syntactically valid but not achievable. This includes all goals of the form \textit{grow} + \bi{color} $\cup$ \{\textit{any}\} + \bi{furniture} $\cup$ \{\textit{furniture}\} (e.g. \textit{grow red lamp}).

\paragraph{IMAGINE Pseudo-Code.}
Algorithm~\ref{alg:example} outlines the pseudo-code of our learning architecture. See Main Section~\ref{sec:architecture} for high-level descriptions of each module and function.

 \begin{algorithm}[h!]
    \caption{\imagine}
    \label{alg:example}

    \begin{algorithmic}[1]
       \STATE \textbf{Input:} env, \SP \hspace{1.cm}
       \STATE \textbf{Initialize:} $L_e$, $\mathcal{R}$, $\pi$, $mem(\mathcal{R})$, $mem(\pi)$, $\G_\text{known}$, $\G_\text{im}$ \\  \hspace{1cm} \# Random initializations for networks  \\
       \hspace{1cm} \# empty sets for memories and goal sets
       
      \FOR{$e=1:N_{episodes}$}
         \IF{$\G_\text{known} \neq \text{\O}$}
            \STATE sample $g_\text{NL}$ from $\G_\text{known} \cup \G_\text{im}$ 
            \STATE $g \gets L_e(g_\text{NL})$
         \ELSE
            \STATE sample $g$ from $\mathcal{N}(0,\mathbf{I})$
         \ENDIF
         \STATE $s_0 \gets$ env.reset()
         \FOR{$t=1:T$}
            \STATE $a_t\gets\pi(s_{t-1},g)$
            \STATE $s_{t}\gets$ env.step($a_t$)
            \STATE $mem_{\pi}$.add($s_{t-1}, a_t, s_t$)
         \ENDFOR
         \STATE $\G_\text{\SP} \gets$ \SP.get\_descriptions($s_T$)
         \STATE $\G_\text{known} \gets \G_\text{known} \: \cup$ $\G_\text{\SP}$
         \STATE $mem(\mathcal{R})$.add($s_T$, $g_\text{NL})$ for $g_\text{NL}$ in $\G_\text{\SP}$
         \IF{goal imagination allowed}
            \STATE $\G_\text{im} \gets$ \textbf{Imagination}$(\G_\text{known})$ \# see Algorithm~\ref{alg:suppl_imagination}
         \ENDIF
         \STATE Batch$_{\pi} \gets$ \textbf{ModularBatchGenerator}$(mem(\pi))$ 
         \hspace{1cm} \# Batch$_{\pi}$=$\{(s,a,s')\}$
         \STATE Batch$_{\pi} \gets$ \textbf{Hindsight}$(\text{Batch}_\pi, \mathcal{R}, \G_\text{known}, \G_\text{im})$ 
         \hspace{0.2cm} \# Batch$_{\pi}$=$\{(s,a,r,g,s')\}$ where $r=\mathcal{R}(s,g)$
         \STATE $\pi \gets $\textbf{RL\_Update}(Batch$_{\pi}$)
      \IF{$e \:\%\: $reward\_update\_freq $==0$}
      \STATE Batch$_{\mathcal{R}} \gets$ ModularBatchGenerator$(mem(\mathcal{R}))$
      \STATE $L_e$, $\mathcal{R} \gets$ \textbf{LE\&RewardFunctionUpdate}$(\text{Batch}_{\mathcal{R}})$
      \ENDIF
      \ENDFOR 
    \end{algorithmic}
\end{algorithm}


\clearpage

\section{Focus on Generalization}
\label{sec:supp_focus_gene}
Because scenes are procedurally-generated, $\SR$ computed on $\G^\train$ measures the generalization to new states. When computed on $\G^\test$, however, $\SR$ measures both this state generalization and the generalization to new goal descriptions from $\G^\test$. As $\SR_\train$ is almost perfect, this section focuses solely on generalization in the language space: $\SR_\test$.

\paragraph{Different types of generalization.}
Generalization can occur in two different modules of the \imagine architecture: in the reward function and in the policy. Agents can only benefit from goal imagination when their reward function is able to generalize the meanings of imagined goals from the meanings of known ones. When they do, they can further train on imagined goals, which might, in turn, reinforce the generalization of the policy. This section characterizes different types of generalizations that the reward and policy can both demonstrate.

\begin{itemize}
    \item Type 1 - \textit{Attribute-object generalization}: This is the ability to accurately associate an attribute and an object that were never seen together before. To interpret the goal \textit{grasp red tree} requires to isolate the \textit{red} and \textit{tree} concepts from other sentences and to combine them to recognize a \textit{red tree}. To measure this ability, we removed from the training set all goals containing the following attribute-object combinations: \textit{\{blue door, red tree, green dog\}} and added them to the testing set (4 goals). 
    
    \item Type 2 - \textit{Object identification}: This is the ability to identify a new object from its attribute. We left out of the training set all goals containing the word \textit{flower} (4 goals). To interpret the goal \textit{grasp red flower} requires to isolate the concept of \textit{red} and to transpose it to the unknown object \textit{flower}. Note that in the case of \textit{grasp any flower}, the agent cannot rely on the attribute, and must perform some kind of complement reasoning:``if these are known objects, and that is unknown, then if must be a \textit{flower}".
    
    \item Type 3 - \textit{Predicate-category generalization}: 
    This is the ability to interpret a predicate for a category when they were never seen together before. As explained in Section~\ref{sec:suppl_env_descr}, a category regroups a set of objects and is not encoded in the object state vector. It is only a linguistic concept. We left out all goals with the \textit{grasp} predicate and the \textit{animal} category (4 goals). To correctly interpret \textit{grasp any animal} requires to identify objects that belong to the animal category (acquired from "growing \textit{animal}" and "growing animal objects" goals), to isolate the concept of \textit{grasping} (acquired from grasping non-\textit{animal} objects) and to combine the two.
    
    \item Type 4 - \textit{Predicate-object generalization}:  This is the ability to interpret a predicate for an object when they were never seen together before. We leave out all goals with the \textit{grasp} predicate and the \textit{fly} object (4 goals). To correctly interpret \textit{grasp any fly}, the agent should  leverage its knowledge about the \textit{grasp} predicate (acquired from the "grasping non-fly objects" goals) and the \textit{fly} object (acquired from the "growing flies" goals).
    
   \item Type 5 - \textit{Predicate dynamics generalization}: This is the ability to generalize the behavior associated with a predicate to another category of objects, for which the dynamics is changed. In the \textit{Playground} environment, the dynamics of \textit{grow} with \bi{animals} and \bi{plants} is a a bit different. \bi{animals} can be grown with \textit{food} and \textit{water} whereas \bi{plants} only grow with \textit{water}. We want to see if \imagine can learn the dynamics of \textit{grow} on \bi{animals} and generalize it to \bi{plants}. We left out all goals with the \textit{grow} predicate and any of the \bi{plant} objects, \textit{plant} and \textit{living thing} categories (48 goals). To interpret, \textit{grow any plant}, the agent should be able to identify the \bi{plant} objects (acquired from the "grasping plants" goals) and that objects need supplies (food or water) to \textit{grow} (acquired from the "growing animals" goals). Type 5 is more complex than Type 4 for two reasons: 1) because the dynamics change and 2) because it mixes objects and categories. Note that, by definition, the zero-shot generalization is tested without additional reward signals (before imagination). As a result, even the best zero-shot generalization possible cannot adapt the \textit{grow} behavior from animals to plant and would bring food and water with equal probability $p=0.5$ for each.
\end{itemize}

 Table~\ref{tab:test_descriptions} provides the exhaustive list of goals used to test each type of generalization.


\paragraph{Different ways to generalize.}
Agent can generalize to out-of-distribution goals (from any of the 5 categories above) in three different ways:
\begin{enumerate}
    \item \textit{Policy zero-shot generalization}: The policy can achieve the new goal without any supplementary training.
    \item \textit{Reward zero-shot generalization}: The reward can tell whether the goal is achieved or not without any supplementary training.
    \item \textit{Policy n-shot generalization or behavioral adaptation}:
    When allowed to imagine goals, \imagine agents can use the zero-shot generalization of their reward function to autonomously train their policy to improve on imagined goals. After such training, the policy might show improved generalization performance compared to its zero-shot abilities. We call this performance \textit{n-shot generalization}. The policy received supplementary training, but did not leverage any external supervision, only  the zero-shot generalization of its internal reward function. This is crucial to achieve Type 5 generalization. As we said, zero-shot generalization cannot figure out that plants only grow with water. Fine-tuning the policy based on experience and internal rewards enables agents to perform \textit{behavioral adaptation}: adapting their behavior with respect to imagined goals in an autonomous manner (see Main Figure~\ref{fig:adaptation}).
\end{enumerate}

\begin{table}[!h]
    \centering
    \caption{Testing goals in $\G^\test$, by type.}
    \label{tab:test_descriptions}
    \begin{tabular}{c|l}
        Type 1 & \textit{Grasp blue door}, \textit{Grasp green dog},\textit{Grasp red tree}, \textit{Grow green dog} \\ 
        \hline
        \multirow{2}{2.75em}{Type 2} & \textit{Grasp any flower}, \textit{Grasp blue flower}, \textit{Grasp green flower}, \textit{Grasp red flower}, \\
                                     & \textit{Grow any flower}, \textit{Grow blue flower}, \textit{Grow green flower}, \textit{Grow red flower}\\
        \hline
        Type 3 & \textit{Grasp any animal}, \textit{Grasp blue animal}, \textit{Grasp green animal}, \textit{Grasp red animal} \\ 
        \hline
       Type 4 & \textit{Grasp any fly}, \textit{Grasp blue fly}, \textit{Grasp green fly}, \textit{Grasp red fly} \\ 
        \hline
        \multirow{11}{2.75em}{Type 5} & \textit{Grow any algae}, \textit{Grow any bonsai}, \textit{Grow any bush}, \textit{Grow any cactus}\\
                                     & \textit{Grow any carnivorous}, \textit{Grow any grass}, \textit{Grow any living\_thing}, \textit{Grow any plant}\\
                                     & \textit{Grow any rose}, \textit{Grow any tea}, \textit{Grow any tree}, \textit{Grow blue algae}\\
                                     & \textit{Grow blue bonsai}, \textit{Grow blue bush},\textit{Grow blue cactus}, \textit{Grow blue carnivorous}\\
                                     & \textit{Grow blue grass}, \textit{Grow blue living\_thing}, \textit{Grow blue plant}, \textit{Grow blue rose}\\
                                     & \textit{Grow blue tea}, \textit{Grow blue tree},\textit{Grow green algae}, \textit{Grow green bonsai}\\
                                     & \textit{Grow green bush}, \textit{Grow green cactus},  \textit{Grow green carnivorous}, \textit{Grow green grass}\\
                                     & \textit{Grow green living\_thing}, \textit{Grow green plant}, \textit{Grow green rose}, \textit{Grow green tea}\\
                                     & \textit{Grow green tree}, \textit{Grow red algae}, \textit{Grow red bonsai}, \textit{Grow red bush}\\
                                     & \textit{Grow red cactus}, \textit{Grow red carnivorous}, \textit{Grow red grass}, \textit{Grow red living\_thing}\\
                                     & \textit{Grow red plant}, \textit{Grow red rose}, \textit{Grow red tea}, \textit{Grow red tree}\\
    \end{tabular}
\end{table}

\begin{figure*}[!h]
    \centering
    \subfigure[\label{fig:reward_func_in_rl}\textbf{Reward Function, no imagination}]{\includegraphics[width=0.48\textwidth]{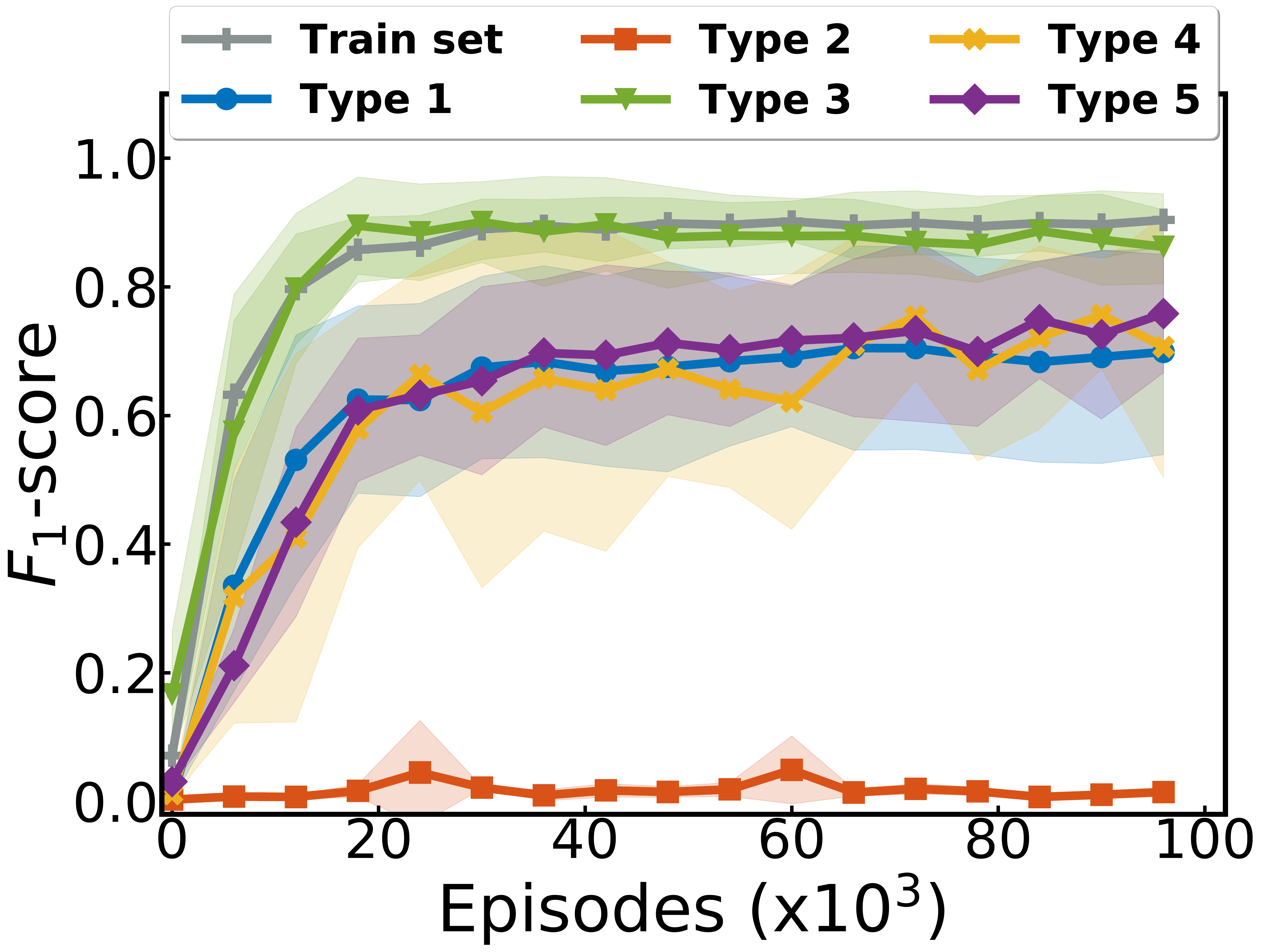}}
    \subfigure[\label{fig:gene_no_imagined}\textbf{Policy, no imagination}]{\includegraphics[width=0.48\textwidth]{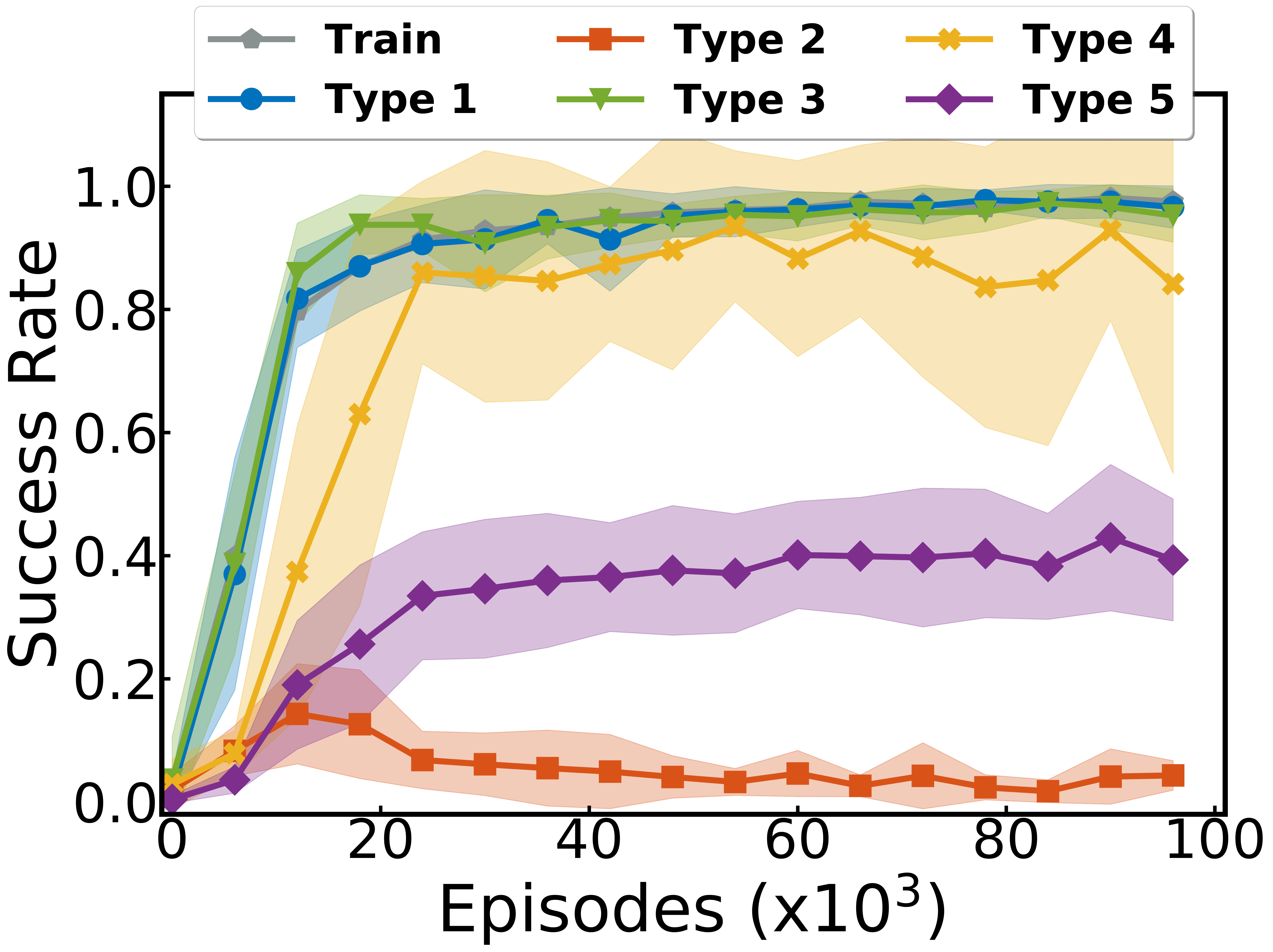}}
    \subfigure[\label{fig:gene10}\textbf{Policy, imagination half way}]{\includegraphics[width=0.48\textwidth]{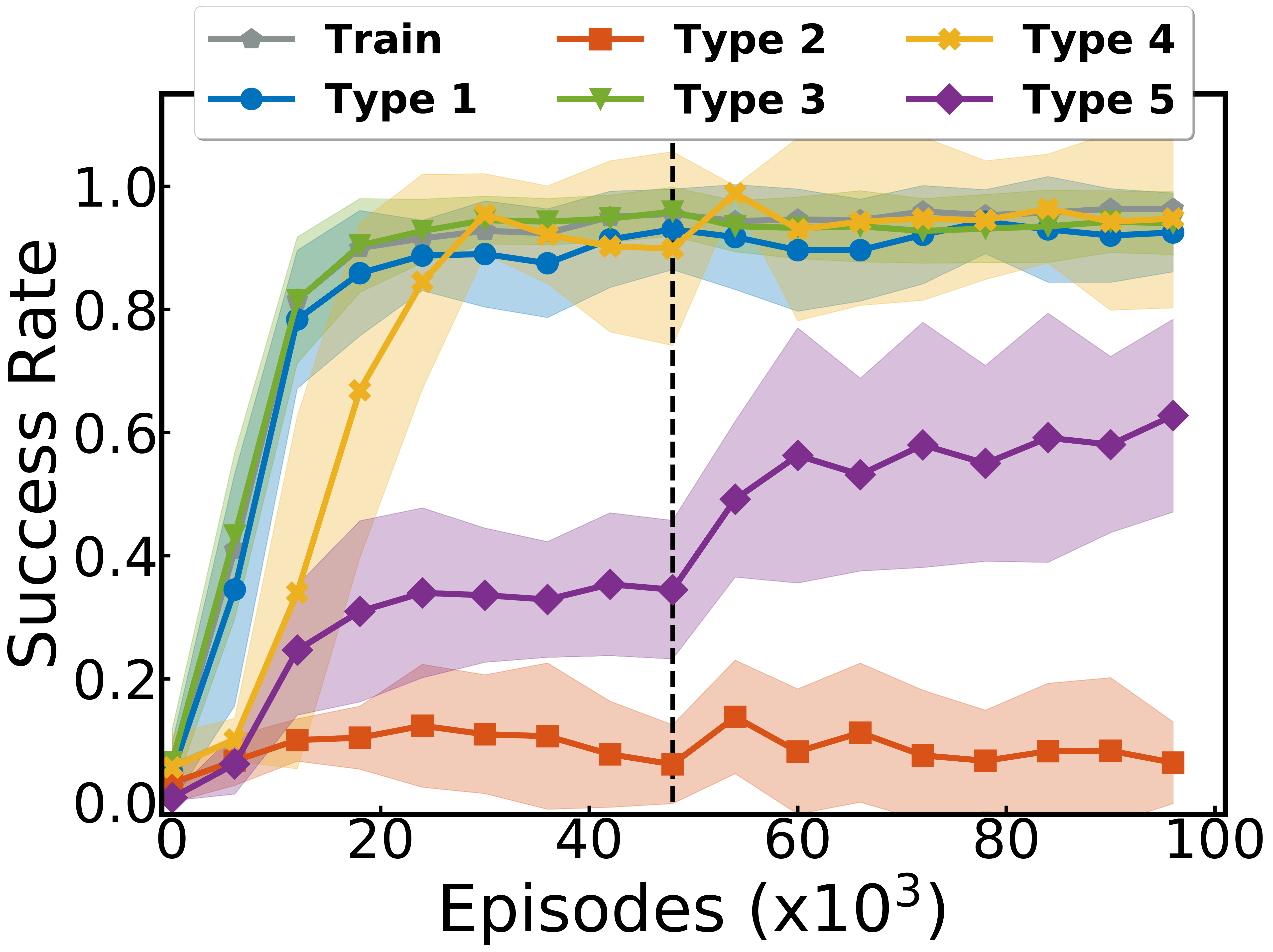}}
    \subfigure[\label{fig:gene80}\textbf{Policy, imagination early}]{\includegraphics[width=0.48\textwidth]{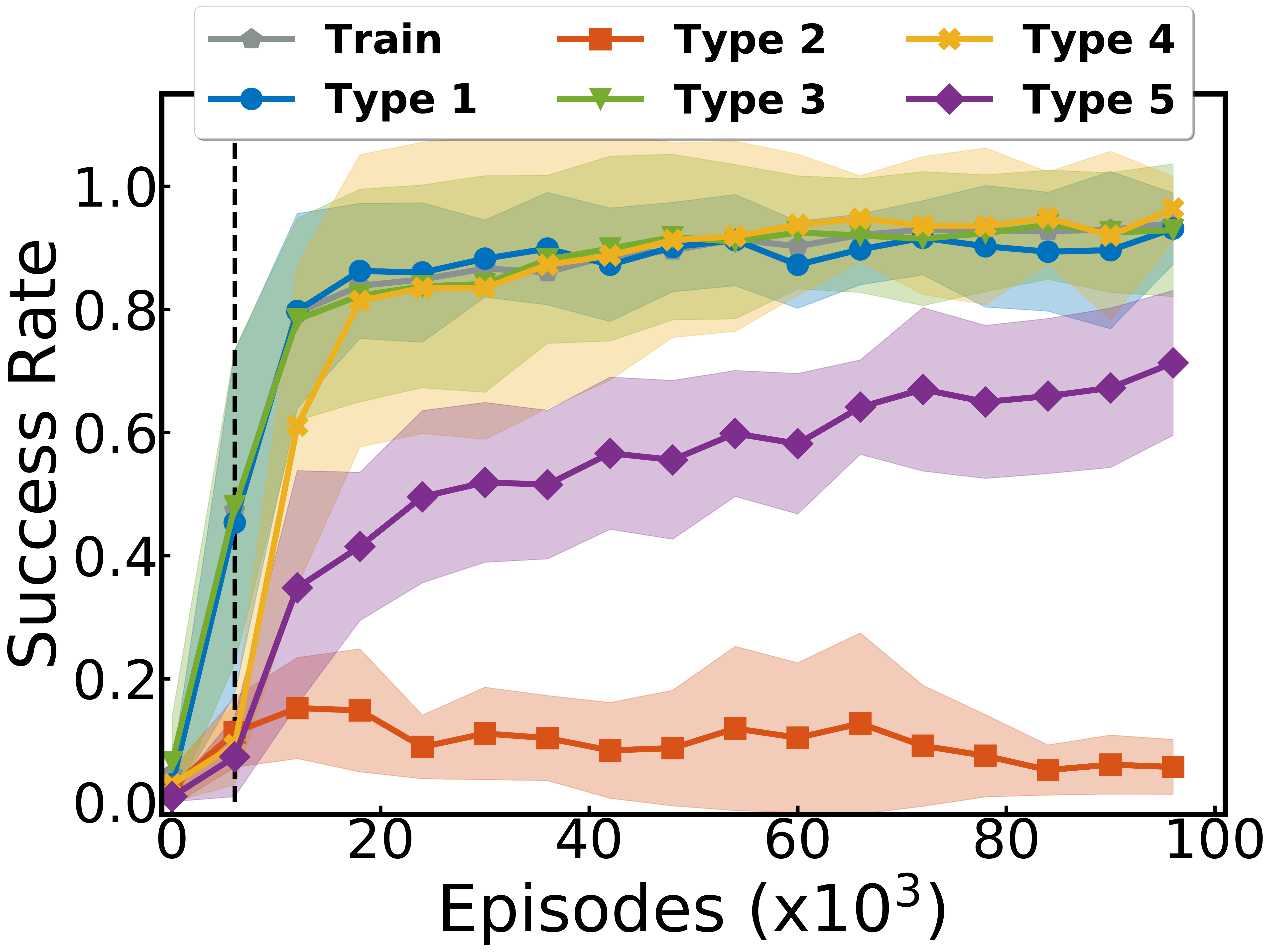}}
    
    \caption{\textbf{Zero-shot and n-shot generalizations of the reward function and policy.} Each figure represents the training and testing performances (split by generalization type) for the reward (a), and the policy (b, c, d). (a) and (b) represent zero-shot performance in the \textit{no imagination} conditions. In (c) and (d), agents start to imagine goals as denoted by the vertical dashed line. Before that line, $\SR$ evaluate zero-shot generalization. After, it evaluates the n-shot generalization, as agent can train autonomously on imagined goals.\label{fig:suppl_gen}}
\end{figure*}

\paragraph{Experiments.}
Figure~\ref{fig:suppl_gen} presents training and generalization performance of the reward function and policy. We evaluate the generalization of the reward function via its average $F_1$ score on $\G^\test$, the generalization of the policy by $\SR_\test$.

\textit{Reward function zero-shot generalization.}
When the reward function is trained in parallel of the policy, we monitor its zero-shot generalization capabilities by computing the $F_1$-score over a dataset collected separately with a trained policy run on goals from  $\G^\test$ (kept fixed across runs for fair comparisons). As shown in Figure~\ref{fig:reward_func_in_rl}, the reward function exhibits good zero-shot generalization properties over 4 types of generalization after $25 \times 10^3$ episodes. Note that, because we test on data collected with a different RL policy, the $F_1$-scores presented in Figure~\ref{fig:reward_func_in_rl} may not faithfully describe the true generalization of the reward function during co-training. 

\textit{Policy zero-shot generalization.}
The zero-shot performance of the policy is evaluated in Figure~\ref{fig:gene_no_imagined} (\textit{no imagination} condition) and in the period preceding goal imagination in Figure~\ref{fig:gene10} and \ref{fig:gene80} (before vertical dashed line). The policy shows excellent zero-shot generalization properties for Type 1, 3 and 4, average zero-shot generalization on Type 5 and fails to generalize on Type 2. Type 1, 3 and 4 can be said to have similar levels of difficulty, as they all require to learn two concepts individually before combining them at test time. Type 2 is much more difficult as the meaning of only one word is known. The language encoder indeed receives a new word token which seems to disturb behavior. As said earlier, zero-shot generalization on Type 5 cannot do better than 0.5, as it cannot infer that plants only require water.

\textit{Policy n-shot generalization.} When goal imagination begins (Figures~\ref{fig:gene10} and \ref{fig:gene80} after the vertical line), agents can imagine goals and train on them. This means that $\SR$ evaluates n-shot policy generalization. Agents can now perform \textit{behavior adaptation}. They can learn that plants need water. As they learn this, their generalization performance on goals from Type 5 increases and goes beyond 0.5. Note that this effects fights the zero-shot generalization. By default, policy and reward function apply zero-shot generalization: e.g. they bring water or food equally to plants. Behavioral adaptation attempts to modify that default behavior. Because of the poor zero-shot generalization of the reward on goals of Type 2, agents cannot hope to learn Type 2 behaviors. Moreover, Type 2 goals cannot be imagined, as the word \textit{flower} is unknown to the agent.

\clearpage

\section{Focus on exploration}
\label{sec:suppl_exploration}
\paragraph{Interesting Interactions.} \textit{Interesting interactions} are trajectories of the agent that humans could infer as goal-directed. If an agent brings water to a plant and grows it, it makes sense for a human. If it then tries to do this for a lamp, it also feels goal-directed, even though it does not work. This type of behavior characterizes the penchant of agents to interact with objects around them, to try new things and, as a result, is a good measure of exploration. 

\paragraph{Sets of interesting interactions.} We consider three sets of interactions: 1) interactions related to training goals; 2) to testing goals; 3) the extra set. This \textit{extra set} contains interactions where the agent brings water or food to a piece of furniture or to another supply. Although such behaviors do not achieve any of the goals, we consider them as interesting exploratory behaviors. Indeed, they testify that agents try to achieve imagined goals that are meaningful from the point of view of an agent that does not already know that doors cannot be grown, i.e. corresponding to a meaningful form of generalization after discovering that animals or plants can be grown (e.g. \textit{grow any door}). 

\paragraph{The Interesting Interaction Count metric.} 
We count the number of interesting interactions computed over all final transitions from the last $600$ episodes (1 epoch). Agents do not need to target these interactions, we just report the number of times they are experienced. Indeed, the agent does not have to target a particular interaction for the trajectory to be interesting from an exploratory point of view. The \her mechanism ensures that these trajectories can be replayed to learn about any goal, imagined or not. Computed on the extra set, the \textit{Interesting Interaction Count} (\itwoc) is the number of times the agent was found to bring supplies to a furniture or to other supplies over the last epoch: 
    \begin{displaymath}
    \itwoc_{\text{extra}} = \sum_{i\in \mathcal{I}=\G_\text{extra}} \sum_{t=1}^{600} \delta_{i,t},
    \end{displaymath}

    where $\delta_{i,t} = 1$ if interaction $i$ was achieved in episode $t$, $0$ otherwise and $\mathcal{I}$ is the set of interesting interactions (here from the extra set) performed during an epoch. 
    
\begin{figure*}[!h]
  \centering
    \subfigure[\label{fig:sc_train}]{\includegraphics[width=0.328\textwidth]{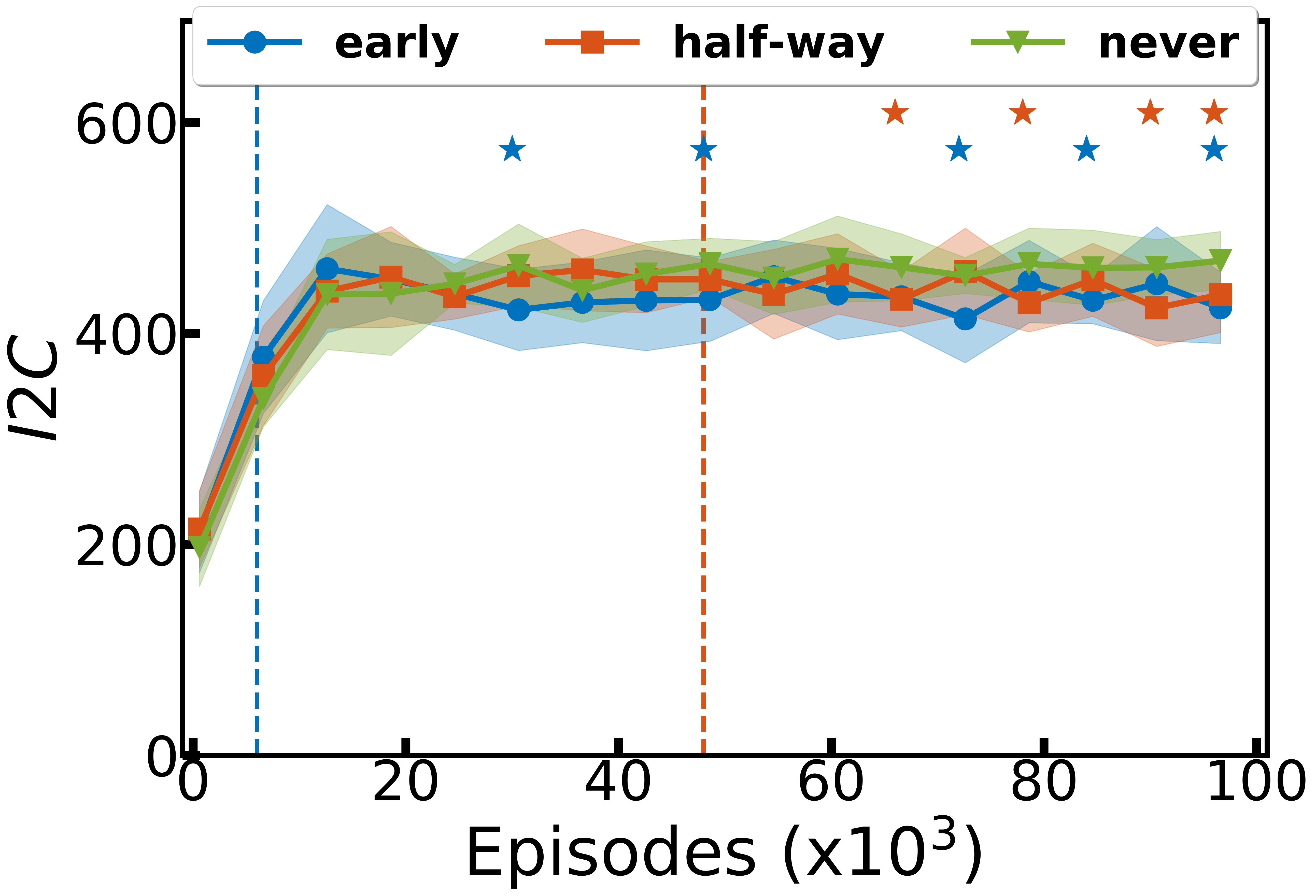}}
    \subfigure[\label{fig:sc_test}]{\includegraphics[width=0.328\textwidth]{figures/GOAL_INVENTION_exploration_count_reward_test_set_compressed.pdf}}
    \subfigure[\label{fig:sc_extra}]{\includegraphics[width=0.328\textwidth]{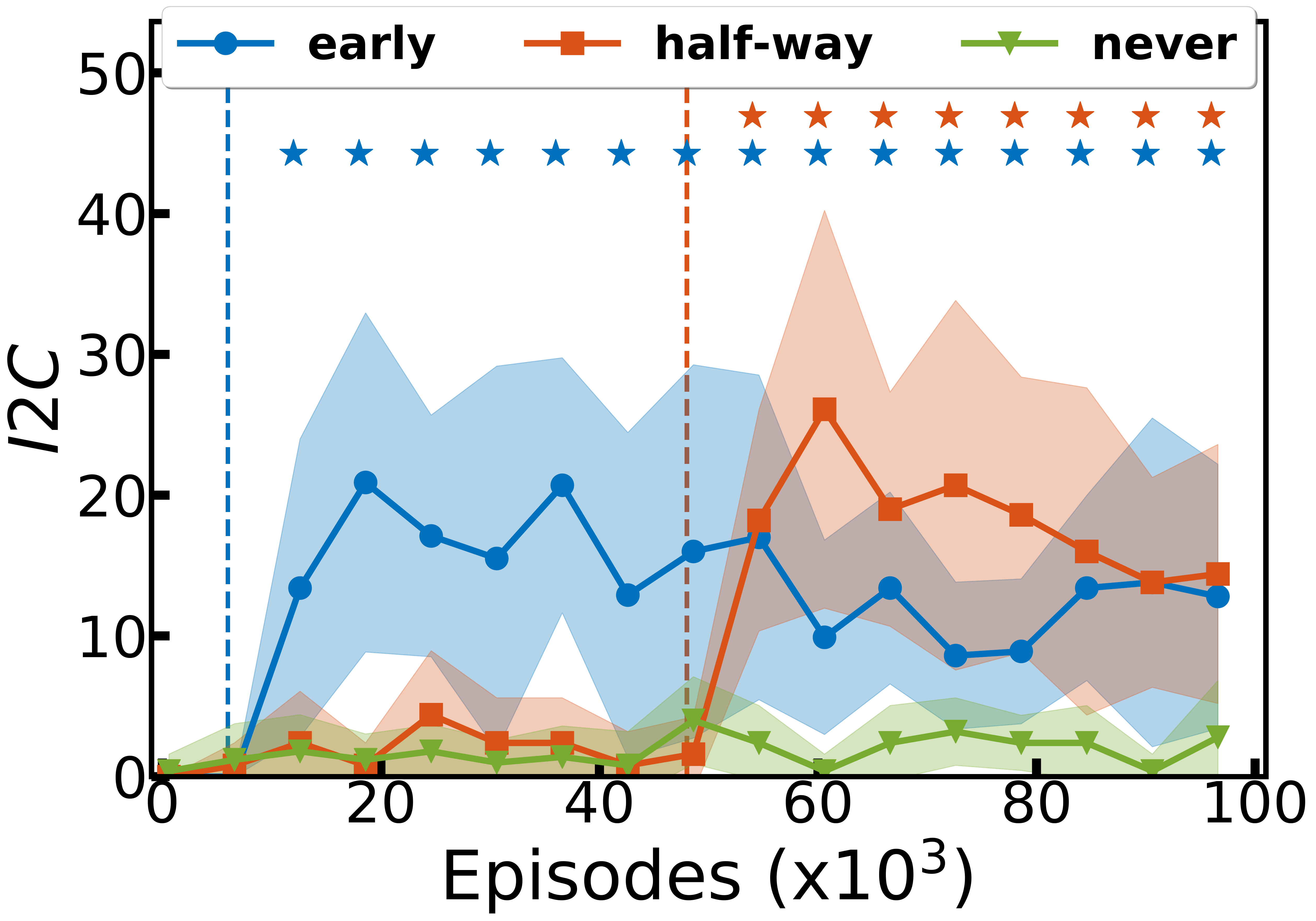}}
  \caption{\textbf{Exploration metrics} (a) Interesting interaction count (\itwoc) on training set, (b) \itwoc on testing set, (c) \itwoc on extra set. Goal imagination starts early (vertical blue line), half-way (vertical orange line) or does not start (\textit{no imagination} baseline in green).}
  \label{fig:explo_metrics}
\end{figure*}

Agents that are allowed to imagine goals achieve higher scores in the testing and extra sets of interactions, while maintaining similar exploration scores on the training set, see Figures~\ref{fig:sc_train} to \ref{fig:sc_extra}. 

\clearpage

\section{Focus on Goal Imagination}
\label{sec:suppl_goal_imagination}

\begin{figure}[!h]
      \centering
      \subfigure[\label{fig:suppl_known_earl}\textbf{CGH}]{\includegraphics[width=0.32\textwidth]{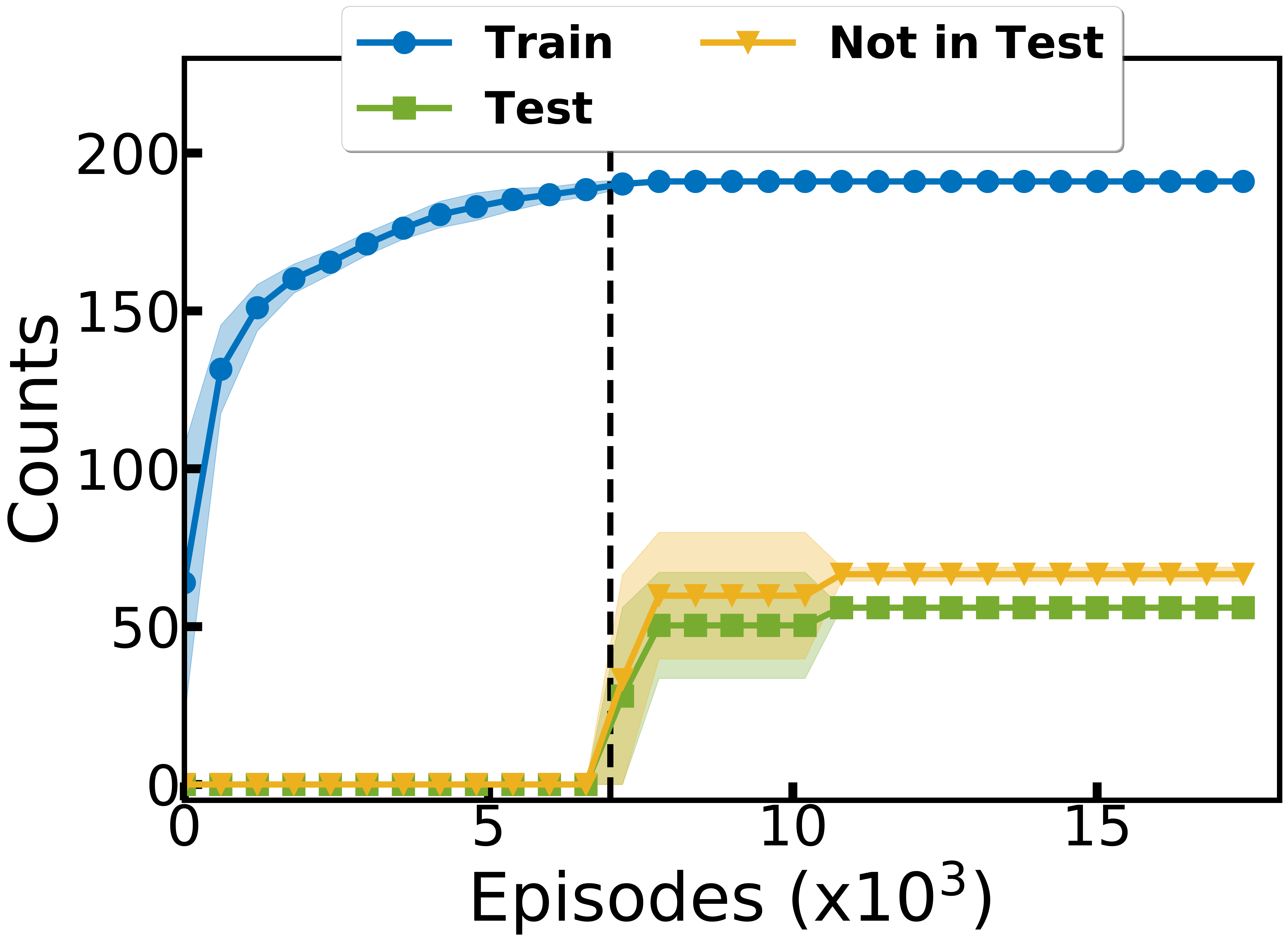}}
      \subfigure[\label{fig:suppl_known_low_cov}\textbf{Low Coverage}]{\includegraphics[width=0.32\textwidth]{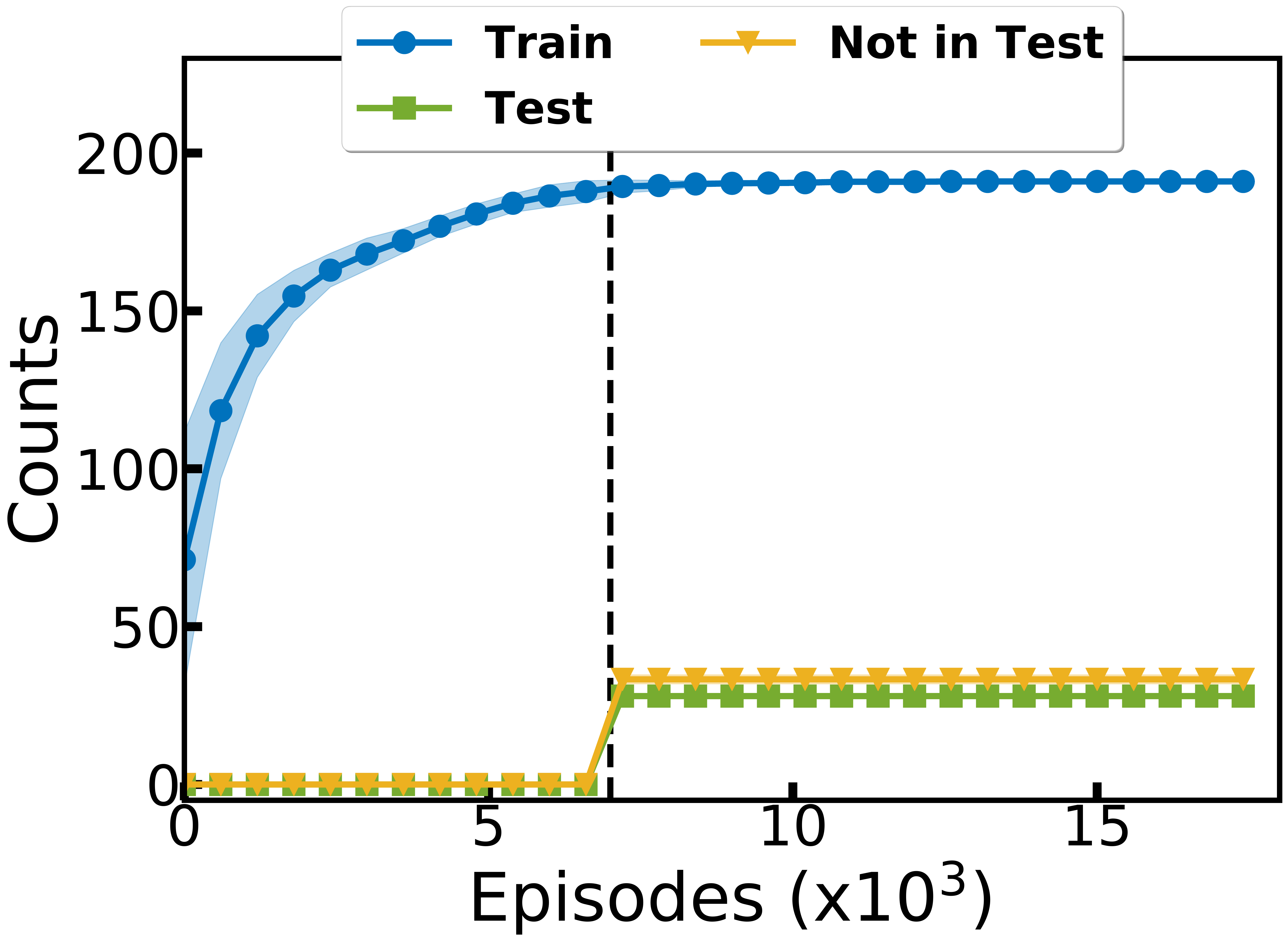}} 
      \subfigure[\label{fig:suppl_known_low_prec}\textbf{Low Precision}]{\includegraphics[width=0.32\textwidth]{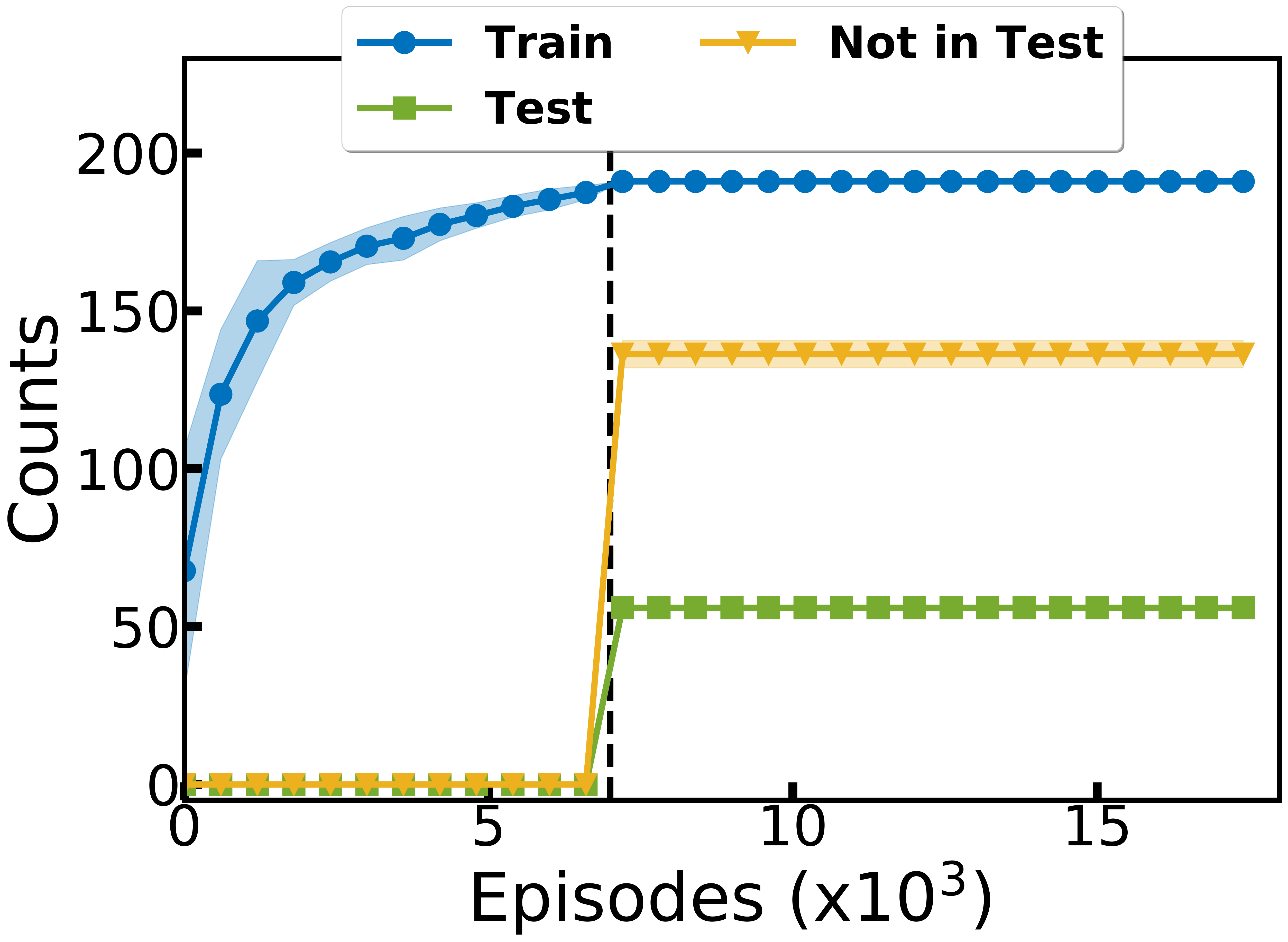}} \\
      \subfigure[\label{fig:suppl_known_oracle}\textbf{Oracle}]{\includegraphics[width=0.32\textwidth]{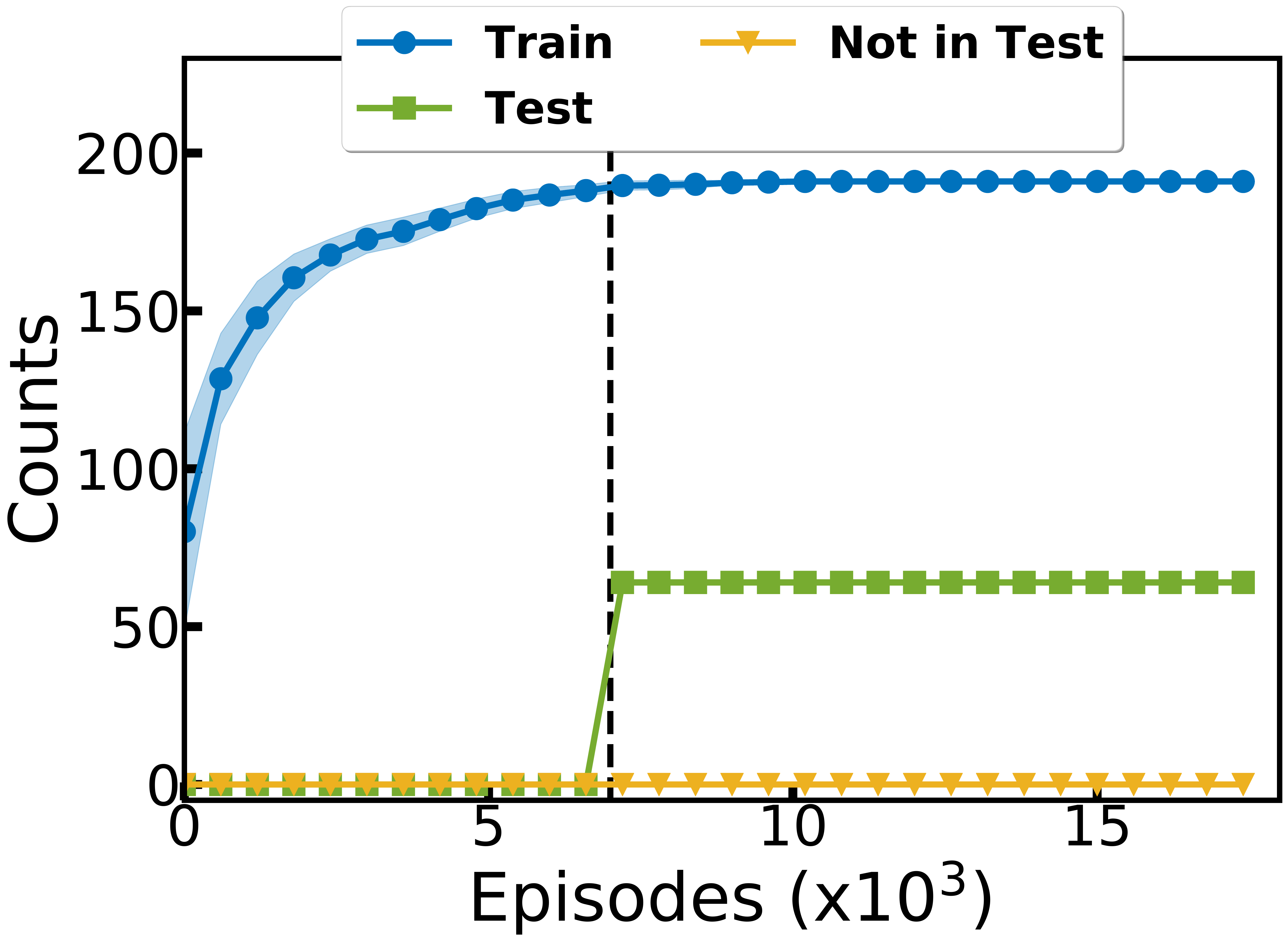}} 
      \subfigure[\label{fig:suppl_known_random}\textbf{Random Goals}]{\includegraphics[width=0.32\textwidth]{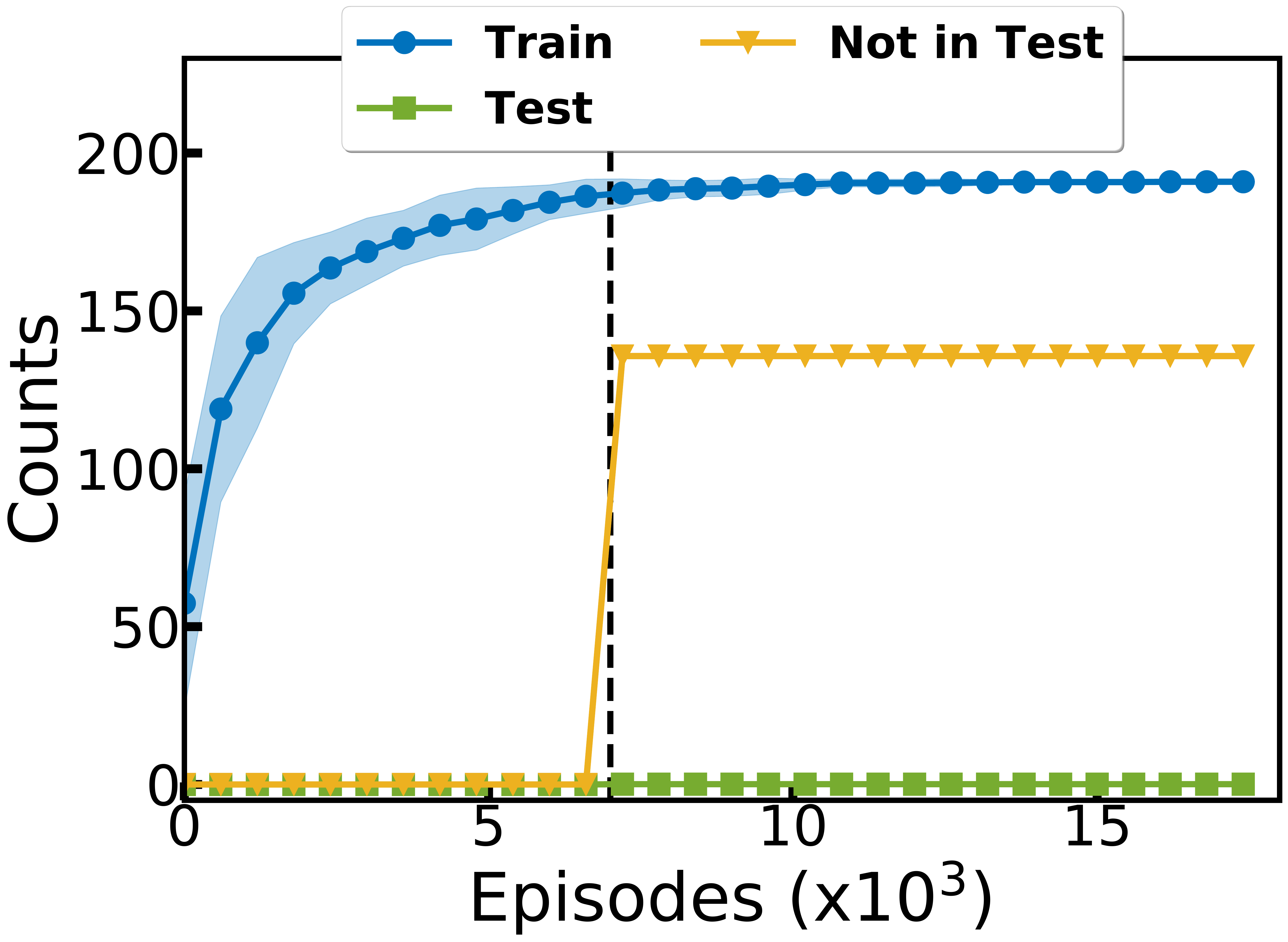}}
      \caption{\textbf{Evolution of known goals for various goal imagination mechanisms.} All graphs show the evolution of the number of goals from $\G^\train$, $\G^\test$ and others in the list of known goals $\G_\text{known}$. We zoom on the first epochs, as most goals are discovered and invented early. Vertical dashed line indicates the onset of goal imagination. (a) \CGH; (b) Low Coverage; (c) Low precision; (d) Oracle; (e) Random Goals. \label{fig:suppl_known_goals}}
\end{figure}

Algorithm~\ref{alg:suppl_imagination} presents the algorithm underlying our goal imagination mechanism. This mechanism is inspired from the \textit{Construction Grammar} (CG) literature and generates new sentences by composing known ones \cite{goldberg2003constructions}. It computes sets of equivalent words by searching for sentences with an edit distance of $1$: sentences where only one word differs. These words are then labelled equivalent, and can be substituted in known sentences. Note that the goal imagination process filters goals that are already known. Although all sentences from $\G^\train$ can be imagined, there are filtered out of the imagined goals as they are discovered. Imagining goals from $\G^\train$ before they are discovered drives the exploration of \imagine agents. In our setup, however, this effect remains marginal as all the goals from $\G^\train$ are discovered in the first epochs (see Figure~\ref{fig:suppl_known_goals}).

\begin{minipage}{.6\textwidth} %
        \vspace{0.5cm}
        \captionof{algocf}{Goal Imagination. The edit distance between two sentences refers to the number of words to modify to transform one sentence into the other.\label{alg:suppl_imagination}}
        \begin{algorithmic}[1]
            \STATE \textbf{Input:} $\G_\text{known}$ (discovered goals)
            \STATE \textbf{Initialize:} \textit{word\_eq} (list of sets of equivalent words, empty) 
            \STATE \textbf{Initialize:} \textit{goal\_template} (list of template sentences used for imagining goals, empty)
            \STATE \textbf{Initialize:} $\G_\text{im}$ (empty)
            \FOR[Computing word equivalences]{$g_\text{NL}$ in $\G_\text{known}$}  
                \STATE \textit{new\_goal\_template} = True
                \FOR{$g_m$ in \textit{goal\_template}}
                    \IF{edit\_distance$(g_\text{NL}, g_m) < 2$}
                        \STATE \textit{new\_goal\_template} = False
                        \IF{edit\_distance$(g_\text{NL}, g_m) == 1$}
                            \STATE $w_1, w_2 \gets $ \textit{get\_non\_matching\_words$(g_\text{NL}, g_m)$}
                            \IF{$w_1$ and $w_2$ not in any of \textit{word\_eq} sets}
                                 \STATE \textit{word\_eq}.add(\{$w_1,w_2$\})
                            \ELSE
                                \FOR{\textit{eq\_set} in \textit{word\_eq}}   
                                    \IF{$w_1 \in \:$\textit{eq\_set} or $w_2 \in \: $\textit{eq\_set}}
                                        \STATE \textit{eq\_set} = \textit{eq\_set} $\cup$ \{$w_1,w_2$\}
                                    \ENDIF
                                \ENDFOR
                            \ENDIF
                        \ENDIF
                    \ENDIF
                \ENDFOR
                \IF{\textit{new\_goal\_template}}
                    \STATE \textit{goal\_template}.add($g_\text{NL}$)
                \ENDIF
            \ENDFOR
            \FOR[Generating new sentences]{$g$ in \textit{goal\_template}}  
                \FOR{$w$ in $g$}
                    \FOR{\textit{eq\_set} in \textit{word\_eq}}
                        \IF{$w \in\:$ \textit{eq\_set}}
                            \FOR{$w'$ in \textit{eq\_set}}
                                \STATE $g_{im} \gets$ replace($g$, $w$, $w'$)
                                \IF{$g_{im} \notin \G_\text{known}$}
                                    \STATE $\G_\text{im} = \G_\text{im} \: \cup $ \{$g_{im}$\}
                                \ENDIF
                            \ENDFOR
                        \ENDIF
                    \ENDFOR
                \ENDFOR
            \ENDFOR
            \STATE $\G_\text{im} = \G_\text{im} \setminus \G_\text{known}$ \hspace{1cm}\{filtering known goals.\}
        \end{algorithmic}
\end{minipage} %
\begin{minipage}{.4\textwidth} %
        \centering
        \includegraphics[width=0.9\textwidth]{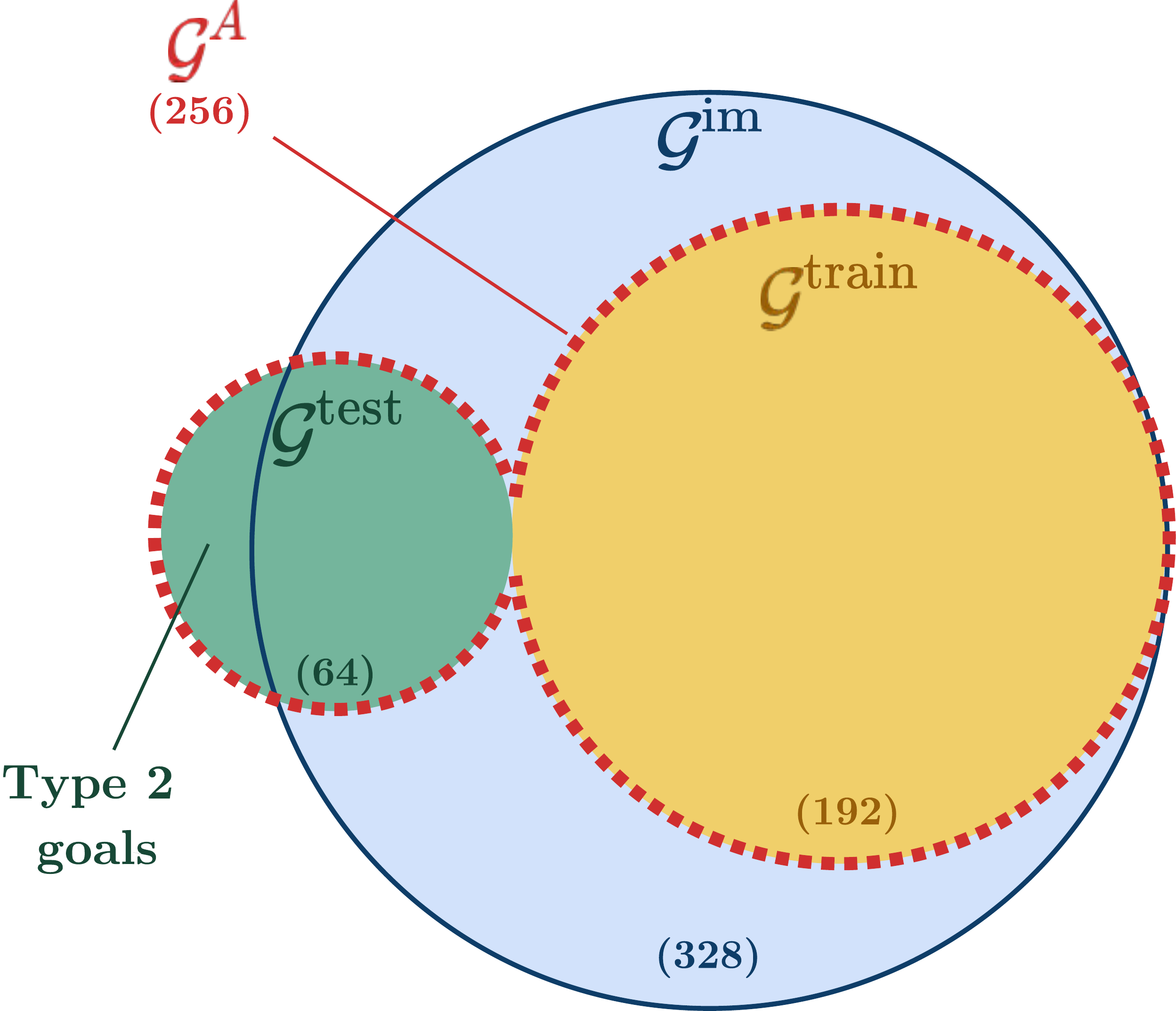}
        \captionof{figure}{Venn diagram of goal spaces.\label{fig:suppl_venn}}  
\end{minipage}

\begin{table*}[!htbp]
    \centering
    \caption{All imaginable goals $\G^\text{im}$ generated by the Construction Grammar Heuristic.}
    \label{tab:imaginable_goals}
    \begin{tabular}{c|l}
        \multirow{3}{*}{Goals from $\G^\train$}
        & \\

        &  $\G^\train$. (Note that known goals are filtered from the set of imagined goals.  \\  & However, any goal from $\G^\train$ can be imagined before it is encountered \\ 
        & in the interaction with \SP.) \\ 
        & \\
        \hline
        \multirow{3}{*}{\shortstack{Goals from  $\G^\test$}} & \multirow{3}{*}{All goals from Type 1, 3, 4 and 5, see Table~\ref{tab:test_descriptions}} \\ & \\ & \\
        \hline
        \multirow{4}{*}{\shortstack{Syntactically\\ incorrect goals}} 
             & \\
             & \textit{Go bottom top}, \textit{Go left right}, \textit{Grasp red blue thing}, \\
             & \textit{Grow blue red thing}, \textit{Go right left}, \textit{Go top bottom}, \\
             & \textit{Grasp green blue thing}, \textit{Grow green red thing}, \textit{Grasp green red thing} \\
             & \textit{Grasp blue green thing}, \textit{Grasp blue red thing}, \textit{Grasp red green thing}. \\ & \\
        \hline
        \multirow{18}{*}{\shortstack{Syntactically\\ correct but\\unachievable goals}} 
            & \\
             & \textit{Go center bottom}, \textit{Go center top}, \textit{Go right center},  \textit{Go right bottom}, \\
             & \textit{Go right top}, \textit{Go left center}, \textit{Go left bottom}, \textit{Go left top}, \\
             & \textit{Grow green cupboard},  \textit{Grow green sink}, \textit{Grow blue lamp}, \textit{Go center right}, \\
             & \textit{Grow green window}, \textit{Grow blue carpet}, \textit{Grow red supply}, \textit{Grow any sofa}, \\
             & \textit{Grow red sink}, \textit{Grow any chair},  \textit{Go top center}, \textit{Grow blue table}, \\
             & \textit{Grow any door},  \textit{Grow any lamp}, \textit{Grow blue sink}, \textit{Go bottom center}, \\
             & \textit{Grow blue door}, \textit{Grow blue supply},  \textit{Grow green carpet}, \textit{Grow blue furniture}, \\
             & \textit{Grow green supply}, \textit{Grow any window},  \textit{Grow any carpet}, \textit{Grow green furniture}, \\
             & \textit{Grow green chair}, \textit{Grow green food}, \textit{Grow any cupboard}, \textit{Grow red food}, \\
             & \textit{Grow any table}, \textit{Grow red lamp} , \textit{Grow red door},  \textit{Grow any food}, \\
             & \textit{Grow blue window}, \textit{Grow green sofa}, \textit{Grow blue sofa},  \textit{Grow blue desk}, \\
             & \textit{Grow any sink}, \textit{Grow red cupboard}, \textit{Grow green door}, \textit{Grow red furniture}, \\
             & \textit{Grow blue food}, \textit{Grow red desk} , \textit{Grow red table},  \textit{Grow blue chair}, \\
             & \textit{Grow red sofa}, \textit{Grow any furniture}, \textit{Grow red window}, \textit{Grow any desk}, \\
             & \textit{Grow blue cupboard}, \textit{Grow red chair}, \textit{Grow green desk}, \textit{Grow green table}, \\
             & \textit{Grow red carpet}, \textit{Go center left}, \textit{Grow any supply},  \textit{Grow green lamp}, \\
             & \textit{Grow blue water}, \textit{Grow red water},  \textit{Grow any water}, \text{Grow green water}, \\
             & \textit{Grow any water}, \text{Grow green water}. \\
             & \\
    \end{tabular}

\end{table*}

\clearpage

\paragraph{Imagined goals.} We run our goal imagination mechanism based on the Construction Grammar Heuristic (\CGH) from $\G^\train$. After filtering goals from $\G^\train$, this produces $136$ new imagined sentences. Table~\ref{tab:imaginable_goals} presents the list of these goals while Figure~\ref{fig:suppl_venn} presents a Venn diagram of the various goal sets. Among these $136$ goals, $56$ belong to the testing set $\G^\test$. This results in a coverage of $87.5\%$ of $\G^\test$, and a precision of $45\%$. In goals that do not belong to $\G^\test$, goals of the form \textit{Grow} + \{\textit{any}\} $\cup$ \textbf{color} + \textbf{furniture} $\cup$ \textbf{supplies} (e.g. \textit{Grow any lamp}) are \textit{meaningful} to humans, but are not achievable in the environment (\textit{impossible}).  

\paragraph{Variants of goal imagination mechanisms.}
Main Section~\ref{sec:res_im_properties} investigates variants of our goal imagination mechanisms:
\begin{enumerate}
    \item \textit{Lower coverage:} To reduce the coverage of \CGH while maintaining the same precision, we simply filter half of the goals that would have been imagined by \CGH. This filtering is probabilistic, resulting in different imagined sets for different runs. It happens online, meaning that the coverage is always half of the coverage that \CGH would have had at the same time of training.
    \item \textit{Lower precision:} To reduce precision while maintaining the same coverage, we sample a random sentence (random words from the words of $\G^\train$) for each goal imagined by \CGH that does not belong to $\G^\test$. Goals from $\G^\test$ are still imagined via the \CGH mechanism. This variants only doubles the imagination of sentences that do not belong to $\G^\test$.
    \item \textit{Oracle:} Perfect precision and coverage is achieved by filtering the output of \CGH, keeping only goals from $\G^\test$. Once the $56$ goals that \CGH can imagine are imagined, the oracle variants adds the $8$ remaining goals: those including the word \textit{flower} (Type 2 generalization).
    \item \textit{Random goals:} Each time \CGH would have imagined a new goal, it is replaced by a randomly generated sentence, using words from the words of $\G^\train$.
\end{enumerate}
Note that all variants imagine goals at the same speed as the \CGH algorithm. They simply filter or add noise to its output, see Figure~\ref{fig:suppl_known_goals}.

\begin{figure}[!h]
      \centering
      \includegraphics[width=0.5\textwidth]{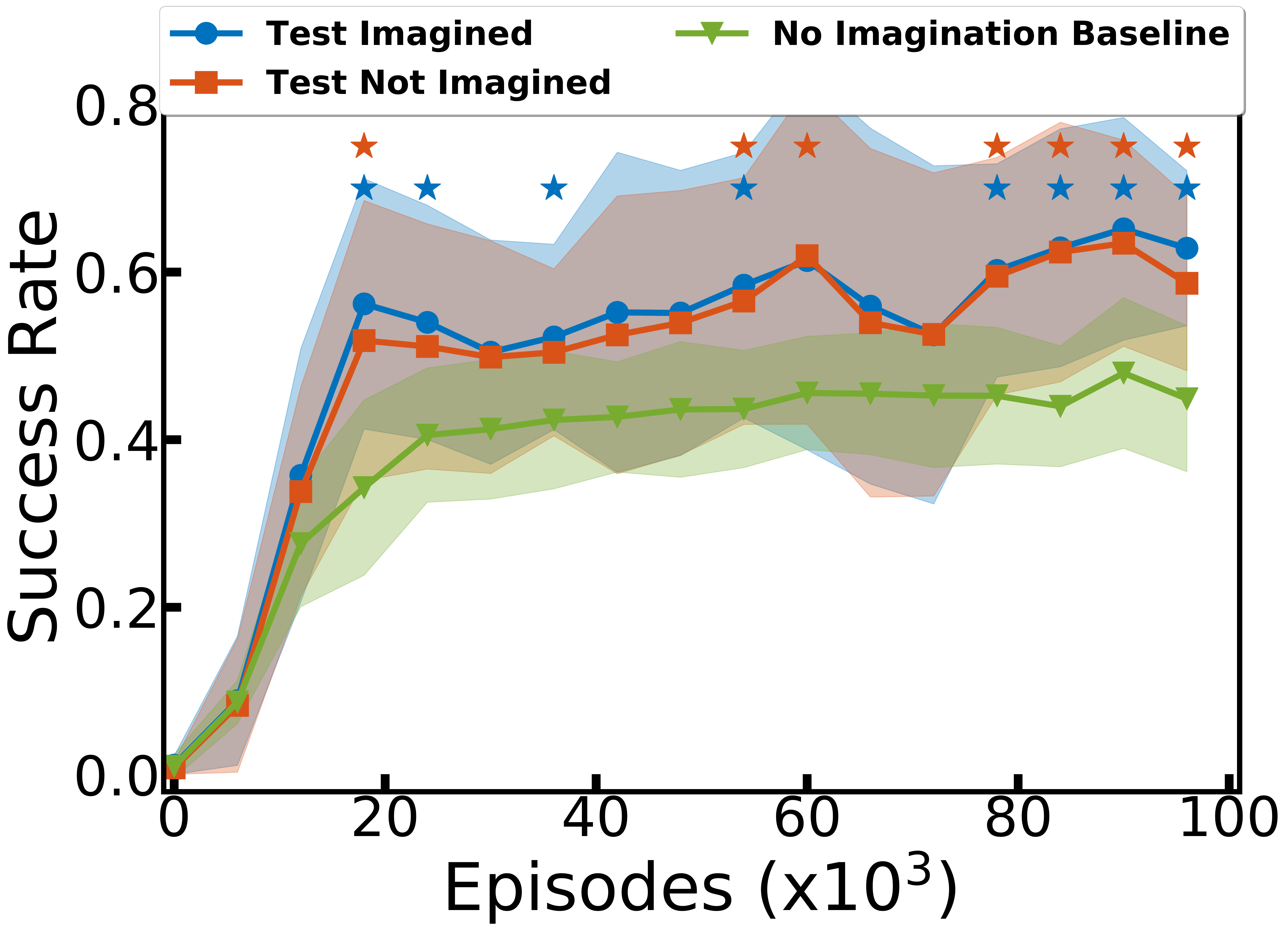}
      \caption{\textbf{Zero-shot versus n-shot.} We look at the \textit{Low Coverage} variant of our goal imagination mechanism that only covers $43.7\%$ the test set with a $45\%$ precision. We report success rates on testing goals of Type 5 (\textit{grow + plant}) and compare with the \textit{no imagination} baseline (green). We split in two: goals that were imagined (blue), and goals that were not (orange). \label{fig:suppl_halftc}}
\end{figure}

\newpage
\paragraph{Effect of low coverage on generalization.}
In Main Section~\ref{sec:res_im_properties}, we compare our goal imagination mechanism to a \textit{Low Coverage} variant that only covers half of the proportion of $\G^\test$ covered by \CGH ($44\%$). Figure~\ref{fig:suppl_halftc} shows that the generalization performance on goals from $\G^\test$ that the agent imagined (n-shot generalization, blue) are not significantly higher than the generalization performance on goals from $\G^\test$ that were not imagined (zero-shot generalization). As they are both significantly higher than the \textit{no imagination} baseline, this implies that training on imagined goals boosts zero-shot generalization on similar goals that were not imagined.

\begin{figure*}[!h]
  \centering
    \subfigure[\label{fig:supp_explo_train}\textbf{\itwoc on $\G^\train$}]{\includegraphics[width=0.328\textwidth]{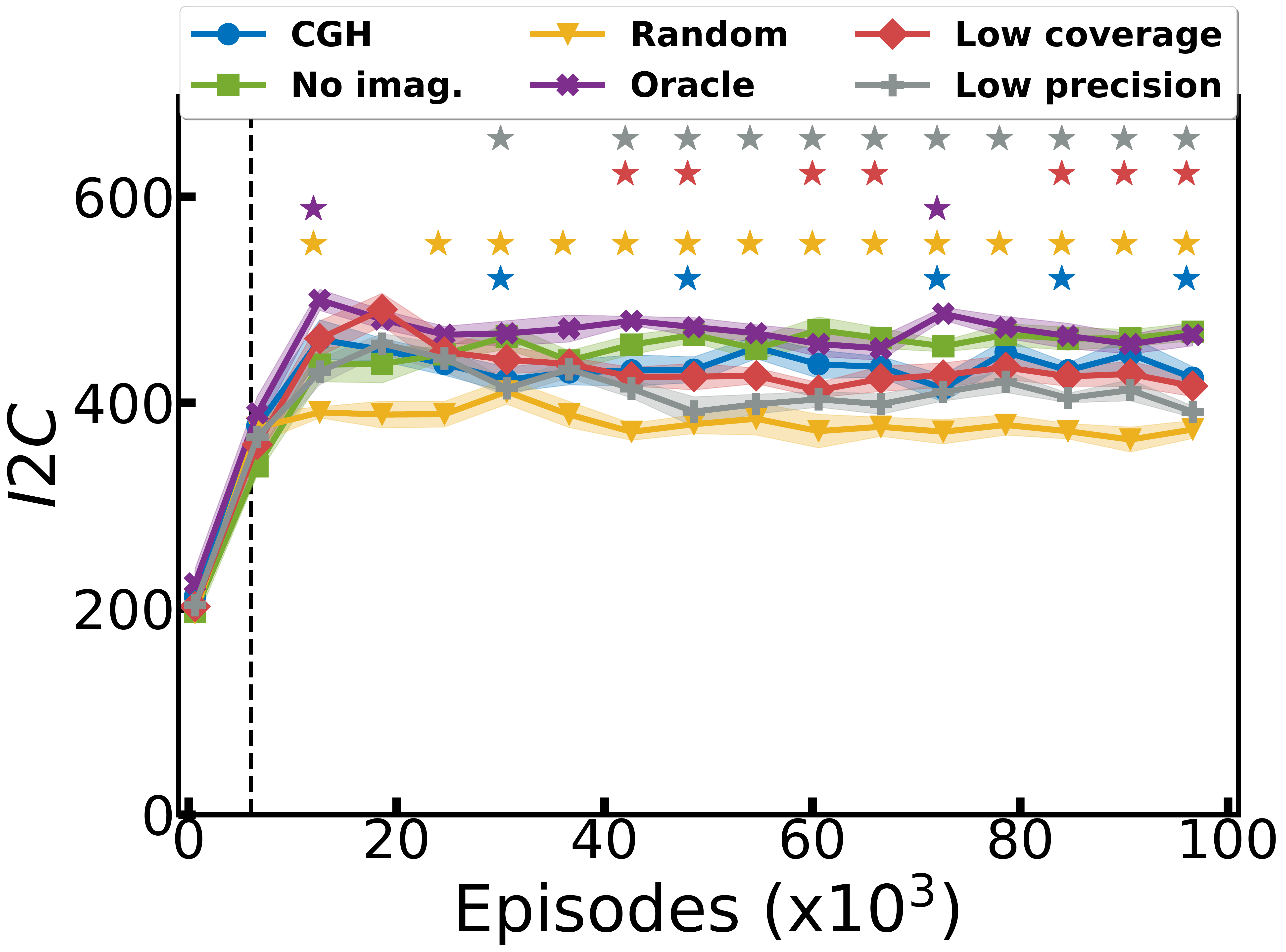}}
    \subfigure[\label{fig:supp_explo_test}\textbf{\itwoc on $\G^\test$}]{\includegraphics[width=0.328\textwidth]{figures/explo_test.pdf}}
    \subfigure[\label{fig:supp_explo_extra}\textbf{\itwoc on $\G^\text{extra}$}]{\includegraphics[width=0.328\textwidth]{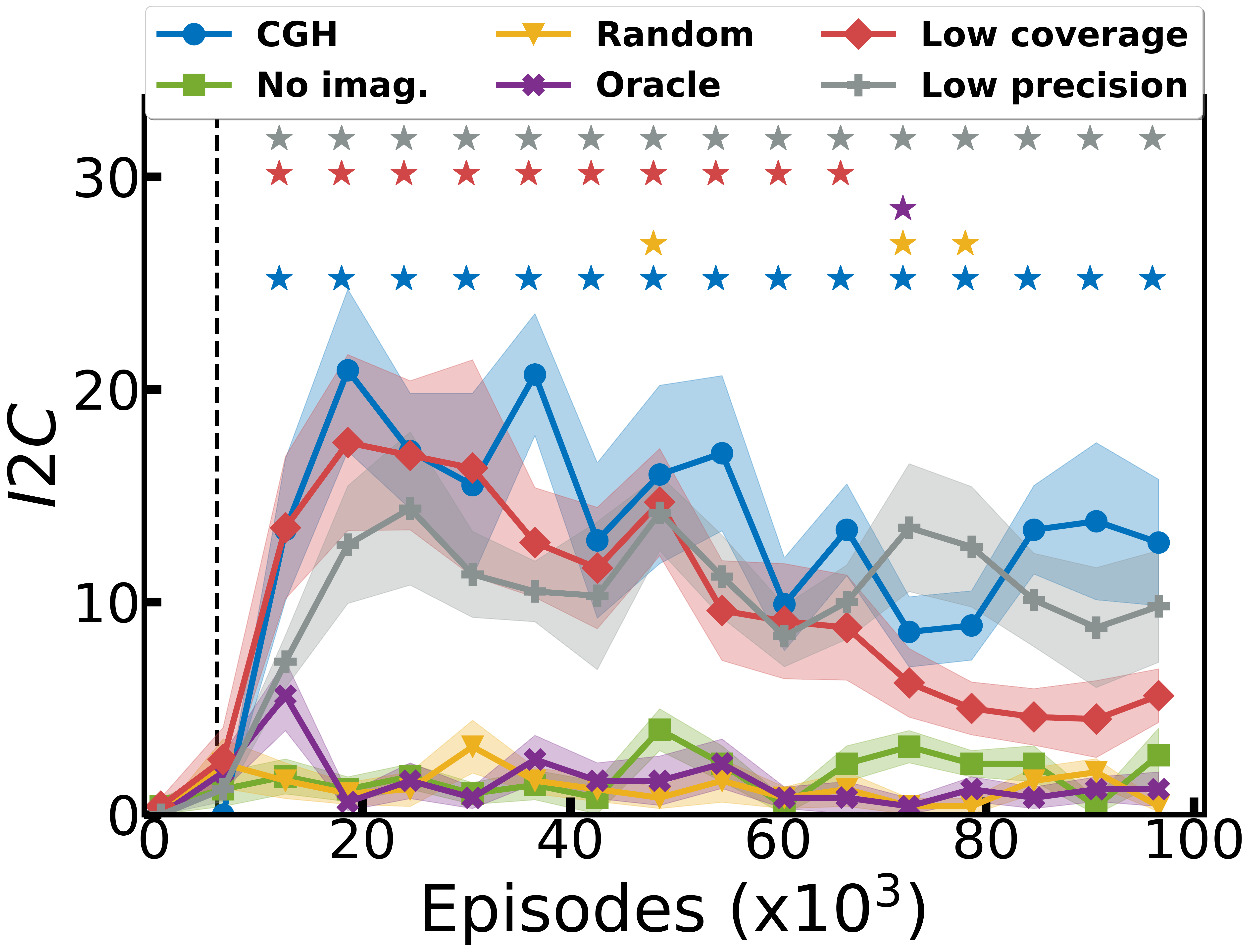}}
  \caption{\textbf{Exploration metrics for different goal imagination mechanisms}: (a) Interesting interaction count (\itwoc) on training set, \itwoc on testing set, (c) \itwoc on extra set.  Goal imagination starts early (vertical line), except for the \textit{no imagination} baseline (green). Standard errors of the mean plotted for clarity (as usual, $10$ seeds).}
  \label{fig:suppl_explo_metrics_goalim}
\end{figure*}

\clearpage
\paragraph{Details on the impacts of various goal imagination mechanisms on exploration.}
Figure~\ref{fig:suppl_explo_metrics_goalim} presents the \itwoc exploration scores on the training, testing and extra sets for the different goal imagination mechanisms introduced in Main Section~\ref{sec:res_im_properties}. Let us discuss each of these scores:

\begin{enumerate}
    \item \textit{Training interactions.} In Figure~\ref{fig:supp_explo_train}, we see that decreasing the precision (Low Precision and Random Goal conditions) affects exploration on interactions from the training set, where it falls below the exploration of the \textit{no imagination} baseline. This is due to the addition of meaningless goals forcing agent to allow less time to meaningful interactions relatively.
    \item \textit{Testing interactions.}
    In Figure~\ref{fig:supp_explo_test}, we see that the highest exploration scores on interactions from the test set comes from the oracle. Because it shows high coverage and precision, its spends more time on the diversity of interactions from the testing set. What is more surprising is the exploration score of the low coverage condition, higher than the exploration score of \CGH. With an equal precision, \CGH should show better exploration, as it covers more test goals. However, the \textit{Low Coverage} condition, by spending more time exploring each of its imagined goals (it imagined fewer), probably learned to master them better, increasing the robustness of its behavior towards those. This insight advocates for the use of goal selection methods based on learning progress \cite{forestier2016modular,curious}. Agents could estimate their learning progress on imagined goals using their internal reward function and its zero-shot generalization. Focusing on goals associated to high learning progress might help agents filter goals they can learn about from others.
    
    \item \textit{Extra interactions.} Figure~\ref{fig:supp_explo_extra} shows that only the goal imagination mechanisms that invent goals not covered by the testing set manage to boost exploration in this extra set. The oracle perfectly covers the testing set, but does not generate goals related to other objects (e.g. \textit{grow any lamp}). 
\end{enumerate}

\clearpage

\section{Focus on Architectures}
\label{sec:suppl_archi}
This section compares our proposed object-based modular architecture \MA for the policy and reward function to a flat architecture that does not use inductive biases for efficient skill transfer. We hypothesize that only the object-based modular architectures enable a generalization performance that is sufficient for the goal imagination to have an impact on generalization and exploration. Indeed, when generalization abilities are low, agents cannot evaluate their performance on imagined goals and thus, cannot improve.

\begin{figure*}[!h]
    \label{fig:reward_all_archi}
    \centering
    \subfigure[\label{fig:archi_far}\far]{\includegraphics[width=0.42\textwidth]{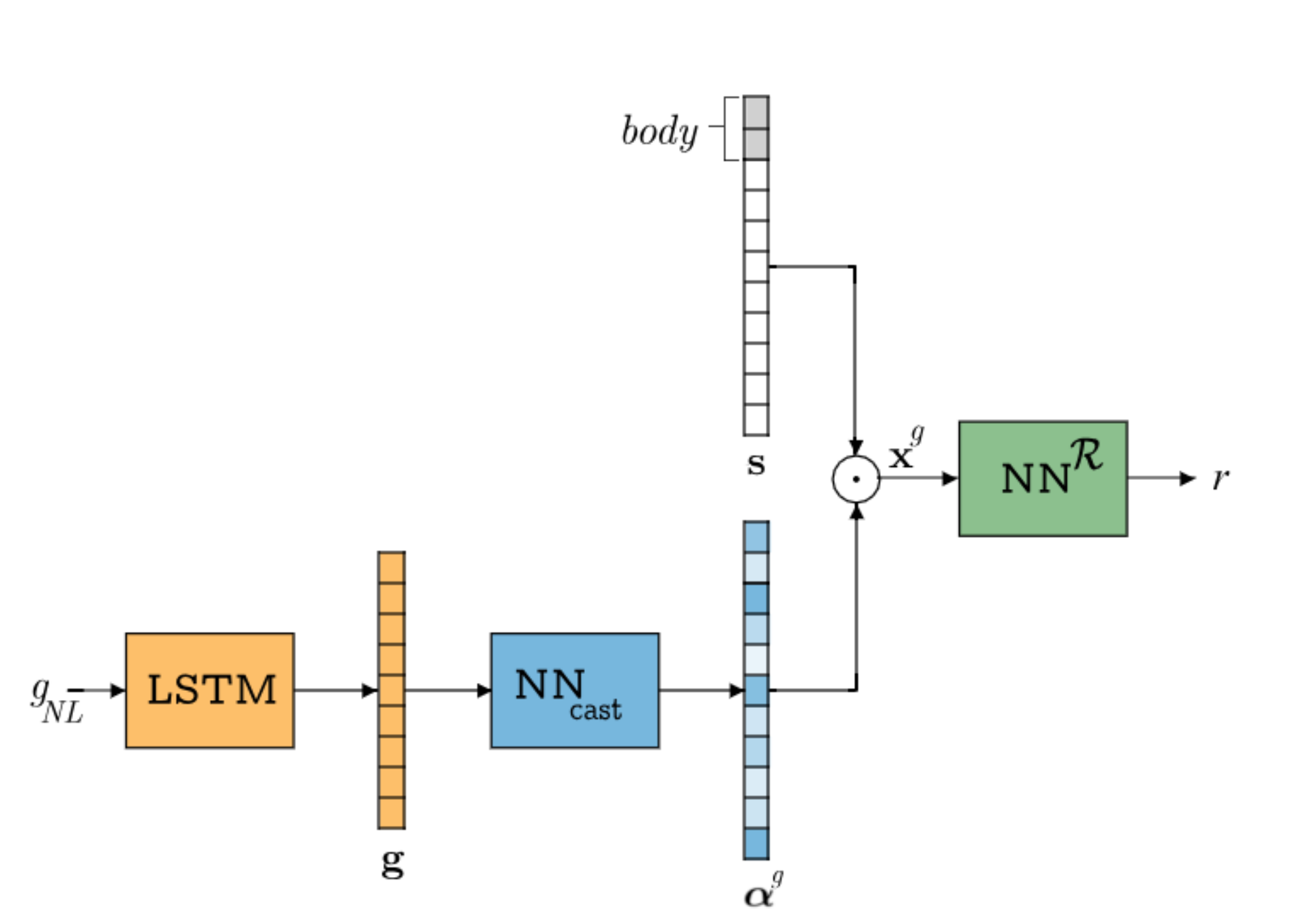}}
    \subfigure[\label{fig:archi_mar}\mar]{\includegraphics[width=0.57\textwidth]{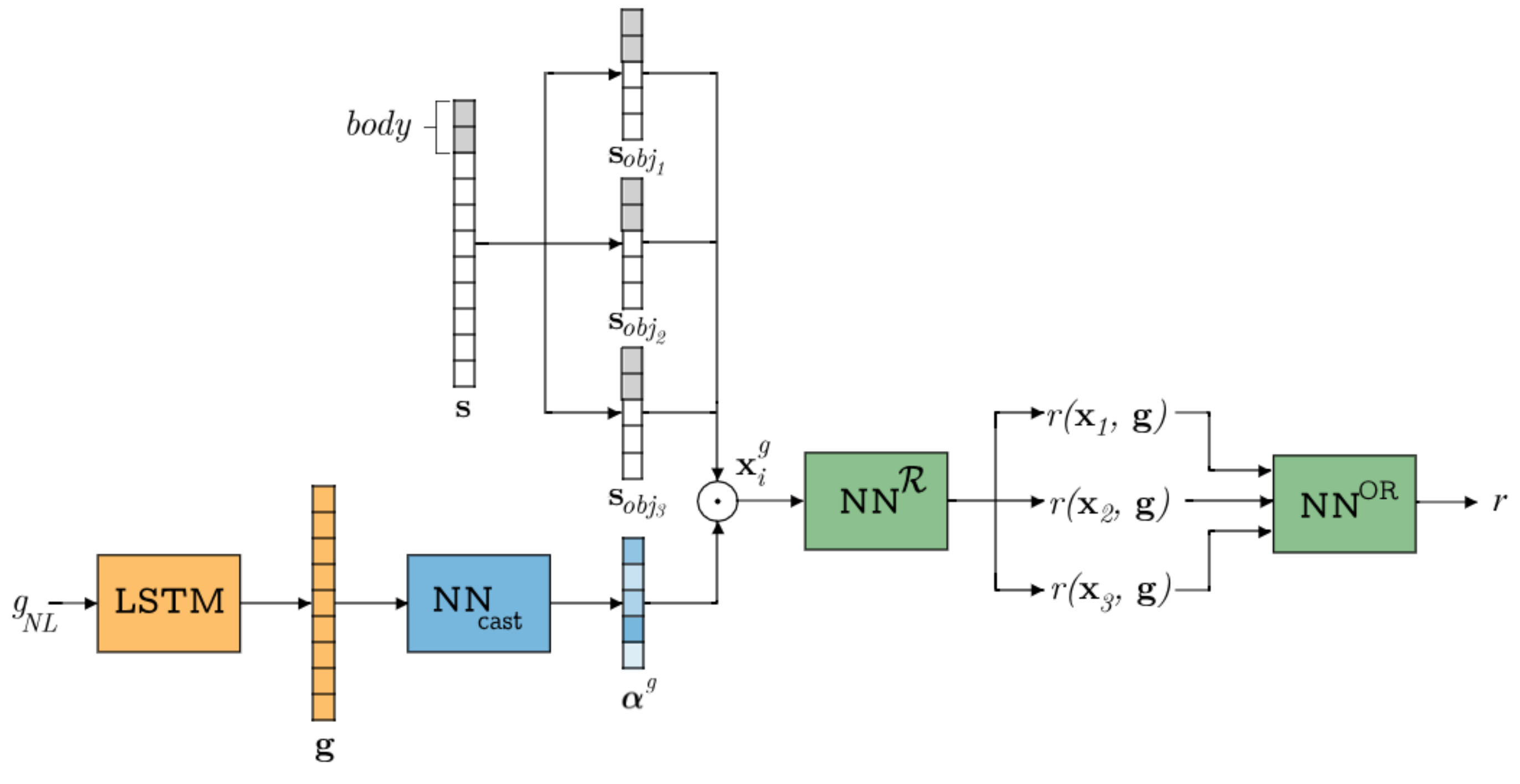}}\\
    \caption{\textbf{Reward function architectures}: (a) \textit{Flat-attention} reward function (\far) and (b) \textit{Modular-attention} reward function (\mar). We use \mar for all experiments except for the experiment in Table~\ref{tab:suppl_reward_function_archi_comparison}
    \label{fig:archi_far_mar}}
\end{figure*}

\paragraph{Preliminary study of the reward function architecture.} We first compared the use of modular and flat architectures for the reward function (\mar vs \far in Figure~\ref{fig:archi_far_mar}). This experiment was conducted independently from policy learning, in a supervised setting. We use a dataset of 50$\times{10^3}$ trajectories and associated goal descriptions collected using a pre-trained policy. To closely match the training conditions of \imagine, we train the reward function on the final states $s_{T}$ and test it on any states $s_t$, $t=[1,~..,~T]$ of other episodes. Table~\ref{tab:suppl_reward_function_archi_comparison} provides the $F_1$ score computed at convergence on $\mathcal{G}^\text{train}$ and $\mathcal{G}^\text{test}$ for the two architectures.

\begin{table}[h!]
    \caption{Reward function architectures performance.}
    \label{tab:suppl_reward_function_archi_comparison}
    \vspace{0.2cm}
    \centering
    \begin{tabular}{l|cc}
    & ${F_1}_\train$ & ${F_1}_\test$\\
    \hline    
    \mar & $0.98 \pm 0.02$ & $0.64 \pm 0.22 $  \\
    \far &  $0.60 \pm 0.10$&$ 0.22 \pm 0.05$ \\
    \end{tabular}
\end{table}

\mar outperforms \far on both the training and testing sets. In addition to its poor generalization performance, \far's performance on the training set are too low to support policy learning. As a result, the remaining experiments in this paper use the \mar architecture for all reward functions. Thereafter, \MA is always used for the reward function and the terms \MA and \FA refer to the architecture of the policy.

\begin{figure*}[!h]
    \label{fig:all_archi}
    \centering
    \subfigure[\label{fig:archi_FA}]{\includegraphics[width=\textwidth]{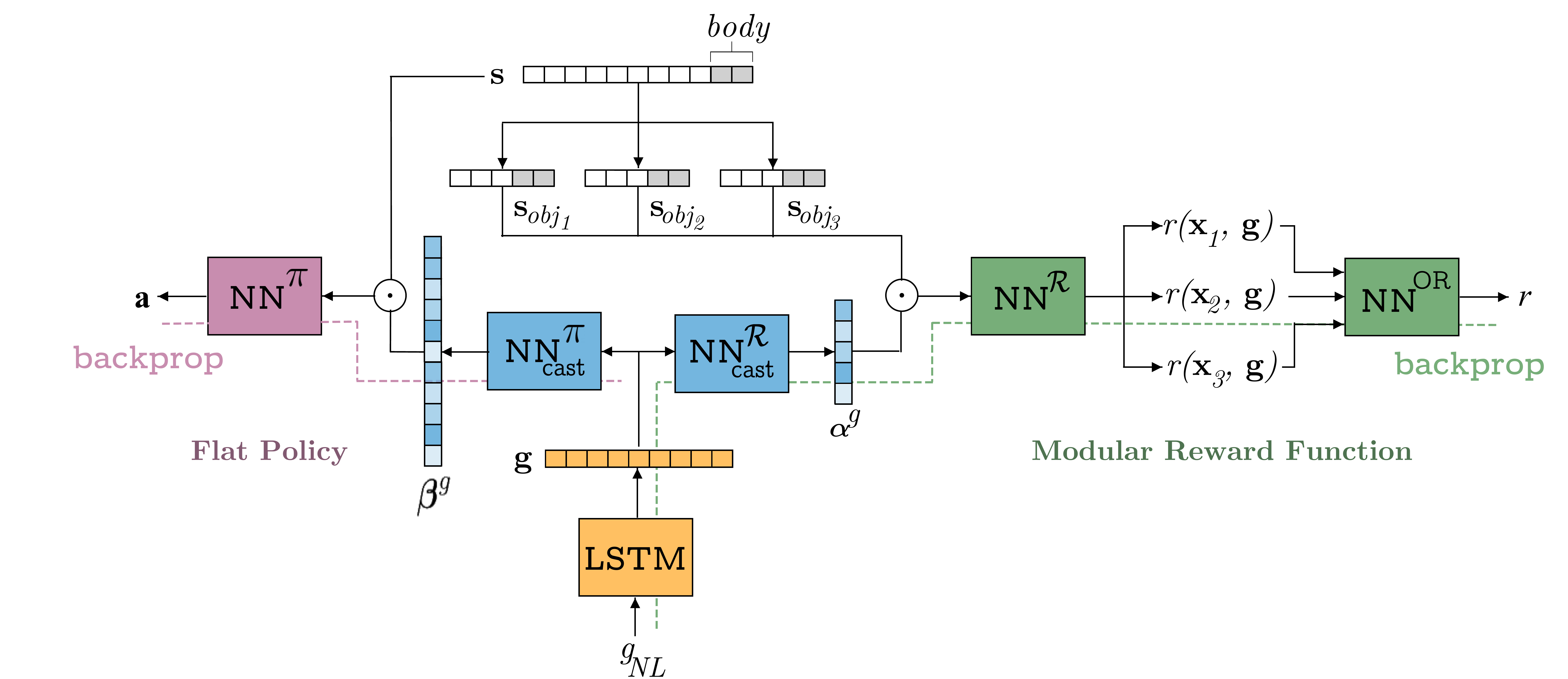}}
    \subfigure[\label{fig:archi_MA}]{\includegraphics[width=\textwidth]{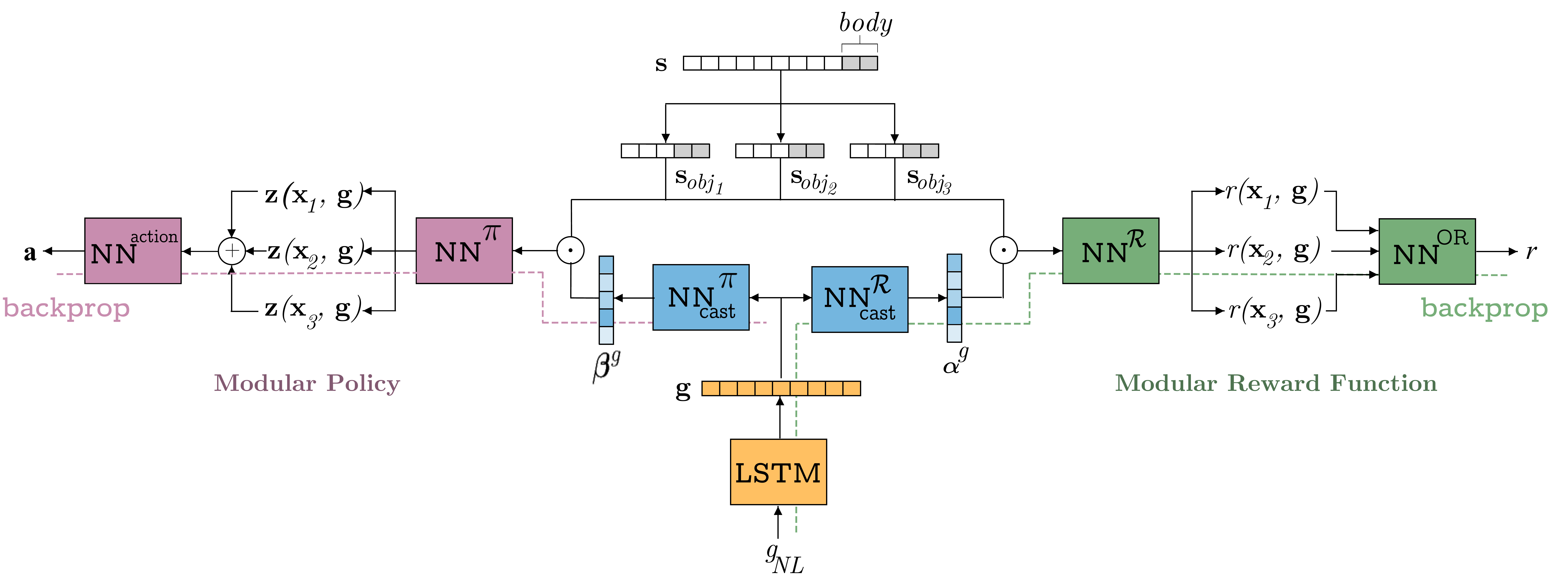}}\\
    \caption{\textbf{Policy and reward function architectures:} (a) \textit{Modular-attention} (\MA) reward + \textit{Flat-attention} (\FA) policy. (b) \MA reward + \MA policy. In both figures, the reward function is represented on the right in green, the policy on the left in pink, the language encoder in the bottom in yellow and the attention mechanisms at the center in blue.
    \label{fig:archi_FA_MA}}
\end{figure*}

\paragraph{Architectures representations.}
The combination of \MA for the reward function and either \MA or \FA for the policy are represented in Figure~\ref{fig:archi_FA_MA}.

\begin{figure}[!h]
    \centering
    \subfigure[\label{fig:suppl_ma_fa_comparison_gene}\textbf{$\SR_\test$}]{\includegraphics[width=0.47\textwidth]{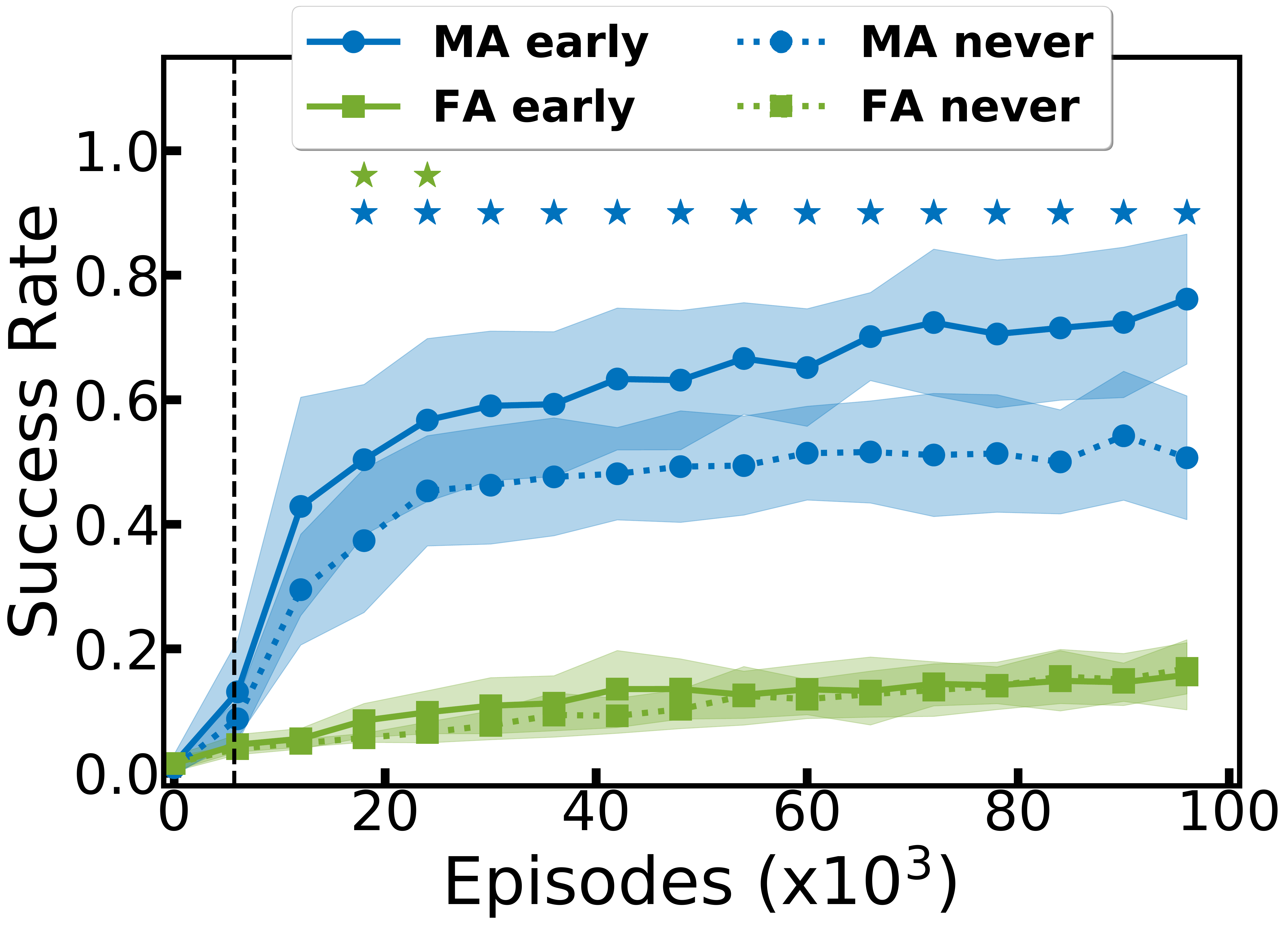}}
    \subfigure[\label{fig:suppl_ma_fa_comparison_i2ctrain}\textbf{$\itwoc_\train$}]{\includegraphics[width=0.47\textwidth]{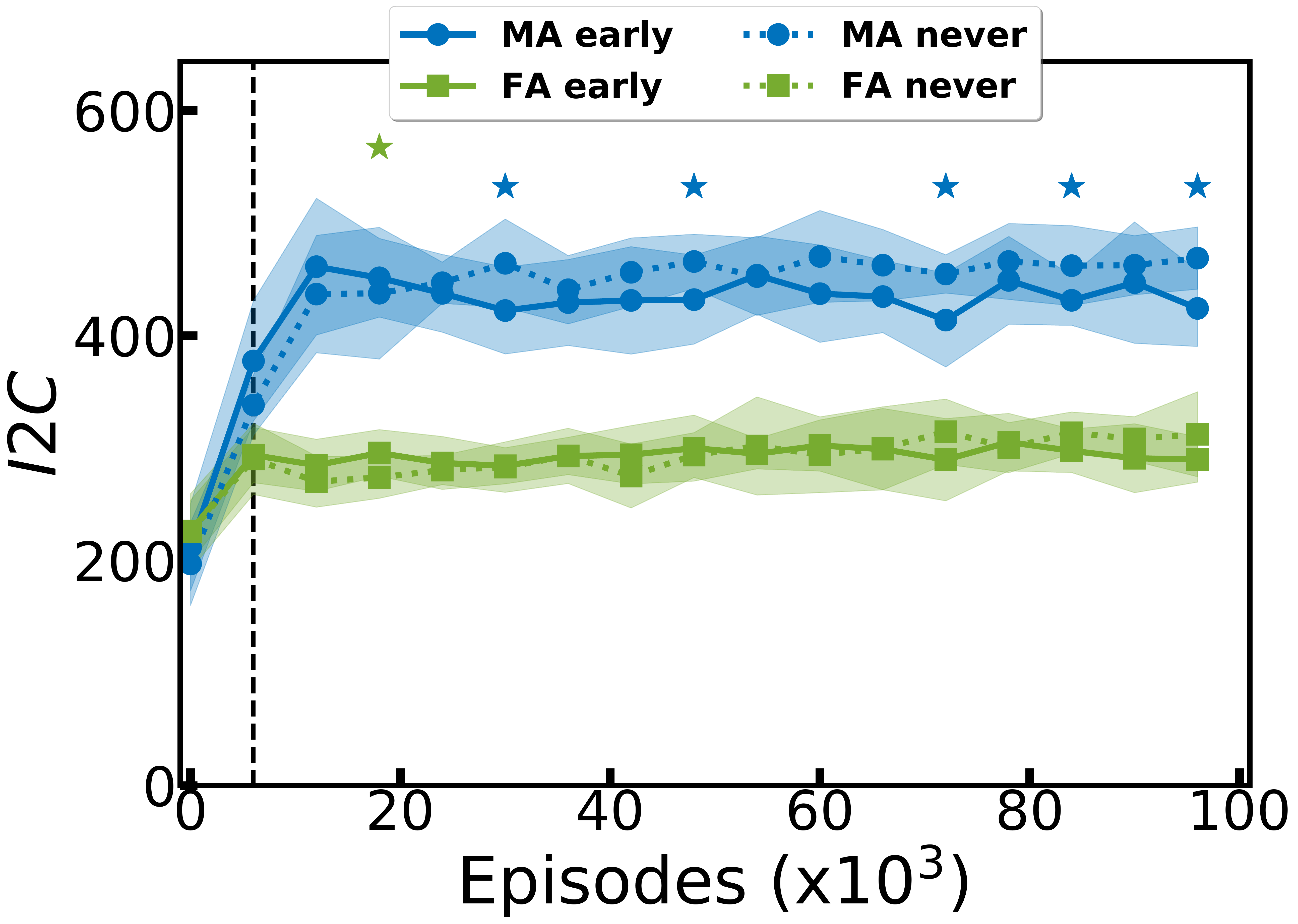}} \\
    \subfigure[\label{fig:suppl_ma_fa_comparison_i2ctest}\textbf{$\itwoc_\test$}]{\includegraphics[width=0.47\textwidth]{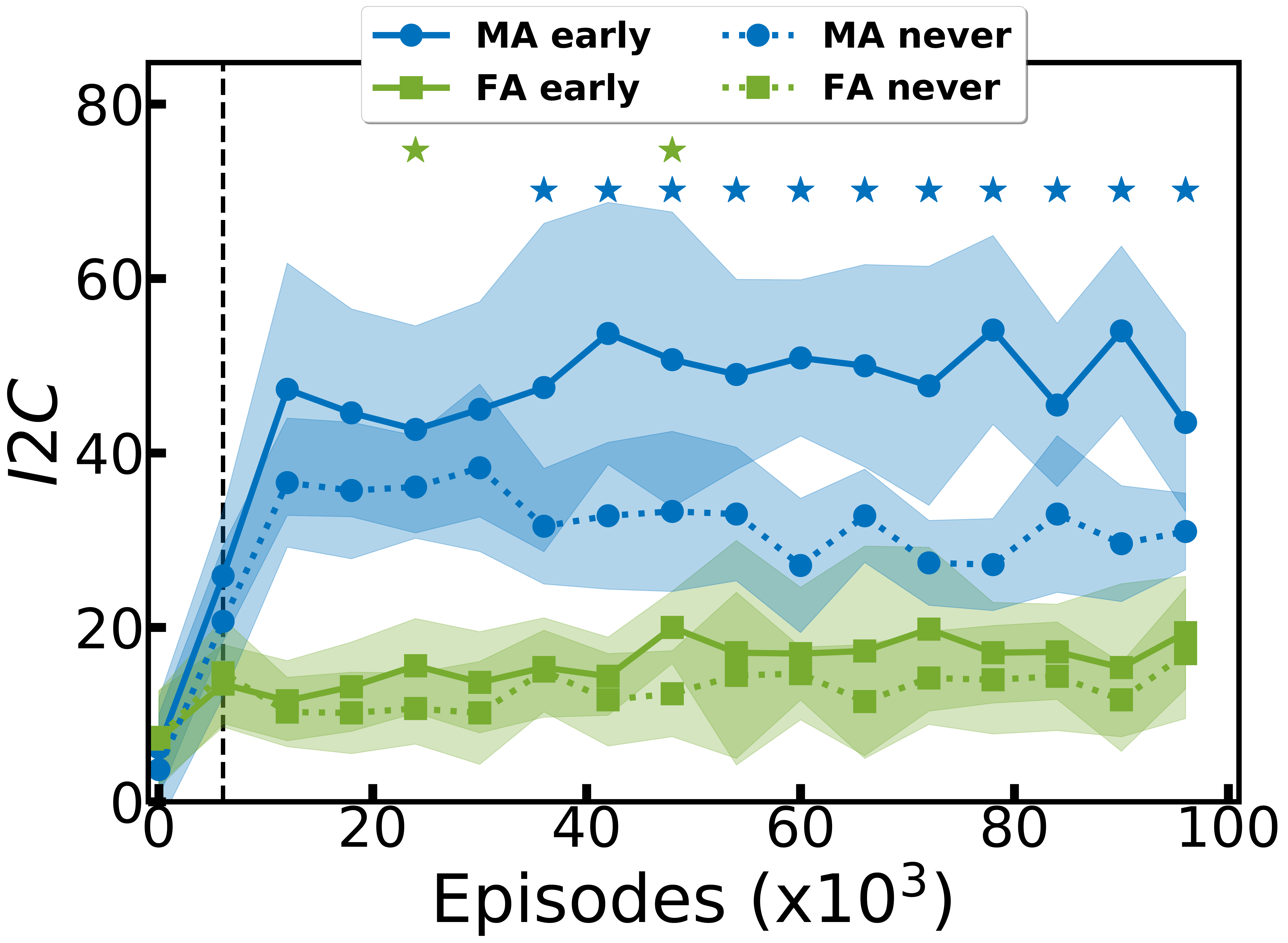}}
    \subfigure[\label{fig:suppl_ma_fa_comparison_i2cextra}\textbf{$\itwoc_\text{extra}$}]{\includegraphics[width=0.47\textwidth]{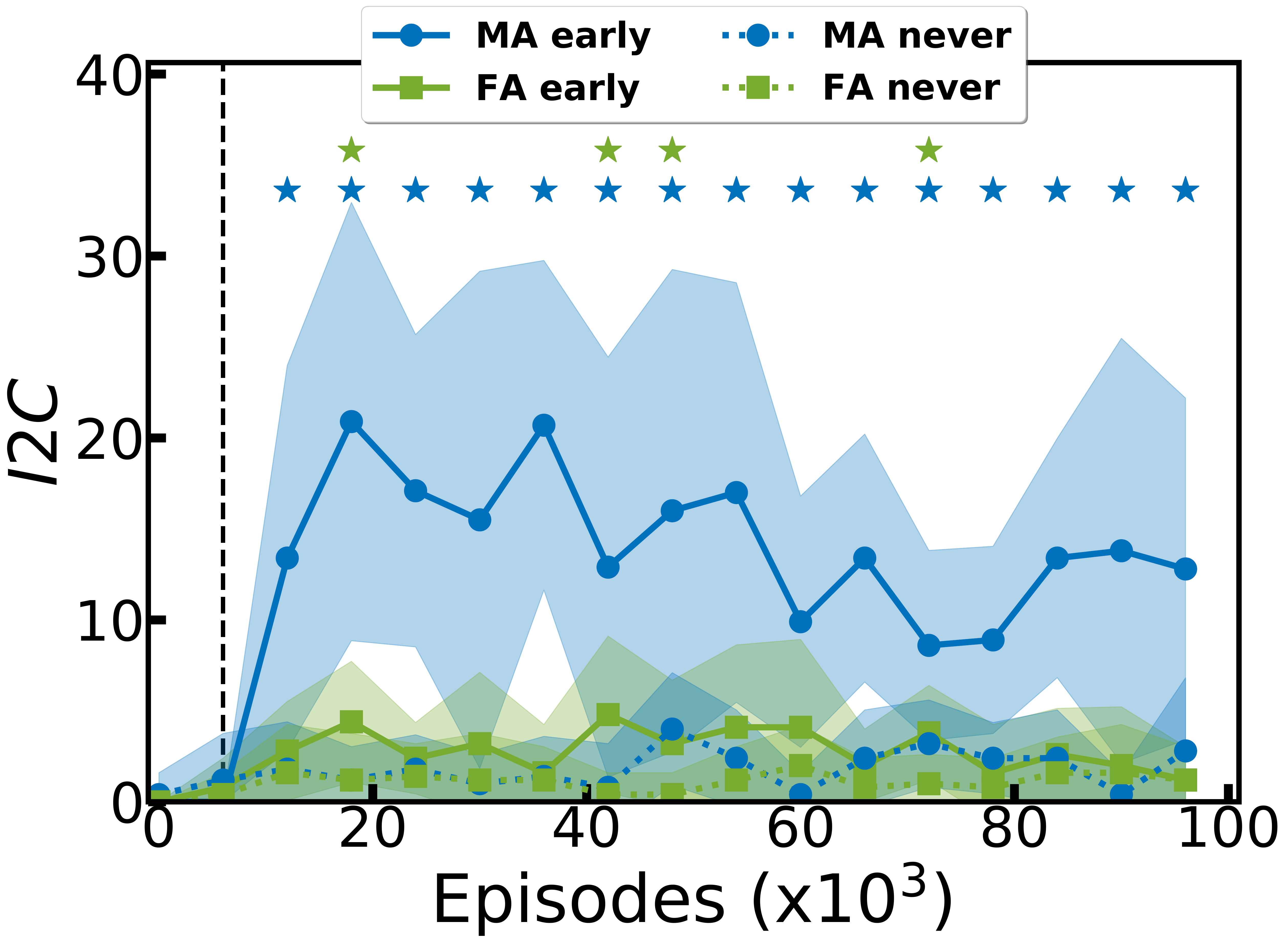}}
    \caption{\textbf{Policy architecture comparison: } (a) $\SR$ on $\G^\test$ for the \FA and \MA architectures when the agent starts imagining goals early (plain, after the black vertical dashed line) or never (dashed). (b, c, d) \itwoc on interactions from the training, testing and extra sets respectively. Imagination is performed using \CGH. Stars indicate significant differences between \CGH and the corresponding \textit{no imagination} baseline.
    \label{fig:suppl_ma_fa_comparison}}
\end{figure}

\paragraph{Policy architecture comparison.} Table~\ref{tab:suppl_archi_comparison} shows that \MA significantly outperforms \FA on both the training and testing sets at convergence. Figure~\ref{fig:suppl_ma_fa_comparison_gene} clearly shows an important gap between the generalization performance of the modular and the flat architecture. In average, less than 20\% of the testing goals can be achieved with \FA when \MA masters half of them without imagination. Moreover, there is no significant difference between the never and the early imagination conditions for the flat architecture. The generalization boost enabled by the imagination is only observable for the modular architecture (see Main Table~\ref{tab:archi_comparison}). Figure~\ref{fig:suppl_ma_fa_comparison_i2ctest} and \ref{fig:suppl_ma_fa_comparison_i2cextra} support similar conclusions for exploration: only the modular architecture enable goal imagination to drive an exploration boost on the testing and extra sets of interactions.

\begin{table}[h!]
    \caption{Architectures performance. Both p-values$~<~10^{-10}$.}
    \label{tab:suppl_archi_comparison}
    \vspace{0.2cm}
    \centering
    \begin{tabular}{l|cc}
    & $\SR_\train$ & $\SR_\test$\\
    \hline    
    \MA & $0.95 \pm 0.05$ & $0.76 \pm 0.10 $  \\
    \FA &  $0.40 \pm 0.13$&$ 0.16 \pm 0.06$ \\
    \end{tabular}
\end{table}

In preliminary experiments, we tested a \textit{Flat-Concatenation} (\textsc{fc}\xspace) architecture where the gated attention mechanism was replaced by a simple concatenation of goal encoding to the state vector. We did not found signficant difference with respect to \FA. We chose to pursue with the attention mechanism, as it improves model interpretability (see Additional Visualization \ref{sec:suppl_visu}).

\clearpage
\section{Focus on Reward Function}
\label{sec:suppl_reward}
Our \imagine agent is autonomous and, as such, needs to learn its own reward function. It does so by leveraging a weak supervision from a social partner that provides descriptions in a simplified language. This reward function can be used for many purposes in the architecture. This paper leverages some of these ideas (the first two), while others are left for future work (the last two):

\begin{itemize}
    \item
    \textbf{Behavior Adaptation.} As Main Section~\ref{sec:res_imag_exp} showed, the reward function enables agents to adapt their behavior with respect to imagined goals. Whereas the zero-shot generalization pushed agents to grow plants with food and water with equal probability, the reward function helped agents to correct that behavior towards more water.
    \item 
    \textbf{Guiding Hindsight Experience Replay (\her).} In multi-goal RL with discrete sets of goals, \her is traditionally used to modify transitions sampled from the replay buffer. It replaces originally targeted goals by others randomly selected from the set of goals \cite{andrychowicz2017hindsight,unicorn}. This enables to transfer knowledge between goals, reinterpreting trajectories in the light of new goals. In that case, a reward function is required to compute the reward associated to that new transition (new goal). To improve on random goal replay, we favor goal substitution towards goals that actually match the state and have higher chance of leading to rewards. In \imagine, we scan a set of $40$ goal candidates for each transition, and select substitute goals that match the scene when possible, with probability $p~=~0.5$.
    \item 
    \textbf{Exploring like Go-Explore.} In Go-Explore \cite{ecoffet2019go}, agents first reach a goal state, then start exploring from there. We could reproduce that behavior in our \imagine agents with our internal reward function. The reward function would scan each state during the trajectory. When the targeted goal is found to be reached, the agent could switch to another goal, add noise on its goal embedding, or increase the exploration noise on actions. This might enable agents to explore sequences of goal-directed behaviors. We leave the study of this mechanism for future work.
    \item 
    \textbf{Filtering of Imagined Goals.} When generating imagined goals, agents also generate meaningless goals. Ideally, we would like agents to filter these from meaningful goals. Meaningful goals, are goals the agent can interpret with its reward function, goals from which it can learn directed behavior. They are interpreted from known related goals via the generalization of the reward function. If we consider an ensemble of reward functions, chances are that all reward functions in the ensemble will agree on the interpretation of meaningful imagined goals. On the other hand, they might disagree on meaningless goals, as their meanings might not be as easily derived from known related goals. Using an ensemble of reward function may thus help agents filter meaningful goals from meaningless ones. This could be done by labeling a dataset of trajectories with positive or negative rewards and comparing results between reward functions, effectively computing agreement measures for each imagined goals. Having an efficient filtering mechanism would drastically improve the efficiency of goal imagination, as Main Section~\ref{sec:res_im_properties} showed that the ratio of meaningful goals determines generalizations performance. This is also left for future work.
\end{itemize}

\clearpage
\section{Additional Visualizations}
\label{sec:suppl_visu}

\begin{figure*}[!hb]
  \centering
  \subfigure[\label{fig:suppl_tsne_predicate}]{\includegraphics[width=0.323\textwidth]{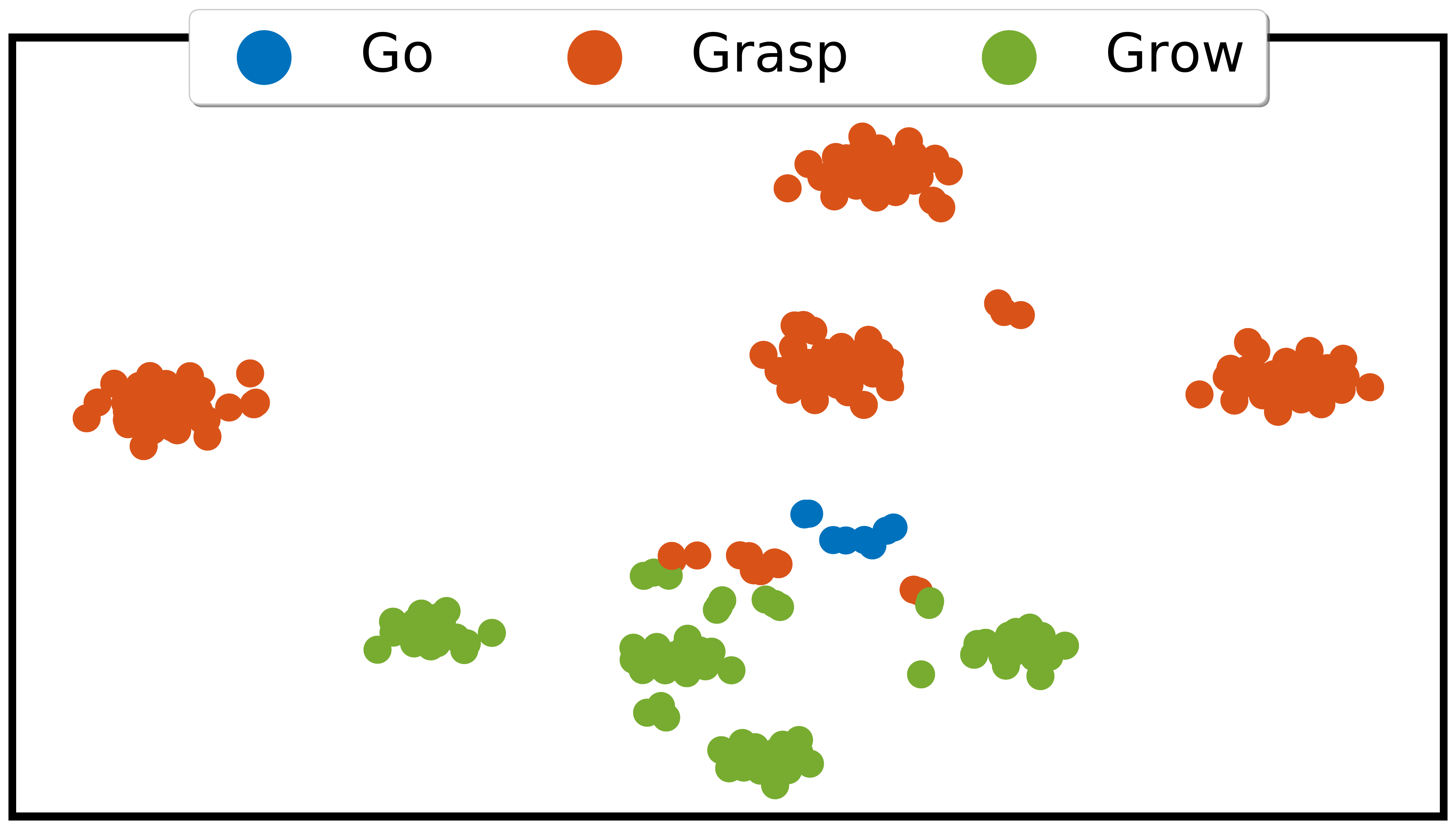}}
  \subfigure[\label{fig:suppl_tsne_color}]{\includegraphics[width=0.323\textwidth]{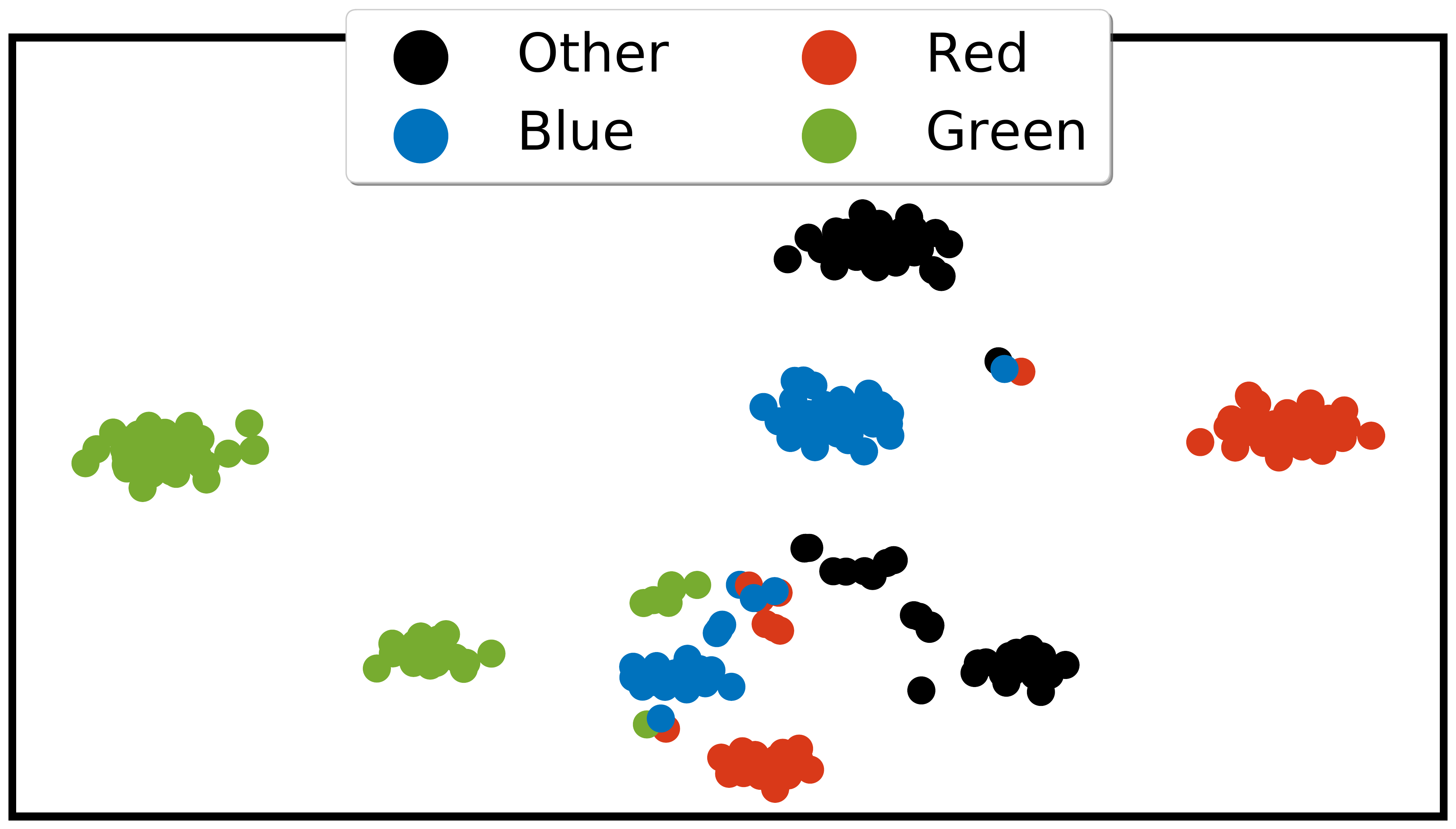}}
  \subfigure[\label{fig:suppl_tsne_cat}]{\includegraphics[width=0.323\textwidth]{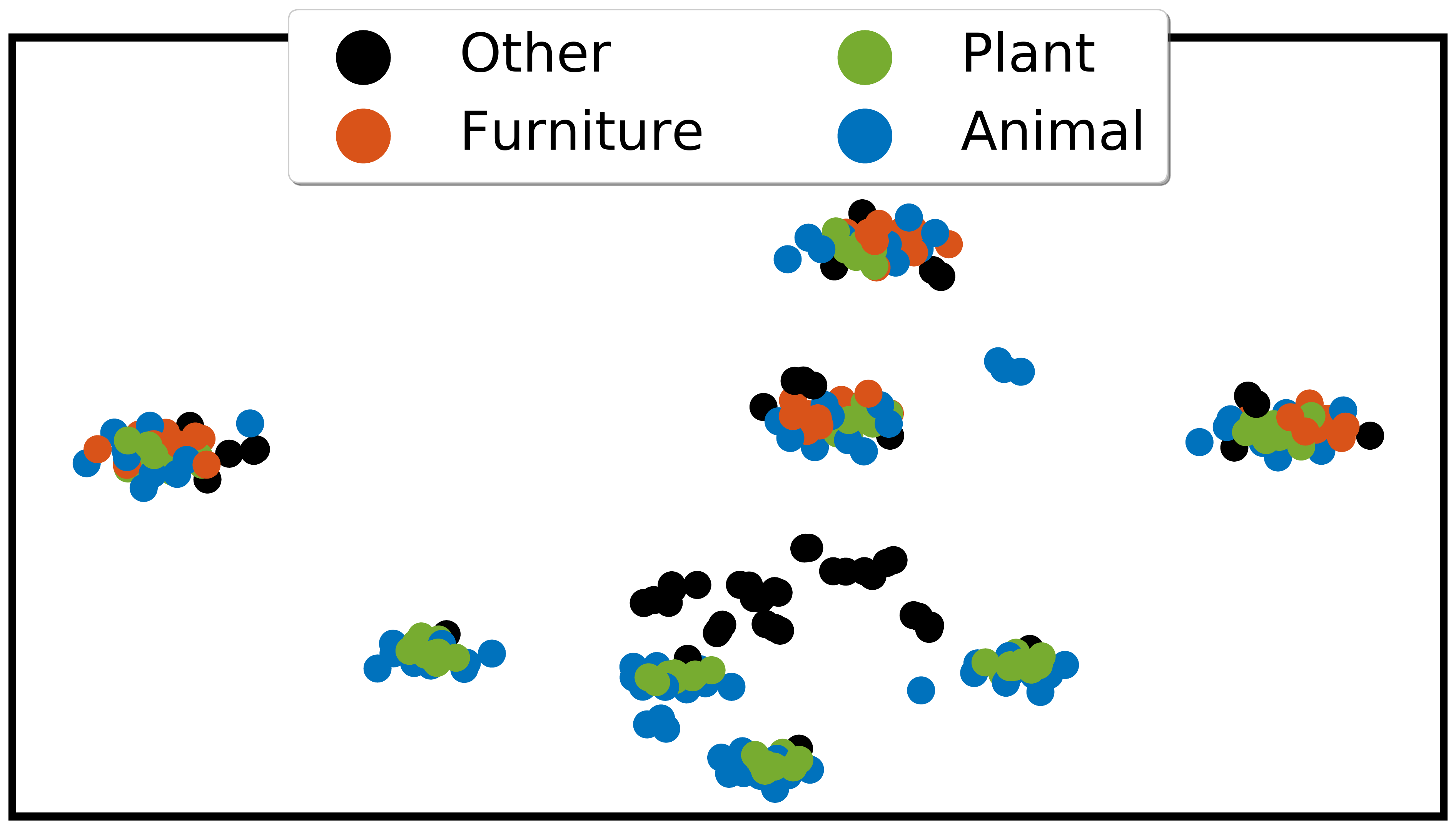}}
  \caption{\textbf{t-SNE of Goal Embedding.} The same t-SNE is presented, with different color codes (a) predicates, (b) colors, (c) object categories.}
  \label{fig:suppl_tsne}
\end{figure*}

\paragraph{Visualizing Goal Embedding}
To analyze the goal embeddings learned by the language encoder $L_e$, we perform a t-SNE using $2$ components, perplexity $20$, a learning rate of $10$ for $5000$ iterations. Figure~\ref{fig:suppl_tsne} presents the resulting projection for a particular run. The embedding seems to be organized mainly in terms of motor predicates (\ref{fig:suppl_tsne_predicate}), then in terms of colors (\ref{fig:suppl_tsne_color}). Object types or categories do not seem to be strongly represented (\ref{fig:suppl_tsne_cat}). 

\paragraph{Visualizing Attention Vectors}

\begin{figure*}[!htbp]
  \centering
   \subfigure[\label{fig:att_rew}]{\includegraphics[width=0.7\textwidth]{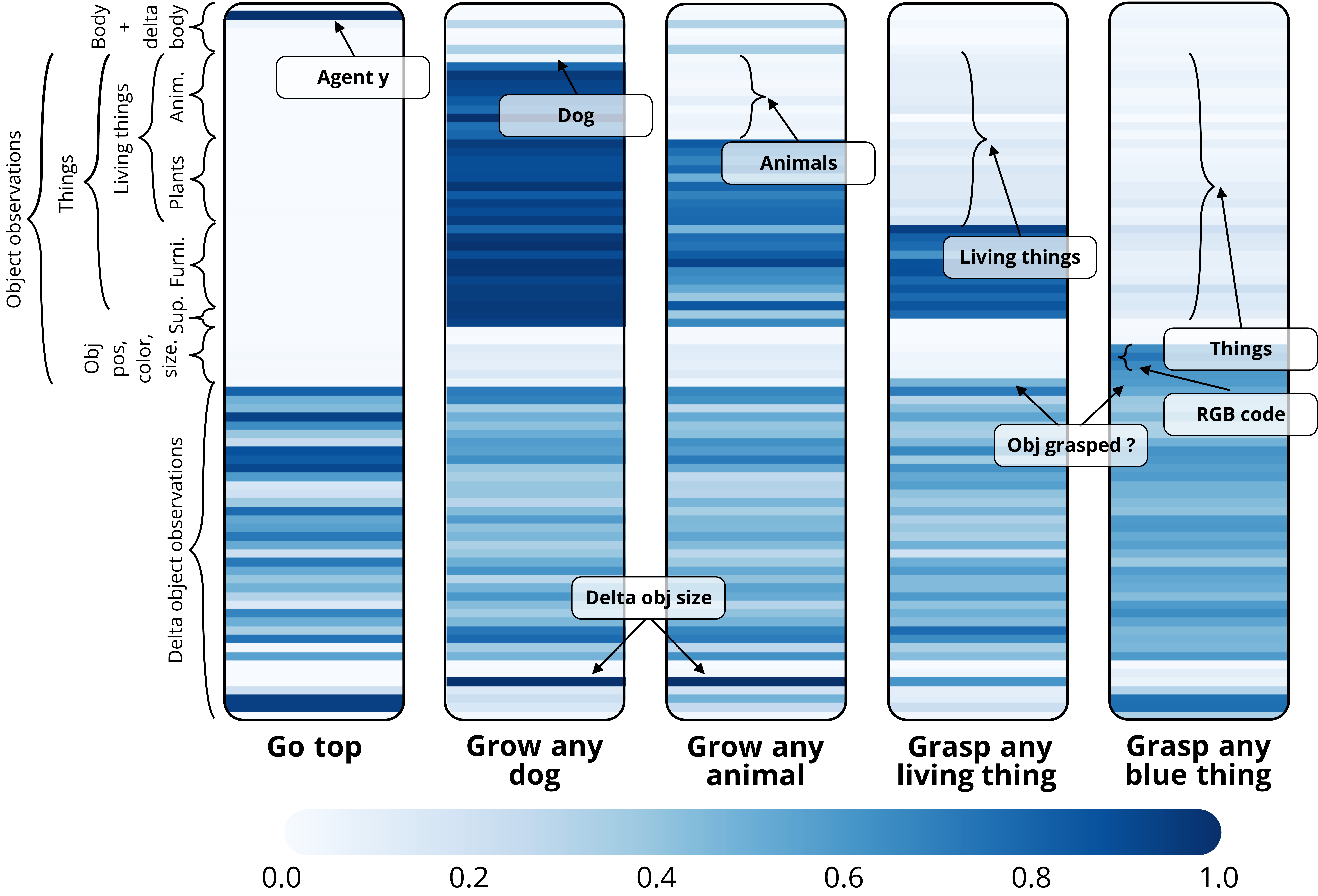}}\\
   \subfigure[\label{fig:att_pol}]{\includegraphics[width=0.7\textwidth]{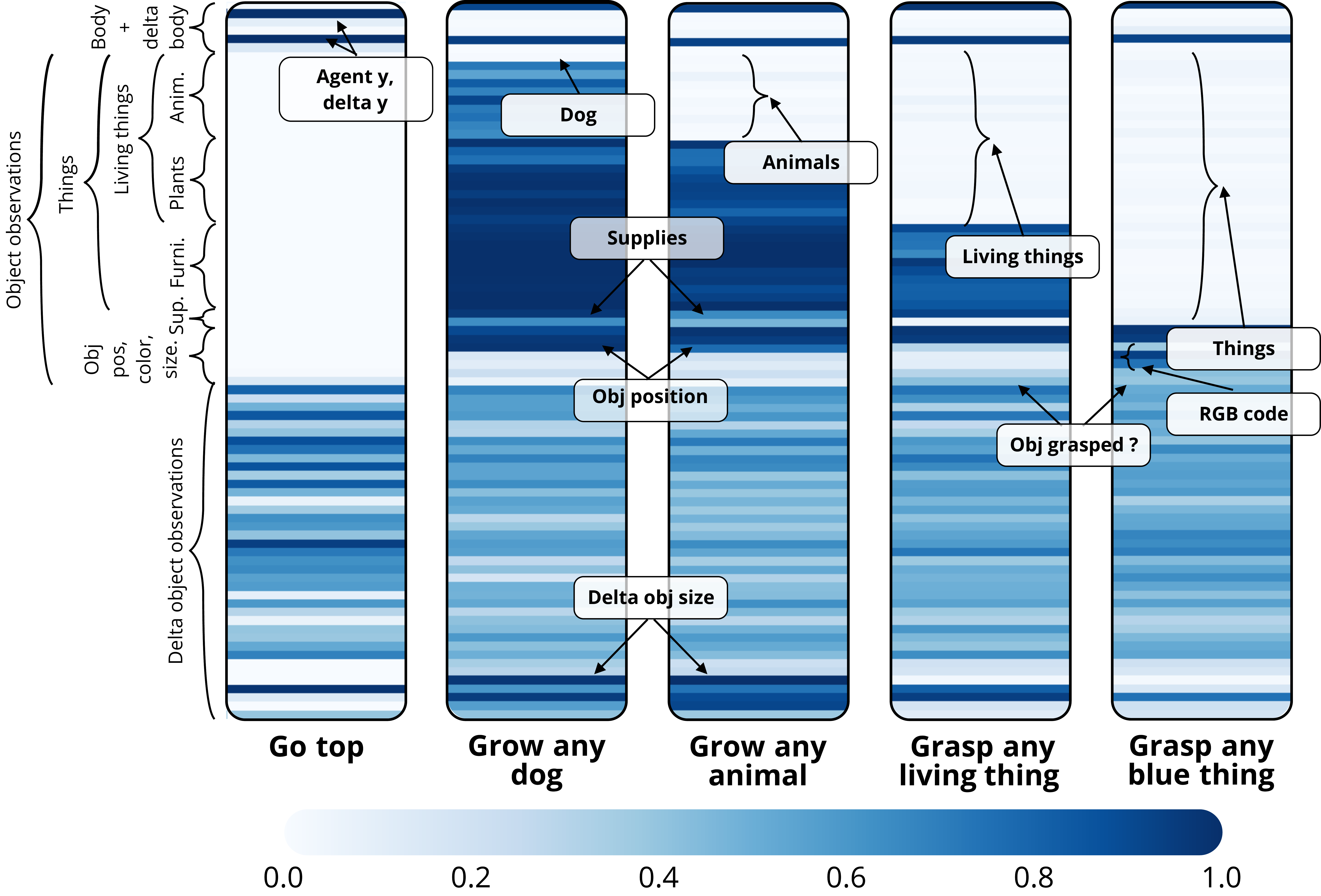}}\\
  \caption{\textbf{Attention vectors} (a) $\balpha^g$ for the reward function ($1$ seed). (b) $\bbeta^g$ for the policy ($1$ seed).}
  \label{fig:att}
\end{figure*}

In the \textit{modular-attention} architectures for the reward function and policy, we train attention vectors to be combined with object-specific features using a gated attention mechanism. In each architecture, the attention vector is shared across objects (permutation invariance). Figure~\ref{fig:att} presents examples of attention vectors for the reward function (\ref{fig:att_rew}) and for the policy (\ref{fig:att_pol}) at the end of training. These attention vectors highlight relevant parts of the object-specific sub-state depending on the \NL  goal:

\begin{itemize}
    \item When the sentence refers to a particular object type (e.g. \textit{dog}) or category (e.g. \textit{living thing}), the attention vector suppresses the corresponding object type(s) and highlights the complement set of object types. If the object does not match the object type or category described in the sentence, the output of the Hadamard product between object types and attention will be close to $1$. Conversely, if the object is of the required type, the attention suppression ensures that the output stays close to zero. Although it might not be intuitive for humans, it efficiently detects whether the considered object is the one the sentence refers to.
    \item When the sentence refers to a navigation goal (e.g. \textit{go top}, the attention highlights the agent's position (here $y$).
    \item When the sentence is a \textit{grow} goal, the reward function focuses on the difference in object's size, while the policy further highlights the object's position.
\end{itemize}
The attention vectors uses information about the goal to highlight or suppress parts of the input using the different strategies described above depending on the type of input (object categories, agent's position, difference in size etc). This type of gated-attention improves the interpretability of the reward function and policy. 

\clearpage
\section{Comparing IMAGINE to goal-as-state approaches.}
\label{sec:suppl_discu}
In the goal-conditioned RL literature, some works have proposed goal generation mechanisms to facilitate the acquisition of skills over large sets of goals \cite{nair2018visual,pong2019skew,curious,nair2019contextual}. Some of them had a special interest in exploration, and proposed to bias goal sampling towards goals from low density areas \cite{pong2019skew}. One might then think that \imagine should be compared to these approaches. However, there are a few catches: 
\begin{enumerate}
    \item 
    \citet{nair2018visual,nair2019contextual,pong2019skew} use generative models of states to sample state-based goals. However, our environment is procedurally generated. This means that sampling a given state from the generative model has a very low probability to \textit{match} the scene. If the present objects are three red cats, the agent has no chance to reach a goal specifying dogs and lions' positions, colors and sizes. Indeed, most of the state space is made of  object features that cannot be acted upon (colors, types, sizes of most objects). One could imagine using \SP to organize the scene, but we would need to ask \SP to find the three objects specified by the generated goal, in the exact colors (RGB codes) and size. Doing so, there would be no distracting object for agent to discover and learn about. A second option is to condition the goal generation on the scene as it is done in \citet{nair2019contextual}. The question of whether it might work in procedurally-generated environments remains open.
    \item 
    Assuming a perfect goal generator that only samples valid goals that do not ask a change of object color or type, the agent would then need to bring each object to its target position and to grow objects to their very specific goal size. These goals are not the same as those targeted by \imagine, they are too specific. These approaches --like most goal-conditioned RL approaches-- represent goals as particular states (e.g. block positions in manipulation tasks, visual states in navigation tasks) \cite{schaul2015universal,andrychowicz2017hindsight,nair2018visual,pong2019skew,curious}. In contrast, language-conditioned agents represent abstract goals, usually defined by specific constraints on states (e.g. \textit{grow any plant} requires the size of at least one plant to increase) \cite{chan2019actrce,Jiang2019,ther}. 
    For this reason, \textit{goal-as-state} and \textit{abstract goal} approaches do not tackle the same problem. The first targets specific coordinates, and cannot be instructed to reach abstract goals, while the second are not trained to reach specific states.
\end{enumerate}

For these reasons, we argue that the goal-conditioned approaches that use state-based goals cannot be easily or fairly compared to our approach \imagine.


\clearpage
\section{Implementation details}
\label{sec:supp_impl_details}


\paragraph{Reward function inputs and hyperparameters.} Supplementary Section~\ref{sec:suppl_archi} details the architecture of the reward function. The following provides extra details about the inputs. The object-dependent sub-state $\mathbf{s}_{obj(i)}$ contains information about both the agent's body and the corresponding object $i$:  $\mathbf{s}_{obj(i)}=[\mathbf{o}_{body}, \Delta{\mathbf{o}_{body}}, \mathbf{o}_{obj(i)}, \Delta{\mathbf{o}_{obj(i)}}]$ where $\mathbf{o}_{body}$ and $\mathbf{o}_{obj(i)}$ are body- and $obj_i$-dependent observations, and $\Delta{\mathbf{o}^t_{body}}~=~\mathbf{o}_{body}^t-\mathbf{o}_{body}^0$ and $\Delta{\mathbf{o}^t_{obj(i)}}~=~\mathbf{o}_{obj(i)}^t-\mathbf{o}_{obj(i)}^0$ measure the difference between the initial and current observations. The second input is the attention vector $\balpha^g$ that is integrated with $\mathbf{s}_{obj(i)}$ through an Hadamard product to form the model input: $\mathbf{x}_i^g=\mathbf{s}_{obj(i)} \odot \balpha^g$. This attention vector is a simple mapping from $\textbf{g}$ to a vector of the size of $\mathbf{s}_{obj(i)}$ contained in $[0,1]^{size(\mathbf{s}_{obj(i)})}$. This cast is implemented by a one-layer neural network with sigmoid activations $\texttt{NN}^\text{cast}$ such that $\balpha^g=\texttt{NN}^\text{cast}(\mathbf{g})$.

For the three architectures the number of hidden units of the \texttt{LSTM} and the sizes of the hidden layers of fully connected networks are fixed to $100$. \texttt{NN} parameters are initialized using He initialization \cite{he} and we use one-hot word encodings. The \texttt{LSTM} is implemented using \texttt{rnn.BasicLSTMCell} from tensorflow 1.15 based on \citet{zaremba2014recurrent}. The states are initially set to zero. The \texttt{LSTM}'s weights are initialized uniformly from $[-0.1,0.1]$ and the biases initially set to zero. The \texttt{LSTM} use a $tanh$ activation function whereas the \texttt{NN} are using ReLU activation functions in their hidden layers and sigmoids at there output.

\paragraph{Reward function training schedule.} The architecture are trained via backpropagation using the Adam Optimizer \cite{kingma2014adam}. The data is fed to the model in batches of $512$ examples. Each batch is constructed so that it contains at least one instance of each goal description $g_\text{NL}$ (goals discovered so far). We also use a modular buffer to impose a ratio of positive rewards of $0.2$ for each description in each batch. When trained in parallel of the policy, the reward function is updated once every $1200$ episodes. Each update corresponds to up to $100$ training epochs ($100$ batches). We implement a stopping criterion based on the $F_1$-score computed from a held-out test set uniformly sampled from the last episodes ($20\%$ of the last $1200$ episodes (2 epochs)). The update is stopped when the $F_1$-score on the held-out set does not improve for $10$ consecutive training epochs.

\paragraph{RL implementation and hyperparameters.} In the policy and critic architectures, we use hidden layers of size $256$ and ReLU activations. Attention vectors are cast from goal embeddings using single-layer neural networks with sigmoid activations. We use the He initialization scheme for \cite{he} and train them via backpropagation using the Adam optimizer ($\beta_1=0.9, \beta_2=0.999$) \cite{kingma2014adam}.

Our learning algorithm is built on top of the OpenAI Baselines implementation of \her-\ddpg.\footnote{ The OpenAI Baselines implementation of \her-\ddpg can be found at https://github.com/openai/baselines, our implementation can be found at \url{https://sites.google.com/view/imagine-drl}.\footnote{Link to our Github repository will be added in the final version.}} We leverage a parallel implementation with $6$ actors. Actors share the same policy and critic parameters but maintain their own memory and conduct their own updates independently. Updates are then summed to compute the next set of parameters broadcast to all actors. Each actor is updated for $50$ epochs with batches of size $256$ every $2$ episodes of environment interactions. Using hindsight replay, we enforce a ratio $p=0.5$ of transitions associated with positive rewards in each batch. We use the same hyperparameters as \citet{plappert2018multi}.

\paragraph{Computing resources.}
The RL experiments contain $8$ conditions of $10$ seeds each, and $4$ conditions with $5$ seeds (\SP study). Each run leverages $6$ cpus ($6$ actors) for about $36$h for a total of $2.5$ cpu years. Experiments presented in this paper requires machines with at least $6$ cpu cores.

\end{document}